\documentclass[preprint,11pt,a4paper]{elsarticle}
\biboptions{numbers,sort&compress}
\usepackage[a4paper,margin=25mm]{geometry}
\usepackage{lmodern}

\usepackage{amssymb}
\usepackage{amsmath}
\usepackage{amsthm}
\usepackage{mathtools}
\usepackage{booktabs}
\usepackage{tabularx}
\usepackage{longtable}
\usepackage{array}
\usepackage{enumitem}
\usepackage{xcolor}
\usepackage{microtype}
\usepackage[section]{placeins}
\usepackage{flafter}
\usepackage{tikz}
\usetikzlibrary{arrows.meta,positioning,fit,calc}
\usepackage{listings}
\usepackage{xr-hyper}
\usepackage[hidelinks]{hyperref}
\usepackage{xurl}
\usepackage{comment}
\usepackage[nameinlink,noabbrev]{cleveref}
\journal{Artificial Intelligence}

\newcommand{\artifacturl}{\url{http://staff.ustc.edu.cn/~jianmin/rat/}}

\theoremstyle{definition}
\newtheorem{definition}{Definition}

\theoremstyle{plain}
\newtheorem{proposition}{Proposition}
\newtheorem{theorem}{Theorem}
\newtheorem{lemma}{Lemma}
\newtheorem{corollary}{Corollary}
\theoremstyle{remark}

\newcommand{\Dof}[1]{\mathcal{D}[#1]}
\newcommand{\Modules}{\mathcal{M}}
\newcommand{\Config}{\mathcal{C}}

\newcommand{\cl}{\mathsf{Cl}}
\newcommand{\enabled}{\mathsf{Enabled}}

\newcommand{\trans}[1]{\xrightarrow{#1}}

\tikzset{
  ratbox/.style={draw,rounded corners,align=center,inner sep=5pt,minimum height=9mm},
  ratarr/.style={-{Latex[length=2mm]},thick},
  ratdash/.style={-{Latex[length=2mm]},thick,dashed}
}

\lstdefinestyle{asp}{
  language=Prolog,
  basicstyle=\ttfamily\small,
  columns=fullflexible,
  frame=single,
  framerule=0.3pt,
  breaklines=true,
  showstringspaces=false,
  commentstyle=\color{gray!70!black}
}

\begin{document}
\begin{frontmatter}
\title{Reconfigurable Action Theories: A Formal Foundation for Dynamically Composable Agent Harnesses}
\author[1]{Jianmin Ji\corref{cor1}}
\ead{jianmin@ustc.edu.cn}
\affiliation[1]{organization={University of Science and Technology of China},
  city={Hefei},postcode={230026},country={China}}
\cortext[cor1]{Corresponding author.}

\begin{abstract}
Action theories typically fix their vocabulary and causal laws, whereas agent harnesses can change their tools, providers, and policies during execution.
We introduce Reconfigurable Action Theories (RAT), in which actions can change the action theory itself.
A theory module in RAT contains a vocabulary, causal laws, and an initializer. A RAT configuration records the active modules and the world state. Module loading and unloading are represented by actions governed by ordinary causal laws.
RAT retains the standard causal action semantics of its language fragment when the active module set is fixed.
We give sufficient conditions for at most one next configuration, preservation of existing behavior when a module is added, and equal results when modules are loaded in either order.
We establish the computational complexity of reasoning tasks in RAT and translate finite RAT systems into nonmonotonic causal theories and Answer Set Programming (ASP), preserving bounded executions.
Experiments compare finite RAT models with executions of official DeepSeek Harness, Pi, and OpenHands implementations.
RAT provides a formal basis for checking how component changes affect a harness's available actions, behavior, and reachable goals.
\end{abstract}
\end{frontmatter}
\hypersetup{pdftitle={Reconfigurable Action Theories: A Formal Foundation for Dynamically Composable Agent Harnesses},pdfauthor={Jianmin Ji},pdfsubject={Module change in causal action theories}}

\section{Introduction}
\label{sec:introduction}

Action theories describe how actions change a world. A conventional action
description fixes a vocabulary of fluents, which represent changeable facts,
and actions, together with the causal laws governing them, while states or
histories evolve
~\cite{gelfondLifschitz1998,giunchigliaLifschitz1998,mccainTurner1997,reiter2001,baral2003}.
We study actions that also change the vocabulary and causal laws used to
describe later actions.

Agentic AI provides a concrete setting for this problem. An agent repeatedly
proposes actions, invokes tools, observes results, and updates its context
while pursuing a goal
~\cite{wangEtAl2024,qinEtAl2025,yao2023react}. The surrounding \emph{agent
harness} constructs model context, exposes and validates tools, selects
providers, dispatches calls, maintains sessions, enforces policies, and
records effects~\cite{liEtAl2026Harness}. Several harnesses support
runtime changes to components, tools, or provider bindings
~\cite{deepseekArchitecture,piSdk,openclawArchitecture,codexAppServer}.
For example, loading a tool can introduce new calls, while replacing a
provider or changing a policy can alter when a call is permitted and
what it does. To reason about these changes, we need to represent both
the current action description and the actions that change it.

Consider a ReAct agent~\cite{yao2023react} asked to compute \(23\times17\)
when no calculator tool is initially available.
Loading a calculator module should introduce its multiplication action, causal laws, and
private state. Removal should withdraw that contribution. A fixed description can encode
availability through fluents and guarded, predeclared laws, but the
vocabulary and laws themselves remain fixed. We make the module structure
explicit: which action-description fragments are in force, and how actions
change their selection.

We introduce \emph{Reconfigurable Action Theories} (RAT), a minimal extension
in which actions can change the action theory itself, using the ordinary
nonmonotonic causal foundation~\cite{mccainTurner1997,giunchigliaEtAl2004}.
A \emph{theory module} is a named action-description fragment. A
\emph{configuration} \(\Config=\langle K,s\rangle\) records active optional
modules \(K\) and world state \(s\), alongside an always-active ``core''.
The actions \(\mathit{load}(M)\) and \(\mathit{unload}(M)\) select
or deselect module \(M\), without rewriting its laws. Ordinary causal laws
govern their world effects and qualification, while RAT fixes their effect on \(K\).

Runtime theory change itself is not new. Modular action languages assemble
reusable descriptions, while evolving logic programs, domain update, and
action-theory revision address changes of rules or theories
~\cite{lifschitzRen2006,inclezanGelfond2016,alferesBantiBrogi2005,eiterErdemFinkSenko2010,varzinczak2010}.
RAT instead makes module selection explicit, distinguishes its roles before
and after an action, and studies preservation and reordering of module
changes. For a fixed execution bound, a causal theory can represent
the same executions.
RAT keeps module declarations and their changes explicit, so conditions
for preserving behavior can be stated in terms of the modules rather
than the auxiliary rules of an encoding. Section~\ref{sec:rat-value}
discusses this role of RAT alongside its causal-theory translation.

\begin{figure}[htbp]
  \centering
  \resizebox{0.92\linewidth}{!}{%
  \begin{tikzpicture}[node distance=9mm and 12mm, every node/.style={font=\small}]
    \node[ratbox,fill=blue!5] (cfg) {modules active\\before \(a_t\): \(K_t\)};
    \node[ratbox,fill=green!5,right=of cfg] (world) {source state\\\(s_t\)};
    \node[ratbox,fill=orange!8,right=25mm of world] (cfg2) {modules active\\after \(a_t\): \(K_{t+1}\)};
    \node[ratbox,fill=green!5,right=of cfg2] (world2) {target state\\\(s_{t+1}\)};
    \node[draw,dashed,rounded corners,fit=(cfg)(world),inner sep=4pt,label=below:{before the action}] {};
    \node[draw,dashed,rounded corners,fit=(cfg2)(world2),inner sep=4pt,label=below:{after the action}] {};
    \draw[ratarr] ($(world.east)+(1mm,0)$) -- node[above=1pt,fill=white,inner sep=1pt]{perform \(a_t\)} ($(cfg2.west)+(-1mm,0)$);
    \draw[ratdash,bend left=24] (cfg.north) to node[above]{check action; derive direct effects} (world.north);
    \draw[ratdash,bend left=24] (cfg2.north) to node[above]{derive static consequences} (world2.north);
  \end{tikzpicture}%
  }
  \caption{An action may change both the world state and the action theory.
  The description \(\Dof{K_t}\), formed by the core and the modules in
  \(K_t\), supplies the executability and direct-effect laws for \(a_t\).
  Loading or unloading changes \(K_t\) to \(K_{t+1}\); the resulting
  description \(\Dof{K_{t+1}}\) supplies the static laws governing
  \(s_{t+1}\).}
  \label{fig:coupled}
\end{figure}
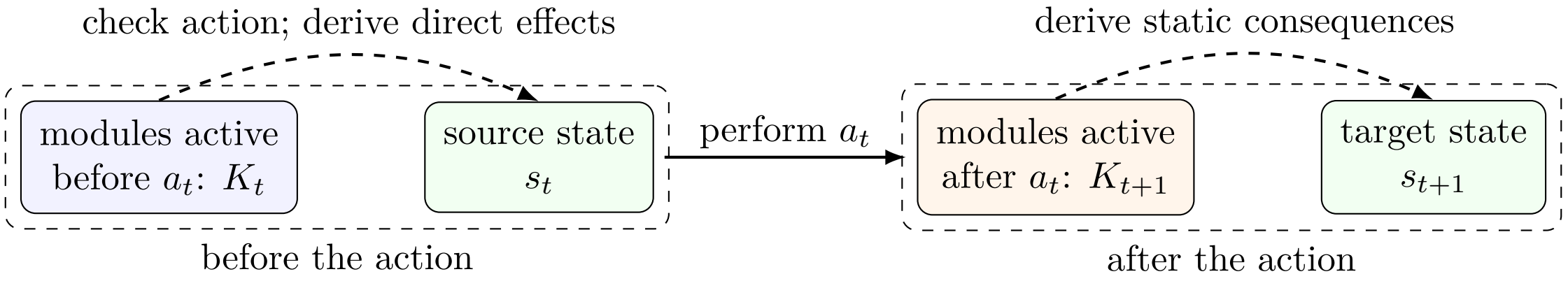

The central semantic issue is which laws govern a step that changes the
theory itself. Write \(\Dof K\) for the ordinary action description formed by
the core and the modules in \(K\) and \(\langle K, s\rangle\) for its configuration. At time \(t\), an action \(a_t\) labels a
transition
\[
  \langle K_t,s_t\rangle\trans{a_t}\langle K_{t+1},s_{t+1}\rangle.
\]
We call the state before the action its \emph{source state} and the state
after it its \emph{target state}.
We check executability using the nonexecutability laws of \(\Dof{K_t}\)
and the current module set \(K_t\). Dynamic laws also come from
\(\Dof{K_t}\). Their effects depend on conditions evaluated in
\(s_t\) and \(s_{t+1}\).
Loading or unloading
changes \(K_t\) to \(K_{t+1}\). The static laws of \(\Dof{K_{t+1}}\)
determine within-state causal consequences, jointly with effects, inertia,
and initialization. A newly loaded module
contributes no executability or direct-effect law to its loading step, but
its static laws already govern the resulting state and may cause indirect
changes to existing fluents (Figure~\ref{fig:coupled}).

For finite ground RAT models whose initialization
conditions are represented by propositional formulas, a proposed
transition can be checked in polynomial time.
Successor existence is NP-complete, while unbounded reachability is
PSPACE-complete. Thus finite module selection does not
raise these worst-case bounds above ordinary finite-state causal reasoning
and planning. For finite RAT systems, causal theories and Answer Set
Programming (ASP)~\cite{lifschitzTurner1999,ferrarisEtAl2012} give bounded
planning, diagnosis, and counterexample search.

RAT is a specification for reasoning about a harness, not a replacement
for its scheduler or tool runtime. A designer can use it to ask whether
a tool can be loaded, whether existing actions retain their behavior,
or whether a goal remains reachable after a module change. Reproducible
experiments compare finite RAT models with local executions of official
DeepSeek Harness\footnote{\url{https://www.deepseek.com/en/harness/}},
Pi\footnote{\url{https://pi.dev/}}, and
OpenHands~\cite{wangEtAl2025OpenHands} implementations. We also test
ASP-based configuration computation and planning on generated models,
measuring how running time and ground-program size grow with problem size.

\FloatBarrier

The main contributions are:
\begin{enumerate}[leftmargin=*,label=(\arabic*)]
  \item \emph{RAT and its properties.} We define how actions change both
    world state and the modules forming an action description.
    For actions other than loading and unloading, RAT has the ordinary
    causal action semantics of the active description. We give
    sufficient conditions for at most one successor, for preserving
    existing behavior when adding a module, including through a read-only
    shared interface, and for the same possible final configurations
    when loading two modules in either order. We also
    show why unloading and reloading need not restore the earlier state.
    Experiments compare RAT predictions with recorded tool calls and
    component changes in three harness implementations.
  \item \emph{Reasoning tasks and complexity.} We define transition checking,
    successor existence, possible and necessary goal satisfaction,
    execution of a supplied action sequence, and reachability. We establish
    their complexity for finite ground inputs with formula-represented
    initialization conditions.
  \item \emph{Causal and ASP representations.} For finite RAT systems,
    we give a bounded causal translation and prove that its explanations
    represent exactly the executions. An established causal-to-ASP
    translation then represents these executions as answer sets, for fixed
    action sequences or bounded action choice.
\end{enumerate}

Section~\ref{sec:background} fixes the logical foundations.
Sections~\ref{sec:language}--\ref{sec:compilation} give RAT's language,
semantics, properties, and translations.
Section~\ref{sec:evaluation} reports executable checks.
Section~\ref{sec:related} compares related work, including Cordis.
Section~\ref{sec:conclusion} discusses use in harnesses, limitations,
and further work. The separate Supplementary Material\footnote{The
experimental code, usage instructions, and Supplementary Material are
available at \artifacturl.\label{fn:artifact}} supplies expanded
proofs, implementation checks, and reproduction details.

\section{Preliminaries}
\label{sec:background}

We first define the finite ground language used in the paper. We then review nonmonotonic causal theories, action descriptions, and Answer Set Programming (ASP). In this paper, we use variables in schematic expressions to abbreviate finite sets of ground instances.

\subsection{Grounding convention}
\label{sec:grounding}

Let \(\mathcal U\) be a finite nonempty set of object constants. 
We use function-free schematic expressions, whose variables are meta-level placeholders. 
For an expression \(E\), let \(\mathsf{Var}(E)\) denote its variables. A grounding substitution
\(\theta:\,\mathsf{Var}(E)\to\mathcal U\) replaces each variable \(x\) by \(\theta(x)\).
We write \(E\theta\) for the expression obtained by applying this substitution to \(E\).
Define
\[
  \mathsf{gr}_{\mathcal U}(E)
  =\{E\theta\mid\theta:\mathsf{Var}(E)\to\mathcal U\},
  \qquad
  \mathsf{gr}_{\mathcal U}(S)=\bigcup_{E\in S}\mathsf{gr}_{\mathcal U}(E)
\]
for a finite set \(S\) of expressions.

A variable may be restricted to a designated subset \(D\subseteq\mathcal U\); in that case, only substitutions assigning it a constant from \(D\) are allowed. This restriction is meta-level and does not introduce a sort into the object language. 
All definitions below are over the resulting finite ground language.

\subsection{Action signatures, formulas, and states}
\label{sec:action-signatures}

A \emph{finite ground action signature}
\(\Sigma=\langle\mathcal F^b,\mathcal F^d,\mathcal A\rangle\) consists of
pairwise disjoint finite sets of \emph{basic fluents}, \emph{defined
fluents}, and \emph{actions}.
Intuitively, a basic fluent records a fact that can persist, such as whether
a tool has been registered. Basic fluents obey \emph{inertia}: their values
normally persist between states, but dynamic or static causal laws may
cause them to change. A defined fluent instead expresses a condition whose
value must be explained by static laws in each state, such as whether a
registered tool is currently ready for use. It has no inertia: a previous
value alone does not justify its current value. The word \emph{defined}
specifies where causal support must come from.

For convenience, write
\(\mathcal F^b(\Sigma)=\mathcal F^b\),
\(\mathcal F^d(\Sigma)=\mathcal F^d\),
\(\mathcal A(\Sigma)=\mathcal A\), and
\(\mathcal F(\Sigma)=\mathcal F^b\cup\mathcal F^d\),
omitting \(\Sigma\) when it is clear from context.

For a fluent set \(\mathcal F\subseteq\mathcal F(\Sigma)\), a \emph{fluent literal} is either \(f\) or \(\neg f\), where \(f\in\mathcal F\).
Let \(\mathsf{Lit}(\mathcal F)=\{f,\neg f\mid f\in\mathcal F\}\). 
For a fluent literal \(\ell\), let \(\mathsf{fl}(\ell)\) denote its underlying fluent, so \(\mathsf{fl}(f)=\mathsf{fl}(\neg f)=f\).

We treat fluents as propositional atoms. More generally, for any finite set
\(\sigma\) of propositional atoms, let \(\mathsf{Form}(\sigma)\) denote the
formulas generated by
\[
\varphi::=\top\mid\bot\mid p\mid\neg\varphi
\mid(\varphi\land\varphi)\mid(\varphi\lor\varphi),
\qquad p\in\sigma.
\]
Here \(\top\) and \(\bot\) denote truth and falsity. For
\(F,G\in\mathsf{Form}(\sigma)\), implication \(G\to F\) abbreviates
\(\neg G\lor F\), and equivalence \(G\leftrightarrow F\) abbreviates
\((G\to F)\land(F\to G)\). We use \(\mathsf{Lit}(\sigma)\) for literals
over any atom set \(\sigma\). A \emph{classical interpretation}
\(I\subseteq\sigma\) makes exactly the atoms in \(I\) true.
We write \(I\models\varphi\) when it satisfies \(\varphi\).
A \emph{state formula} is a formula in \(\mathsf{Form}(\mathcal F)\). For a state formula
\(\varphi\), \(\mathsf{Fl}(\varphi)\) denotes the set of fluents occurring in \(\varphi\), regardless of polarity.

A literal set \(X\subseteq\mathsf{Lit}(\mathcal F)\) is \emph{consistent} if it does not contain both \(f\) and \(\neg f\) for any \(f\in\mathcal F\). It is \emph{complete over \(\mathcal F\)} if it contains exactly one of \(f\) and \(\neg f\) for every \(f\in\mathcal F\).

Define
\[
\mathsf{St}(\mathcal F)
    =\{s\subseteq\mathsf{Lit}(\mathcal F)  \mid s \text{ is consistent and }
        s\text{ is complete over }\mathcal F\},
\]
a \emph{state} in $\mathsf{St}(\mathcal F)$ assigns every fluent in \(\mathcal F\).
For \(s\in\mathsf{St}(\mathcal F)\), let $I_s=\{f\in\mathcal F\mid f\in s\}$.
We write \(s\models\varphi\) for a state formula $\varphi\in \mathsf{Form}(\mathcal F)$ iff the classical interpretation \(I_s\) satisfies \(\varphi\).

\subsection{Nonmonotonic causal theories}
\label{sec:causal-foundation}

A causal theory specifies not only which formulas are true, but also what
causes them to be true~\cite{mccainTurner1997,giunchigliaEtAl2004}.
Fix a finite set \(\sigma\) of propositional atoms. A \emph{causal rule}
\(F\Leftarrow G\), where \(F,G\in\mathsf{Form}(\sigma)\), states that its
head \(F\) is caused when its body \(G\) holds. A finite set \(T\) of causal
rules is a \emph{causal theory}. Define its \emph{causal reduct} at
\(I\subseteq\sigma\) and its \emph{causally explained interpretations} by
\[
\begin{aligned}
  T^I&=\{F\mid(F\Leftarrow G)\in T,\ I\models G\},\\
  \mathsf{CE}(T)&=\{I\subseteq\sigma\mid
    I\text{ is the unique classical model of }T^I\}.
\end{aligned}
\]
The reduct contains the heads whose bodies are true in \(I\). They must
determine \(I\) uniquely over \(\sigma\); satisfying the implications
\(G\to F\) is insufficient. For \(\sigma=\{p\}\), the empty theory has
no explanation, but \(\{p\Leftarrow\top\}\) explains \(\{p\}\).
Adding rules can create or remove explanations, so the semantics is
nonmonotonic.

\subsection{Causal action descriptions}
\label{sec:action-theory-foundation}

An action description specifies constraints on states, effects of actions,
and conditions that forbid actions. Our baseline is the Boolean,
literal-headed, single-action fragment of \(\mathcal C{+}\), with its
distinction between simple and statically determined fluent constants
~\cite{giunchigliaEtAl2004}. We call these basic and defined fluents,
respectively, and declare every basic fluent inertial. Fix
\(\Sigma=\langle\mathcal F^b,\mathcal F^d,\mathcal A\rangle\) and
\(\mathcal F=\mathcal F^b\cup\mathcal F^d\).
For a step performing action \(a\), we call the state \(s\) before the
action its \emph{source state} and the state \(s'\) after it its
\emph{target state}.
A \emph{static causal law}
\[
  \mathsf{caused}\ h\ \mathsf{if}\ \psi,
  \qquad h\in\mathsf{Lit}(\mathcal F)\cup\{\bot\},\quad
  \psi\in\mathsf{Form}(\mathcal F),
\]
causes \(h\) when \(\psi\)
holds in the same state. If \(h=\bot\), it is a \emph{static constraint}
excluding such states. A \emph{dynamic causal law}
\[
  \mathsf{caused}\ \ell\ \mathsf{if}\ \psi\
  \mathsf{after}\ a\land\varphi,
  \qquad \ell\in\mathsf{Lit}(\mathcal F^b),\quad a\in\mathcal A,\quad
  \varphi,\psi\in\mathsf{Form}(\mathcal F),
\]
causes \(\ell\) in the target state \(s'\), provided the action is \(a\),
\(\varphi\) holds in \(s\), and \(\psi\) holds in \(s'\).
The abbreviation \(a\ \mathsf{causes}\ \ell\ \mathsf{if}\ \varphi\)
sets \(\psi=\top\); omitting the condition also sets \(\varphi=\top\).
A \emph{nonexecutability law}
\[
  \mathsf{nonexecutable}\ a\ \mathsf{if}\ \varphi,
  \qquad a\in\mathcal A,\quad\varphi\in\mathsf{Form}(\mathcal F),
\]
forbids performing \(a\) in any state satisfying \(\varphi\).

An \emph{action description}
\(\mathcal D=\langle\Sigma,L^s,L^d,L^x\rangle\) consists of the signature
and finite sets \(L^s,L^d,L^x\) of static causal laws, dynamic causal laws,
and nonexecutability laws, respectively. For \(a\in\mathcal A\) and
\(s\in\mathsf{St}(\mathcal F)\), put
\[
\begin{aligned}
  \mathsf{Blocked}_{\mathcal D}(a,s)
    &\ \text{if{f}}\
      \exists\varphi:\,
      (\mathsf{nonexecutable}\ a\ \mathsf{if}\ \varphi)\in L^x
      \ \land\ s\models\varphi,\\
  \mathsf{Enabled}_{\mathcal D}(s)
    &=\{a\in\mathcal A\mid\neg\mathsf{Blocked}_{\mathcal D}(a,s)\}.
\end{aligned}
\]
In this paper, exactly one ground action occurs at each step.
We do not use the full language \(\mathcal C{+}\):
our law heads are literals, except for \(\bot\) in constraints, and
we do not allow concurrent action occurrences.

\paragraph{Causal theories for states and transitions}
We define the state and transition semantics by causal theories, using
\(\mathsf{CE}\) from Section~\ref{sec:causal-foundation}. For an action
description \(\mathcal D=\langle\Sigma,L^s,L^d,L^x\rangle\), one theory
specifies which states are possible, while another specifies the possible target
states for a given source and action. Both are over the fluent atoms
\(\mathcal F\).

For a complete basic-fluent assignment
\(X\in\mathsf{St}(\mathcal F^b)\), define the \emph{state theory}
\begin{equation}
  \mathsf{CT}_{\mathcal D}(X)
  =\{\ell\Leftarrow\top\mid\ell\in X\}
   \cup\{h\Leftarrow\psi\mid
     (\mathsf{caused}\ h\ \mathsf{if}\ \psi)\in L^s\}.
  \label{eq:ordinary-state-theory}
\end{equation}
A complete state \(s\in\mathsf{St}(\mathcal F)\) is \emph{causally
admissible for \(\mathcal D\)} iff
\[
  I_s\in\mathsf{CE}\bigl(
    \mathsf{CT}_{\mathcal D}(s\cap\mathsf{Lit}(\mathcal F^b))\bigr).
\]
The first set of rules fixes the basic values of this candidate state,
while the static laws must explain its defined-fluent values. All fluent
values must also satisfy those laws. Fixing basic values here specifies
which individual states are admissible.

For a causally admissible source \(s\) and an action \(a\in\mathcal A\),
define the \emph{single-step causal theory}
\begin{equation}
\begin{split}
  T_{\mathcal D,s,a}={}&
    \{h\Leftarrow\psi\mid
      (\mathsf{caused}\ h\ \mathsf{if}\ \psi)\in L^s\}\\
  &\cup\{\ell\Leftarrow\psi\mid
    (\mathsf{caused}\ \ell\ \mathsf{if}\ \psi\
      \mathsf{after}\ a\land\varphi)\in L^d,\ s\models\varphi\}\\
  &\cup\{\ell\Leftarrow\ell\mid
    \ell\in s\cap\mathsf{Lit}(\mathcal F^b)\}.
\end{split}
  \label{eq:ordinary-step-causal-theory}
\end{equation}
Its three sets of rules represent static laws, dynamic laws for \(a\)
whose formula \(\varphi\) holds in \(s\), and inertia, respectively.
The remaining rule bodies are evaluated in the candidate target state.
Unlike the facts in \(\mathsf{CT}_{\mathcal D}(X)\), the inertia rules
in \(T_{\mathcal D,s,a}\) support source basic literals only if they
remain true in the target. Static or dynamic laws may support a changed
value.

For causally admissible states \(s,s'\), the \emph{ordinary transition
relation} is
\begin{equation}
  s\trans{a}_{\mathcal D}s'
  \quad\text{if{f}}\quad
  a\in\mathsf{Enabled}_{\mathcal D}(s)
  \ \text{and}\ 
  I_{s'}\in\mathsf{CE}(T_{\mathcal D,s,a}).
  \label{eq:ordinary-transition}
\end{equation}
Thus \(a\) must not be blocked in \(s\), and
\(T_{\mathcal D,s,a}\) must causally explain \(s'\).
This is the fixed-source form of the standard \(\mathcal C{+}\) causal
semantics~\cite[Sections~4.2 and~4.4]{giunchigliaEtAl2004}.

\subsection{Answer Set Programming}
\label{sec:asp-foundation}

ASP is a form of nonmonotonic logic programming for knowledge
representation and declarative problem solving, supported by efficient
answer set solvers~\cite{brewkaEiterTruszczynski2011}. Its languages allow
rules, constraints, and defaults to describe combinatorial search
problems. We first recall ground normal rules and their answer set
semantics~\cite{gelfondLifschitz1988,gelfondLifschitz1991}.

Fix a finite set \(\sigma\) of propositional atoms. A \emph{ground normal
rule} over \(\sigma\) has the form
\[
  h\leftarrow\ell_1,\ldots,\ell_m,
  \mathsf{not}\ \ell_{m+1},\ldots,\mathsf{not}\ \ell_n,
  \qquad 0\leq m\leq n.
\]
Here \(h\in\mathsf{Lit}(\sigma)\cup\{\bot\}\) is the \emph{head}, and
the expressions to the right of \(\leftarrow\) form the \emph{body}.
Each body literal \(\ell_i\in\mathsf{Lit}(\sigma)\), for \(1\leq i\leq n\),
is either an atom \(p\) or its
\emph{classical negation} \(\neg p\), for \(p\in\sigma\).
Thus classical negation is allowed in both the head and the body;
\(\mathsf{not}\) denotes a different operation, \emph{default negation}.
A rule with \(h=\bot\) is a \emph{constraint}, usually written without
a head. A finite set \(\Pi\) of such rules is a \emph{ground normal
program}.

An \emph{ASP interpretation} is a consistent set
\(X\subseteq\mathsf{Lit}(\sigma)\). For an atom \(p\in\sigma\), it may
omit both \(p\) and \(\neg p\); unlike a state, it need not assign a
truth value to every atom. For \(\ell\in\mathsf{Lit}(\sigma)\), define
\[
  X\models\ell\ \text{if{f}}\ \ell\in X,
  \qquad
  X\models\mathsf{not}\ \ell\ \text{if{f}}\ \ell\notin X,
  \qquad X\not\models\bot.
\]
In particular, \(X\models\neg p\) requires \(\neg p\in X\), whereas
\(X\models\mathsf{not}\ p\) requires only \(p\notin X\).
For a rule \(r\) of the displayed form, \(X\) \emph{satisfies} \(r\),
written \(X\models r\), iff
\[
  \{\ell_1,\ldots,\ell_m\}\subseteq X
    \ \text{and} \
    \{\ell_{m+1},\ldots,\ell_n\}\cap X=\emptyset\ \text{ implies }\
   X\models h.
\]
Thus a rule with a literal head requires that literal whenever its body
is true. A constraint is satisfied exactly when its body is false.
An empty body is true. An interpretation is a \emph{model} of a program
\(\Pi\), written \(X\models\Pi\), iff \(X\models r\) for every
\(r\in\Pi\).

For an ASP interpretation \(X\), the \emph{Gelfond--Lifschitz reduct}
\(\Pi^X\) is obtained by deleting every rule whose body contains
\(\mathsf{not}\ \ell\) with \(\ell\in X\), and then removing all
remaining default-negated literals from rule bodies. The interpretation
\(X\) is an \emph{answer set} of \(\Pi\) iff it is a model of
\(\Pi^X\) and no proper subset of \(X\) is a model of \(\Pi^X\).
In other words, \(X\) must be a subset-minimal model of its own reduct.
We consider only consistent answer sets and write
\[
  \mathsf{AS}(\Pi)=
  \{X\subseteq\mathsf{Lit}(\sigma) \mid
    X\text{ is consistent},\ X\models\Pi^X,\ \text{and }
   \nexists Y\subset X:\ Y\models\Pi^X\}.
\]
Satisfaction alone is not enough: for example, both \(\emptyset\) and
\(\{p\}\) satisfy \(p\leftarrow p\). The reduct is the same rule for
either interpretation, and \(\emptyset\) is its only subset-minimal
model. Thus this program has only the answer set \(\emptyset\).

\paragraph{Rules with nested bodies}
The translation in Section~\ref{sec:rat-asp-translation} uses Boolean
bodies and repeated default negation. These go beyond normal-rule syntax.
We therefore also recall the standard extension to \emph{nested
expressions}~\cite{lifschitzTangTurner1999}, generated by
\[
  B::=\ell\mid\top\mid\bot\mid(B\land B)\mid(B\lor B)
       \mid\mathsf{not}\ B,
  \qquad \ell\in\mathsf{Lit}(\sigma).
\]
A rule \(h\leftarrow B\) keeps the head \(h\in \mathsf{Lit}(\sigma)\cup\{\bot\}\). A finite
set of these rules is a \emph{program with nested bodies}.
An ASP interpretation \(X\) evaluates literals as above, \(\top\) as true,
and conjunction and disjunction in the usual way. It satisfies
\(\mathsf{not}\ B\) iff it does not satisfy \(B\), and satisfies the
rule iff \(X\models B\) implies \(X\models h\).

The reduct replaces each occurrence of \(\mathsf{not}\ B\) that is not
inside another default negation by \(\bot\) if \(X\models B\), and by
\(\top\) otherwise. Answer sets are again the consistent subset-minimal
models of their own reducts. We still use \(\mathsf{AS}(\Pi)\) to denote the set of answer sets of a program $\Pi$ with nested expressions.
This definition agrees with the preceding one on normal
programs. In particular,
\[
  {(\mathsf{not}\,\mathsf{not}\ B)}^X
  =\begin{cases}
     \top,&X\models B,\\
     \bot,&X\not\models B.
   \end{cases}
\]
Double default negation cannot in general be removed. For example,
\(p\leftarrow\mathsf{not}\,\mathsf{not}\ p\) has answer sets
\(\emptyset\) and \(\{p\}\), whereas \(p\leftarrow p\) has only
\(\emptyset\).

ASP computation typically involves two stages: grounding and solving.
A grounder, such as \texttt{gringo}~\cite{gebserSchaubThiele2007gringo}, converts a finite-domain non-ground program into a ground program.
A solver, such as \texttt{clasp}~\cite{gebserEtAl2007clasp}, computes answer sets of the resulting ground program.
Literal-headed causal theories can be translated into ASP to compute
their causally explained interpretations~\cite{ferrarisEtAl2012}. Furthermore,
Ji and Lin~\cite{jiLin2013} study related translations of Turner's logic
of universal causation into propositional logic and logic programs.

\section{Reconfigurable Action Theories}
\label{sec:language}

In RAT, actions can load or unload modules of an action description.
We define theory modules and configurations, and specify which action
description each configuration uses. Section~\ref{sec:semantics} defines
their transitions.

\subsection{Motivating example}
\label{sec:calculator-motivation}

Consider a ReAct-style agent~\cite{yao2023react} asked to compute
\(23\times17\). It initially has no calculator, so the harness first
loads one. After loading, the harness
registers the calculator so that the agent can call it. A multiplication
call then stores the answer \(391\), which can later be cleared. The
harness can also unload the calculator, after which the agent can no
longer use its actions. In this example, multiplication is allowed only while the
calculator is registered and the task has not finished.

Loading changes what the agent can do: it adds calculator actions and the
rules that describe when they can be used and what they do. These actions
and rules were absent before loading. Unloading removes them again, along
with the calculator's registration and answer state. A fixed action
description can simulate this behavior using predeclared availability
flags; RAT instead represents the calculator as a module that is added to
or removed from the current action description.

\subsection{Theory modules and RAT systems}
\label{sec:rat-core-language}

A module contains an action description and an initializer. Its vocabulary
and laws become active together. Actions select from a fixed finite module
set without changing their laws.

Let \(\Modules\subseteq\mathcal U\) be a finite set of optional module
identifiers represented by object constants. Meta-variables \(M,N\) range
over \(\Modules\)\footnote{All schematic expressions below stand for their finite
ground instances, as in Section~\ref{sec:grounding}.}.  The always-present core
\(\mathcal D_{\mathsf{core}}\), with \(\mathsf{core}\notin\Modules\), and
optional \(\mathcal D_M=\langle\Sigma_M,L_M^s,L_M^d,L_M^x\rangle\) are
ordinary descriptions, each declaring all symbols used in its laws.
Two signatures are \emph{compatible} iff every shared symbol has the same
category in both: basic fluent, defined fluent, or action. We require
pairwise compatibility and take signature unions componentwise.
Define the \emph{ambient
signature} and each module's \emph{private basic-fluent set} by
\[
  \Sigma^*
  =\Sigma_{\mathsf{core}}\cup\bigcup_{M\in\Modules}\Sigma_M,\qquad
  \mathcal F^b_M
  =\mathcal F^b(\Sigma_M)\setminus
    \mathcal F^b(\Sigma_{\mathsf{core}}).
\]
We require \(\mathcal F^b_M\cap\mathcal F^b_N=\emptyset\) for all distinct
\(M,N\in\Modules\).
Thus each noncore basic fluent has one declaring module, whose initializer
supplies its values on loading. This ownership condition keeps the
responsibility for initialization unambiguous. Core fluents remain visible;
compatible actions and defined fluents may be shared. A defined fluent
needs no initializing owner: its value requires support from the active
static laws. 

Loading a module introduces its private basic fluents into the state.
These fluents have no values in the preceding state, so inertia cannot
supply their initial values. An \emph{initializer} specifies the values
they may have immediately after loading, possibly depending on the current
state. For example, a newly loaded calculator may start unregistered and
without a stored answer.

For a fluent set \(V\), write \(\mathsf{PVal}(V)\) for the set of
consistent subsets of \(\mathsf{Lit}(V)\). Such a \emph{partial valuation}
may leave some fluents unassigned.

\begin{definition}[Theory module]
\label{def:module}
For a module identifier \(M\in\Modules\), a RAT \emph{theory module} is
\(\mathsf{Mod}(M)=\langle\mathcal D_M,\mathsf{Init}_M\rangle\), with ordinary
description \(\mathcal D_M = \langle\Sigma_M,L_M^s,L_M^d,L_M^x\rangle\) and an \emph{initializer}
\[
  \mathsf{Init}_M:
  \mathsf{PVal}(\mathcal F(\Sigma^*))
  \longrightarrow
  2^{\mathsf{St}(\mathcal F^b_M)}.
\]
\end{definition}

The input \(v\in\mathsf{PVal}(\mathcal F(\Sigma^*))\) records the
fluent values available before loading. The output \(\mathsf{Init}_M(v)\)
is a set of permitted initial assignments. Each member
\(Y\in\mathsf{Init}_M(v)\) is a complete assignment to \(\mathcal F^b_M\).
For every basic fluent \(f\in\mathcal F^b_M\), it contains exactly one
of \(f\) and \(\neg f\).
A loading successor must be causally explained using one such assignment
together with the applicable laws (Section~\ref{sec:semantics}).
The selected assignment fixes the new basic values; the laws can reject it,
but cannot replace those values. We allow several assignments because the
same loading action from the same source state may initialize a module
in different ways. Initialization may depend on external conditions,
such as service availability, that are not represented in the source state.

As an example, consider loading a connector module \(M\) that
attempts to connect to an external service. The connection attempt may
succeed or fail without changing whether the connector itself is installed.
If its outcome is not determined by the represented source state, we model
both possibilities using the private basic fluent \(\mathit{connected}(M)\):
\[
\begin{gathered}
  \mathcal F^b_M=\{\mathit{connected}(M)\},\\
  \mathsf{Init}_M(v)
    =\bigl\{\{\mathit{connected}(M)\},\{\neg\mathit{connected}(M)\}\bigr\}
    \quad\text{for every input }v.
\end{gathered}
\]
Let the core basic fluent \(\mathit{mustConnect}\) record whether loading
requires a successful connection. The module declares this shared fluent
and has the static constraint
\[
  \mathsf{caused}\ \bot\ \mathsf{if}\
    \mathit{mustConnect}\land\neg\mathit{connected}(M).
\]
Take no other fluents or laws. Inertia preserves \(\mathit{mustConnect}\).
When it is false, loading has two successors, one connected and one
disconnected; when it is true, only the connected successor is allowed.

\begin{definition}[RAT system and module actions]
\label{def:rat-system}
A finite RAT system is
\[
  \mathcal R=\left\langle
    \Modules,\mathcal D_{\mathsf{core}},
    {\{\mathsf{Mod}(M)\}}_{M\in\Modules}\right\rangle.
\]
For every module identifier \(M\in\Modules\), it reserves two ground action
symbols, \(\mathit{load}(M)\) and \(\mathit{unload}(M)\), with
\[
  \mathcal A^{\mathsf r}
  =\{\mathit{load}(M),\mathit{unload}(M)\mid M\in\Modules\}
  \subseteq\mathcal A(\Sigma_{\mathsf{core}}).
\]
Distinct identifiers give distinct loading actions and distinct unloading
actions, and no loading action equals an unloading action.
\end{definition}

Ordinary dynamic and nonexecutability laws specify these actions' world
effects and when they are blocked. RAT also specifies how they change the
active module set
(Section~\ref{sec:module-update-semantics}). 

\subsection{Active action descriptions}
\label{sec:active-descriptions}

Let \(K\subseteq\Modules\) be the set of active optional modules. The
current signature and law sets are their unions with those of the core:
\[\begin{aligned}
  \Sigma[K]&=\Sigma_{\mathsf{core}}\cup\bigcup_{M\in K}\Sigma_M,\\
  L^z[K]&=L^z_{\mathsf{core}}\cup\bigcup_{M\in K}L^z_M,
  \qquad z\in\{s,d,x\}.
\end{aligned}\]
The \emph{active action description} is
\(\Dof{K}=\langle\Sigma[K],L^s[K],L^d[K],L^x[K]\rangle\), with
\(\mathcal F^b[K]=\mathcal F^b(\Sigma[K])\),
\(\mathcal F^d[K]=\mathcal F^d(\Sigma[K])\),
\(\mathcal F[K]=\mathcal F(\Sigma[K])\), and
\(\mathcal A[K]=\mathcal A(\Sigma[K])\).

Exactly the modules in \(K\) are \emph{active}; \emph{inactive} modules
contribute neither signatures nor laws. A symbol disappears only when no
active fragment declares it.

\subsection{RAT configurations}
\label{sec:rat-configurations}

A world state records fluent values. RAT also records the active module
set, which determines the current action description.

\begin{definition}[RAT configuration]
\label{def:rat-configuration}
A RAT \emph{configuration} is
\[
  \Config=\langle K,s\rangle,
  \qquad
  K\subseteq\Modules,\quad s\in\mathsf{St}(\mathcal F[K]).
\]
\end{definition}

The ground actions \(a\in\mathcal A[K]\) are those present in the active
description. A present action may still be blocked or have no successor.
This definition specifies the components of a configuration;
Section~\ref{sec:semantics} determines which states satisfy the causal
semantics. RAT records only the active modules and the world state.

\subsection{Formalization of the running example}
\label{sec:running-example}

We formalize the running example in Section~\ref{sec:calculator-motivation}.
We specify the core first, then the calculator. The only optional module
identifier is the object constant \(\mathit{cal}\), so \(\Modules=\{\mathit{cal}\}\).
The core basic fluent \(\mathit{taskDone}\) records task completion.
The action \(\mathit{finishTask}\) marks the task complete; \(\mathit{load}(\mathit{cal})\)
and \(\mathit{unload}(\mathit{cal})\) add and remove the calculator. Its signature is
\[
  \Sigma_{\mathsf{core}}
    =\langle\{\mathit{taskDone}\},\emptyset,
       \{\mathit{load}(\mathit{cal}),\mathit{unload}(\mathit{cal}),\mathit{finishTask}\}\rangle.
\]
The core has the single dynamic law
\(\mathit{finishTask}\ \mathsf{causes}\ \mathit{taskDone}\); its static and nonexecutability
law sets are empty. Initially, the task is not finished and the
calculator is not loaded:
\[
  K_0=\emptyset,\qquad
  \Config_0=\langle K_0,\{\neg\mathit{taskDone}\}\rangle,\qquad
  \Sigma[K_0]=\Sigma_{\mathsf{core}}.
\]

The calculator's private basic fluents are \(\mathit{registered}(\mathit{cal})\)
and \(\mathit{answer}(391)\): they record whether the calculator is
registered and whether the answer \(391\) is stored. Its defined fluent
\(\mathit{ready}(\mathit{cal})\) records whether multiplication is allowed. Its three actions are
\(\mathit{register}(\mathit{cal})\), \(\mathit{multiply}(23,17)\), and
\(\mathit{clearAnswer}(391)\). All terms are ground; the numeral arguments
are object constants. Its signature is
\begin{multline*}
  \Sigma_{\mathit{cal}}=\langle\,\{\mathit{taskDone},\mathit{registered}(\mathit{cal}),\mathit{answer}(391)\},\,
       \{\mathit{ready}(\mathit{cal})\},\\
    \{\mathit{register}(\mathit{cal}),\mathit{multiply}(23,17),
       \mathit{clearAnswer}(391)\}\,\rangle.
\end{multline*}
Registration and the stored answer persist until changed. Readiness must
instead reflect the current registration and task status; it must not
remain true merely because it was true before the task finished.
The calculator reads \(\mathit{taskDone}\) but never changes it. In the notation of
Section~\ref{sec:action-theory-foundation}, the calculator's static laws are
\[\begin{gathered}
  \mathsf{caused}\ \mathit{ready}(\mathit{cal})\ \mathsf{if}\
    \mathit{registered}(\mathit{cal})\land\neg\mathit{taskDone},\\
  \mathsf{caused}\ \neg\mathit{ready}(\mathit{cal})\ \mathsf{if}\
    \neg\mathit{registered}(\mathit{cal})\lor\mathit{taskDone}.
\end{gathered}\]
Its dynamic laws and nonexecutability law are
\[\begin{gathered}
  \mathit{register}(\mathit{cal})\ \mathsf{causes}\ \mathit{registered}(\mathit{cal}),\\
  \mathit{multiply}(23,17)\ \mathsf{causes}\ \mathit{answer}(391),\\
  \mathit{clearAnswer}(391)\ \mathsf{causes}\ \neg\mathit{answer}(391),\\
  \mathsf{nonexecutable}\ \mathit{multiply}(23,17)\ \mathsf{if}\ \neg\mathit{ready}(\mathit{cal}).
\end{gathered}\]
These are exactly the laws in \(L_{\mathit{cal}}^s,L_{\mathit{cal}}^d,L_{\mathit{cal}}^x\), respectively.

Loading the calculator starts it unregistered and without a stored answer.
Its initializer therefore returns the same singleton set for every input
\(v\in\mathsf{PVal}(\mathcal F(\Sigma^*))\):
\[
  \mathsf{Init}_{\mathit{cal}}(v)
    =\bigl\{\{\neg\mathit{registered}(\mathit{cal}),\neg\mathit{answer}(391)\}\bigr\}.
\]
The initializer assigns neither \(\mathit{taskDone}\),
which belongs to the core, nor \(\mathit{ready}(\mathit{cal})\), which is defined.
The second static law makes \(\mathit{ready}(\mathit{cal})\) false because the
calculator is initially unregistered. This completes the RAT system
\(
  \mathcal R_{\times}
  =\left\langle\{\mathit{cal}\},\mathcal D_{\mathsf{core}},
    \{\mathsf{Mod}(\mathit{cal})\}\right\rangle
\).

Before loading, the calculator's three fluents and its multiplication
action are absent:
\[
\begin{gathered}
  \{\mathit{registered}(\mathit{cal}),\mathit{answer}(391),\mathit{ready}(\mathit{cal})\}
    \cap\mathcal F[K_0]=\emptyset,\\
  \mathit{multiply}(23,17)\notin\mathcal A[K_0].
\end{gathered}
\]
With the calculator active, the selection \(K_1=\{\mathit{cal}\}\) instead gives
\[
  \Sigma[K_1]=\Sigma_{\mathsf{core}}\cup\Sigma_{\mathit{cal}},\qquad
  \Dof{K_1}=\langle\Sigma[K_1],L_{\mathit{cal}}^s,
    L^d_{\mathsf{core}}\cup L_{\mathit{cal}}^d,L_{\mathit{cal}}^x\rangle.
\]
Section~\ref{sec:running-transitions} derives the transitions of this system.

\section{RAT transition semantics}
\label{sec:semantics}

We now define when an action \(a\) takes a configuration
\(\langle K,s\rangle\) to \(\langle K',s'\rangle\).
We first specify which states satisfy the static causal laws of the core
and active modules. We then specify which actions are enabled, how
\(K\) changes to \(K'\), and which fluent values can hold after the
action. The definition uses the nonmonotonic causal semantics of
Section~\ref{sec:causal-foundation}. When loading introduces new basic
fluents, their initial values come from the module's initializer.

\subsection{Static causal closure and admissibility}
\label{sec:static-closure}

A complete assignment of fluent values need not be a possible state of
the system. It must satisfy the static laws, and the truth value of each
defined fluent must have causal support from those laws.
Let \(K\subseteq\Modules\) be the active module set in a RAT system
\(\mathcal R\).
The description \(\Dof K\) has signature \(\Sigma[K]\), with basic and
defined fluent sets \(\mathcal F^b[K]\) and \(\mathcal F^d[K]\).
For \(s\in\mathsf{St}(\mathcal F[K])\), its \emph{basic projection}
\(\mathsf{Base}_K(s)=s\cap\mathsf{Lit}(\mathcal F^b[K])\) keeps only
the basic literals.

We treat the values of basic fluents as given when imposing these
conditions on a state. For a complete assignment
\(X\in\mathsf{St}(\mathcal F^b[K])\), define the following causal theory
over \(\mathcal F[K]\):
\begin{equation}
  \mathsf{CT}_K(X)
  =\{\ell\Leftarrow\top\mid\ell\in X\}
   \cup
   \{h\Leftarrow\psi\mid
     (\mathsf{caused}\ h\ \mathsf{if}\ \psi)\in L^s[K]\}.
  \label{eq:static-causal-theory}
\end{equation}
The first set fixes the basic-fluent values to \(X\); the second
translates every static law of \(\Dof K\) into a causal rule.
Using the classical interpretation \(I_s\) for the state \(s\) from
Section~\ref{sec:action-signatures}, define the \emph{causal static closure}
\begin{equation}
  \cl_K(X)=
  \{s\in\mathsf{St}(\mathcal F[K])\mid
    \mathsf{Base}_K(s)=X\text{ and }
    I_s\in\mathsf{CE}(\mathsf{CT}_K(X))\}.
  \label{eq:causal-static-closure}
\end{equation}

Thus \(\cl_K(X)\) contains the complete states causally explained by
\(\mathsf{CT}_K(X)\): each \(I_s\) must be the unique model of its own
causal reduct. The set can be empty, contain one state, or contain several
states. A state \(s\) is \emph{causally admissible under
\(K\)} iff \(s\in\cl_K(\mathsf{Base}_K(s))\); we then call the
configuration \(\langle K,s\rangle\) causally admissible as well.


\subsection{Executability and module change}
\label{sec:module-update-semantics}

An action is enabled only if no nonexecutability law of \(\Dof K\) blocks it.
Loading an already active module and unloading an inactive one are also
excluded. For a causally admissible configuration \(\langle K,s\rangle\),
with \(s\) over \(\Sigma[K]\), define the set of \emph{enabled actions} by
\begin{multline}
  \enabled_K(s)=\{a\in\enabled_{\Dof K}(s)\mid{}
    \forall M\in\Modules:\\
   (a=\mathit{load}(M)\text{ implies }M\notin K)\ \text{and }
   (a=\mathit{unload}(M)\text{ implies }M\in K)\}.
  \label{eq:enabled}
\end{multline}
Thus RAT uses the ordinary test \(\enabled_{\Dof K}\) from
Section~\ref{sec:action-theory-foundation} and also checks the active status
of the module being loaded or unloaded. Passing these tests does not
guarantee a transition: the causal laws may allow no target state.

The action label determines the next module set. For
\(K\subseteq\Modules\) and \(a\in\mathcal A[K]\), define
\begin{equation}
  \mathsf{Next}_K(a)=
  \begin{cases}
    K\cup\{M\},&
      \text{if }a=\mathit{load}(M),\\
    K\setminus\{M\},&
      \text{if }a=\mathit{unload}(M),\\
    K,&
      \text{if }a\notin\mathcal A^{\mathsf r}.
  \end{cases}
  \label{eq:module-successor}
\end{equation}

Loading and unloading labels can occupy the action position in dynamic
and nonexecutability laws, just like any other ground action. For example,
fix \(M\in\Modules\) and let the core basic fluents
\(\mathit{canLoad}(M)\) and \(\mathit{reserved}(M)\) record permission to
load \(M\) and whether a resource is reserved for it. The core may contain
\[
\begin{gathered}
  \mathit{load}(M)\ \mathsf{causes}\ \mathit{reserved}(M),\\
  \mathsf{nonexecutable}\ \mathit{load}(M)\ \mathsf{if}\
    \neg\mathit{canLoad}(M),\\
  \mathit{unload}(M)\ \mathsf{causes}\ \neg\mathit{reserved}(M).
\end{gathered}
\]
Thus loading requires permission and causes a reservation, while unloading
releases it. 

\subsection{Initialization and the single-step causal theory}

Let \(\langle K,s\rangle\) be a source configuration of the RAT system
\(\mathcal R\), with active module set \(K\) and state \(s\).
For a ground action \(a\in\mathcal A[K]\), let
\(K'=\mathsf{Next}_K(a)\) be the target active module set.
We construct a causal theory over \(\mathcal F[K']\) whose explanations
give the possible target states \(s'\). It describes one action from
the fixed source, rather than a sequence of actions; we therefore call
it a \emph{single-step causal theory}.

\paragraph{Initialization}
For each newly active module, choose an assignment allowed by its
initializer. The set of combined assignments is
\[
  \mathsf{Init}_{K\rightarrow K'}(s)=
  \left\{\bigcup_{M\in K'\setminus K}Y_M\ \middle|\;
    Y_M\in\mathsf{Init}_M(s)
    \text{ for every }M\in K'\setminus K\right\}.
\]
Each assignment \(Y_M\) gives the initial values of module \(M\)'s
private basic fluents. These fluent sets are disjoint, so the union of
the assignments is consistent. If no module is new,
\(\mathsf{Init}_{K\rightarrow K'}(s)=\{\emptyset\}\).
One RAT step loads at most one module; the set-based form is also useful
when discussing several module changes in the later proofs.

\paragraph{Selecting the laws}
Loading or unloading can change both the laws and the fluents in the
action description. We therefore need to specify which laws describe
the action and which laws describe the resulting state.
The target state uses all static laws of \(\Dof{K'}\), giving
\[
  \mathsf{Static}(K')
    =\{h\Leftarrow\psi\mid
      (\mathsf{caused}\ h\ \mathsf{if}\ \psi)\in L^s[K']\}.
\]
The action's dynamic laws come from \(\Dof K\), the description in
force before the action. In particular, a newly loaded module supplies
static laws for the target state, but no dynamic law for its own loading.
Likewise, its nonexecutability laws cannot block its own loading, but
can block its unloading while the module is active.
For the selected action \(a\), consider a dynamic law
\[
  \mathsf{caused}\ \ell\ \mathsf{if}\ \psi\
  \mathsf{after}\ a\land\varphi.
\]
The formula \(\varphi\) is tested in the state \(s\) before the action,
while \(\psi\) is tested in the candidate state \(s'\) after the action.
The law can cause \(\ell\)
in \(s'\) only when both conditions hold.

Unloading may remove fluents mentioned in the law. We call a dynamic law
\emph{target applicable} if the fluent in its head and all fluents written
in \(\psi\) belong to \(\mathcal F[K']\). This tests whether the law
can be evaluated there, not whether \(\psi\) is true. The formula \(\varphi\)
may still refer to a removed fluent, since its value is available in \(s\).
No such additional test is needed for the static laws above: their
symbols already belong to \(\Sigma[K']\).
The selected dynamic laws give
\begin{multline*}
  \mathsf{Dynamic}(K,s,a,K')
    =\{\ell\Leftarrow\psi\mid
      (\mathsf{caused}\ \ell\ \mathsf{if}\ \psi\
        \mathsf{after}\ a\land\varphi)\in L^d[K],\ s\models\varphi,\\
    \mathsf{Fl}(\psi)\subseteq\mathcal F[K'],\
      \mathsf{fl}(\ell)\in\mathcal F^b[K']\}.
\end{multline*}
\paragraph{Combining the causes}
Let \(S=\mathcal F^b[K]\cap\mathcal F^b[K']\) be the set of all basic
fluents present before and after the action. For an allowed initialization
assignment \(Y\in\mathsf{Init}_{K\rightarrow K'}(s)\), the single-step causal theory
is
\begin{equation}
  T_{K,s,a,K',Y}=\mathsf{Static}(K')
    \cup\mathsf{Dynamic}(K,s,a,K')
   \cup\{\ell\Leftarrow\ell\mid\ell\in s\cap\mathsf{Lit}(S)\}
    \cup\{\ell\Leftarrow\top\mid\ell\in Y\}.
  \label{eq:step-causal-theory}
\end{equation}
These are exactly four families of causal rules:
\begin{enumerate}[leftmargin=*,label=(\roman*)]
  \item \emph{Static rules.} Each static law of \(\Dof{K'}\) contributes
    \(h\Leftarrow\psi\). Its body is evaluated in the candidate target
    state \(s'\).
  \item \emph{Dynamic rules.} Each target-applicable dynamic law for
    \(a\) in \(\Dof K\) with \(s\models\varphi\) contributes
    \(\ell\Leftarrow\psi\). The formula \(\varphi\) has already
    been tested in \(s\), while \(\psi\) remains in the rule body
    to be tested in \(s'\).
  \item \emph{Inertia rules.} For every basic fluent \(f\in S\), take
    its literal \(\ell\) true in \(s\) and include
    \(\ell\Leftarrow\ell\). This includes all such fluents, even when
    a static or dynamic law may change their values. If \(\ell\) remains
    true in \(s'\), the rule supplies a cause for it. If \(\ell\) is
    false in \(s'\), the rule's body is false and contributes nothing
    to the causal reduct; the opposite value must have another cause.
  \item \emph{Initialization rules.} For every literal \(\ell\in Y\),
    include \(\ell\Leftarrow\top\). These facts assign the new basic
    fluents, which had no values in \(s\) to preserve by inertia.
    Incompatible laws reject this assignment rather than replace it.
\end{enumerate}
Every fluent in \(\mathcal F^b[K']\) is either in \(S\) or newly introduced,
so items (iii) and (iv) cover all of them. Fluents removed by the action
do not belong to \(\mathcal F[K']\). Defined fluents have no inertia rules;
their values must be supported by the static rules in item (i).
All four families are evaluated together by causal explanation, without
an order of application or a priority between them.

\paragraph{References to removed fluents}
Target applicability checks the formula as written, not an equivalent
simplification. The written fluent set is part of the law's interface.
Removing a fluent needed in its head or after-state condition disables
that effect rule, rather than blocking the whole action.
For example, suppose unloading removes private basic
\(q\) but retains basic \(p\), and the source is \(\{\neg p,\neg q\}\).
A dynamic law causing \(p\) with \(\psi=q\lor\neg q\)
is omitted because it refers to the removed fluent. With no other laws,
the successor is \(\{\neg p\}\). Replacing that condition by \(\top\)
keeps the law and gives \(\{p\}\) instead.

The formula \(\varphi\) is evaluated before unloading, so it
may refer to private fluents that the action removes. We call a dynamic law for
\(\mathit{unload}(M)\) \emph{target-safe} if its head and formula \(\psi\)
mention only fluents in \(\mathcal F[K\setminus\{M\}]\) for every active
set \(K\) containing \(M\) in which the law is present. Since there are
finitely many possible sets \(K\), this condition can be checked directly
from the module signatures.

RAT also allows laws that are not target-safe. Such a law is omitted from
a step if its head or \(\psi\) mentions a fluent absent after
unloading. The translation in Section~\ref{sec:compilation} makes the
same check using the fluents in the head and the unsimplified formula
\(\psi\).

\subsection{RAT transitions and executions}

We can now define a transition for RAT.

\begin{definition}[RAT transition]
\label{def:rat-transition}
For causally admissible configurations \(\langle K,s\rangle\) and
\(\langle K',s'\rangle\),
\[
  \langle K,s\rangle\trans{a}_{\mathrm{RAT}}\langle K',s'\rangle
\]
iff all of the following hold:
\begin{enumerate}[leftmargin=*,label=(\roman*)]
  \item \(a\in\enabled_K(s)\);
  \item \(K'=\mathsf{Next}_K(a)\);
  \item for some initialization assignment
    \(Y\in\mathsf{Init}_{K\rightarrow K'}(s)\),
    \(I_{s'}\in\mathsf{CE}(T_{K,s,a,K',Y})\) for the state \(s'\).
\end{enumerate}
Static laws, dynamic laws, inertia, and initialization are solved jointly,
including when a dynamic law's formula \(\psi\) depends on fluent values
in the state after the action.
This definition is declarative:
the conditions are simultaneous semantic constraints, not an operational
evaluation order.
\end{definition}

The last condition also ensures that the target is causally admissible.
In the causal reduct for a target state \(s'\), only static laws can support
the truth values of defined fluents. If we replace the causes of basic literals by the facts
\(\mathsf{Base}_{K'}(s')\), the reduct still has the same unique model.
This is exactly the static admissibility test. The truth values of basic
fluents may nevertheless change indirectly during the step.

These transitions form a \emph{labeled transition system}: its nodes are
the causally admissible RAT configurations, and an edge labeled \(a\)
connects two configurations exactly when they satisfy
Definition~\ref{def:rat-transition} for action \(a\).
For actions \(a\in\mathcal A[K]\setminus\mathcal A^{\mathsf r}\),
that is, actions other than loading and unloading, \(K\) does not change.
Their transitions coincide with those of the ordinary description
\(\Dof K\). Proposition~\ref{prop:fixed-theory} in
Section~\ref{sec:fixed-theory} proves this for one step;
Corollary~\ref{cor:fixed-theory-path} in the same section extends it to
finite executions.

An action \emph{occurs} when the transition relation holds, not merely when
its label is available or enabled. From a fixed source configuration,
the same action may have several successors because of alternative
initializations or causal explanations. All these successors have the
same module set.

A finite \emph{action sequence} is an ordered tuple
\(\alpha=\langle a_1,\ldots,a_n\rangle\), where \(a_i\in\mathcal A(\Sigma^*)\) for
\(1\leq i\leq n\).
A finite \emph{execution}, or \emph{path}, is a sequence
\[
  \Config_0\trans{a_1}_{\mathrm{RAT}}\Config_1
  \trans{a_2}_{\mathrm{RAT}}\cdots
  \trans{a_n}_{\mathrm{RAT}}\Config_n,
\]
with causally admissible configurations
\(\Config_i=\langle K_i,s_i\rangle\) for \(0\leq i\leq n\), and
each step satisfying Definition~\ref{def:rat-transition}.
We write \(\Config\trans{\alpha}_{\mathrm{RAT}}\Config'\) when such a
path starts at \(\Config\) and ends at \(\Config'\). The empty sequence
leaves its starting configuration unchanged. 

\subsection{Transitions for the running example}
\label{sec:running-transitions}

For the instance in Section~\ref{sec:running-example},
\(\Config_0=\langle K_0,\{\neg\mathit{taskDone}\}\rangle\), with \(K_0=\emptyset\),
is causally admissible over its core vocabulary \(\{\mathit{taskDone}\}\).
Loading is enabled because the core has no blocking law and \(\mathit{cal}\) is
inactive, with \(\mathsf{Next}_{K_0}(\mathit{load}(\mathit{cal}))=\{\mathit{cal}\}=K_1\).
No dynamic law in \(\Dof{K_0}\) changes a fluent when \(\mathit{cal}\) is loaded.
Inertia preserves \(\neg\mathit{taskDone}\), and initialization supplies
\(\{\neg\mathit{registered}(\mathit{cal}),\neg\mathit{answer}(391)\}\).
The calculator's static laws explain \(\neg\mathit{ready}(\mathit{cal})\).
Together these conditions give the unique target state
\[
\begin{gathered}
  s_1=\{\neg\mathit{taskDone},\neg\mathit{registered}(\mathit{cal}),
    \neg\mathit{answer}(391),\neg\mathit{ready}(\mathit{cal})\},\\
  \Config_1=\langle K_1,s_1\rangle.
\end{gathered}
\]
Hence \(\Config_0\trans{\mathit{load}(\mathit{cal})}_{\mathrm{RAT}}\Config_1\).
No calculator law is active before loading; its static laws apply afterward.
Multiplication is visible but disabled because \(\neg\mathit{ready}(\mathit{cal})\in s_1\).

Registration causes \(\mathit{registered}(\mathit{cal})\). Inertia preserves
\(\neg\mathit{answer}(391)\), and the static laws explain
\(\mathit{ready}(\mathit{cal})\). Multiplication then causes \(\mathit{answer}(391)\).
For the configurations
\[\begin{aligned}
  \Config_2&=\langle K_1,\{\neg\mathit{taskDone},\mathit{registered}(\mathit{cal}),
    \neg\mathit{answer}(391),\mathit{ready}(\mathit{cal})\}\rangle,\\
  \Config_3&=\langle K_1,\{\neg\mathit{taskDone},\mathit{registered}(\mathit{cal}),
    \mathit{answer}(391),\mathit{ready}(\mathit{cal})\}\rangle,
\end{aligned}\]
the remaining execution is
\[
  \Config_1\trans{\mathit{register}(\mathit{cal})}_{\mathrm{RAT}}\Config_2
  \trans{\mathit{multiply}(23,17)}_{\mathrm{RAT}}\Config_3
  \trans{\mathit{unload}(\mathit{cal})}_{\mathrm{RAT}}\Config_0.
\]
The first two steps preserve \(K_1\); unloading retains
\(\neg\mathit{taskDone}\) by inertia. Multiplication stores the answer
but does not mark the task complete.
The return to \(\Config_0\) occurs because no core fluent has changed.
It does not make unloading and reloading inverse operations: reloading
from \(\Config_0\) gives \(\Config_1\), not \(\Config_3\), because
initialization does not restore the earlier registration or stored answer.
Section~\ref{sec:loading-order} gives a further example of why
unloading and reloading need not restore the earlier state.

After multiplication, we can also finish the task instead of unloading
the calculator. From \(\Config_3\), the core action
\(\mathit{finishTask}\) leads to
\[
  \Config_4=\langle K_1,\{\mathit{taskDone},\mathit{registered}(\mathit{cal}),
    \mathit{answer}(391),\neg\mathit{ready}(\mathit{cal})\}\rangle.
\]
The calculator remains loaded and registered, and the stored answer is
unchanged. Its static laws make \(\mathit{ready}(\mathit{cal})\) false because
\(\mathit{taskDone}\) is now true, so further multiplication is blocked.
If we then unload the calculator, inertia preserves the now-true core
fluent \(\mathit{taskDone}\), giving
\[
  \Config_4\trans{\mathit{unload}(\mathit{cal})}_{\mathrm{RAT}}
  \langle\emptyset,\{\mathit{taskDone}\}\rangle\ne\Config_0.
\]

With or without the calculator, \(\mathit{finishTask}\) still makes
\(\mathit{taskDone}\) true. The module reads this core fluent without
preventing the core action or changing its effect on that fluent.
Theorem~\ref{thm:readonly-extension} in
Section~\ref{sec:readonly-extension} gives sufficient conditions for
this preservation of existing behavior.

\section{Properties of RAT}
\label{sec:rat-properties}

Changing a harness component can affect tasks that use other components.
A designer may therefore want to know whether existing actions remain
usable, whether the model determines one next configuration for a given action, and whether
loading two tools in either order permits the same final configurations.
For a RAT model, these questions can be answered by examining its causal
laws and initializers. We give sufficient conditions for these properties,
then define the corresponding reasoning tasks and establish their
computational complexity.

Throughout this section, we use a finite RAT system
\(
  \mathcal R=\left\langle\Modules,\mathcal D_{\mathsf{core}},
    {\{\mathsf{Mod}(M)\}}_{M\in\Modules}\right\rangle
\)
as in Definition~\ref{def:rat-system}.
Let \(K\subseteq\Modules\) be the active module set in
\(\mathcal R\). The symbols \(\Dof K\), \(\mathcal F[K]\), and
\(\mathcal A[K]\) are its active description, fluents, and actions.
All transition relations and initializers refer to \(\mathcal R\).
The properties hold for every such system.

The proofs use a simple fact about literal-headed causal theories.
Let \(T\) be such a theory over atoms \(\sigma\) and
\(I\subseteq\sigma\) an interpretation. Its causal reduct \(T^I\),
defined in Section~\ref{sec:causal-foundation}, contains the heads of the
rules whose bodies are true in \(I\). Here those heads are literals or
\(\bot\). Thus \(I\) is the unique classical model of \(T^I\) iff
\(T^I=\{p\mid p\in I\}\cup\{\neg p\mid p\in\sigma\setminus I\}\).
The reduct must therefore specify exactly the truth value assigned by
\(I\) to each atom.
We use this criterion to check
causal explanations below.

\subsection{Transitions without loading or unloading}
\label{sec:fixed-theory}

Let \(K\) be the active module set in the RAT system \(\mathcal R\),
and let \(a\in\mathcal A[K]\setminus\mathcal A^{\mathsf r}\) be a
ground action that neither loads nor unloads a module. The core and the
modules in \(K\) form the ordinary action description \(\Dof K\).
We compare its states and transitions, defined in
Section~\ref{sec:action-theory-foundation}, with those of RAT.

For a state \(s\in\mathsf{St}(\mathcal F[K])\), its basic literals
form the assignment \(\mathsf{Base}_K(s)\). The ordinary action
description and RAT use exactly the same causal theory to determine
whether the static laws explain \(s\) with this basic assignment.
Equation~\eqref{eq:ordinary-state-theory} and
Equation~\eqref{eq:static-causal-theory} give
\[
  \mathsf{CT}_{\Dof K}(\mathsf{Base}_K(s))
  =\mathsf{CT}_K(\mathsf{Base}_K(s)).
\]
Both theories contain facts for the basic values in \(s\) and all
static laws of \(\Dof K\). Thus \(s\) is a causally admissible state
of the action description \(\Dof K\) exactly when \(\langle K,s\rangle\)
is a causally admissible RAT configuration. We now show that executing
\(a\) permits the same next states in both descriptions.

\begin{proposition}
\label{prop:fixed-theory}
Let \(\langle K,s\rangle\) and \(\langle K',s'\rangle\) be causally
admissible configurations of the RAT system \(\mathcal R\), and let
\(a\in\mathcal A[K]\setminus\mathcal A^{\mathsf r}\) be a ground action.
Then
\[
  \langle K,s\rangle\trans{a}_{\mathrm{RAT}}\langle K',s'\rangle
  \quad\text{if{f}}\quad
  K'=K\ \text{ and }\ s\trans{a}_{\Dof K}s'.
\]
\end{proposition}
\begin{proof}
Since \(a\notin\mathcal A^{\mathsf r}\),
\(  \mathsf{Next}_K(a)=K\),
  \(\mathsf{Init}_{K\rightarrow K}(s)=\{\emptyset\}
\).
No fluent is added or removed, so every selected dynamic law is target
applicable and there are no initialization facts. Comparing
Equation~\eqref{eq:step-causal-theory} with
Equation~\eqref{eq:ordinary-step-causal-theory} gives
\(T_{K,s,a,K,\emptyset}=T_{\Dof K,s,a}\).
Ordinary executability, \(\mathsf{Enabled}_{\Dof K}(s)\), tests whether
a nonexecutability law blocks the action. RAT's
\(\mathsf{Enabled}_K(s)\) in Equation~\eqref{eq:enabled} adds the
module-status tests for loading and unloading. Since \(a\) is neither,
\(a\in\mathsf{Enabled}_K(s)\) iff
\(a\in\mathsf{Enabled}_{\Dof K}(s)\).
The ordinary transition test in Equation~\eqref{eq:ordinary-transition}
and the RAT test in Definition~\ref{def:rat-transition} therefore impose
the same conditions, proving both directions.
\end{proof}

Applying Proposition~\ref{prop:fixed-theory} to each action gives the
following result for finite executions. No step changes \(K\), and
each step permits exactly the same next states in both systems. For an
empty sequence, both executions consist of the initial state alone.

\begin{corollary}
\label{cor:fixed-theory-path}
Let \(K\) be the active module set in the RAT system \(\mathcal R\),
and let \(\alpha=\langle a_1,\ldots,a_n\rangle\) be a finite action
sequence with
\(a_i\in\mathcal A[K]\setminus\mathcal A^{\mathsf r}\) for
\(1\leq i\leq n\). For any states
\(s_0,\ldots,s_n\in\mathsf{St}(\mathcal F[K])\) that are causally
admissible under \(K\), the chain
\[
  \langle K,s_0\rangle\trans{a_1}_{\mathrm{RAT}}
  \langle K,s_1\rangle\trans{a_2}_{\mathrm{RAT}}\cdots
  \trans{a_n}_{\mathrm{RAT}}
  \langle K,s_n\rangle
\]
is a RAT execution iff
\[
  s_0\trans{a_1}_{\Dof K}s_1\trans{a_2}_{\Dof K}\cdots
  \trans{a_n}_{\Dof K}s_n
\]
is an execution of the action description \(\Dof K\).
\end{corollary}

Thus, for sequences without loading or unloading actions, RAT and the
action description \(\Dof K\) have exactly the same paths, including
all indirect effects permitted by its static laws.

\subsection{When the next configuration is unique}
\label{sec:rat-determinacy}

After registering a tool or finishing a task, a harness needs updated
status information before choosing its next action. If the model gives
one next configuration, it can predict which tools are ready and which
actions are enabled without maintaining several possible states.
A designer can check sufficient conditions for this uniqueness before
using the action.
The result below gives a way to compute this configuration, or determine
that no successor exists, once the initialization assignment is supplied.
We first give a uniqueness condition for defined values when the basic
assignment is fixed. We then add conditions on initialization and on
the dependencies among effects to determine a unique next configuration.

Let \(K\) be the active module set in the RAT system \(\mathcal R\).
We record
dependencies in the \emph{static dependency graph}
\(G_K=\langle\mathcal F^d[K],E_K\rangle\), where
\[
  E_K=\{(f,g)\in\mathcal F^d[K]\times\mathcal F^d[K]\mid
    \text{there exists}\,
    (\mathsf{caused}\ \ell\ \mathsf{if}\ \psi)\in L^s[K],\,
    \ell\in\{g,\neg g\},\,f\in\mathsf{Fl}(\psi)\}.
\]
Here \(f,g\) are defined fluents in \(\mathcal F^d[K]\),
\(\ell\) is a literal on \(g\), and \(\psi\) is the condition of
the displayed static law. An edge \((f,g)\)
means that a law defining \(g\) mentions \(f\) in its condition.
Only static laws with defined-fluent heads contribute edges.
If this graph has no directed cycle, its fluents can be ordered so that
each depends only on basic fluents and earlier defined fluents. We use
such an order in the proof below. RAT also permits cyclic definitions.
Acyclicity is only a sufficient condition for uniqueness.

\begin{proposition}
\label{prop:closure}
Let \(K\) be the active module set in the RAT system \(\mathcal R\), and let
\(G_K\) be the static dependency graph of \(\Dof K\).
If \(G_K\) is acyclic, then
\(
  |\cl_K(X)|\leq 1
\)
for every complete assignment \(X\in\mathsf{St}(\mathcal F^b[K])\)
to the basic fluents.
\end{proposition}
\begin{proof}
We show that any two states in \(\cl_K(X)\) must be equal.
Fix a complete basic assignment \(X\), and let
\(s_1,s_2\in\cl_K(X)\) be two states. Their classical interpretations
are \(I_{s_1}=s_1\cap\mathcal F[K]\) and
\(I_{s_2}=s_2\cap\mathcal F[K]\).
The states agree on all basic fluents because both contain \(X\).
Since \(G_K\) is acyclic, choose
an ordering \(g_1,\ldots,g_r\) of \(\mathcal F^d[K]\) in which every
static law with head \(g_i\) or \(\neg g_i\) mentions only basic fluents and
\(g_1,\ldots,g_{i-1}\) in its condition.

Suppose the states differ, and let \(g_j\) be the first fluent in this
order on which they differ. They agree on every fluent used in the
conditions of laws with head \(g_j\) or \(\neg g_j\). The reducts of
\(\mathsf{CT}_K(X)\) at \(I_{s_1}\) and \(I_{s_2}\) therefore
contain exactly the same heads concerning \(g_j\).
The facts supplied by \(X\) concern only basic fluents, so these static
heads are the only possible support for \(g_j\)'s value. Each reduct
must contain exactly one of \(g_j,\neg g_j\). With neither, it would
not have a unique model. With both, it would have no model.
The same literal must therefore belong to both \(s_1\) and \(s_2\),
contrary to their assumed difference at \(g_j\). Thus \(s_1=s_2\),
which proves \(|\cl_K(X)|\leq1\).
\end{proof}

Proposition~\ref{prop:closure} concerns the defined fluents after all
basic values have been fixed. To obtain a unique next configuration,
we also need to determine the next basic values. The following result
extends the same dependency-order argument to both basic and defined
fluents in the state after the action.
The conditions below also cover indirect changes to basic fluents and
dynamic laws whose effects depend on values after the action.

\begin{proposition}
\label{prop:unique-successor}
Let \(\langle K,s\rangle\) be a causally admissible configuration of
the RAT system \(\mathcal R\), and let
\(a\in\mathsf{Enabled}_K(s)\) be an enabled ground action.
Its target module set is \(K'=\mathsf{Next}_K(a)\).
Suppose the following conditions hold:
\begin{enumerate}[leftmargin=*,label=(\roman*)]
  \item The combined initializer supplies one assignment, i.e.,
    \(\mathsf{Init}_{K\rightarrow K'}(s)=\{Y\}\) for an assignment $Y$.
  \item There is an ordering of all fluents \(f_1,\ldots,f_m\) in \(\mathcal F[K']\),
    so that for every literal-headed rule \(\ell\Leftarrow\psi\) in
    \(\mathsf{Static}(K')\cup\mathsf{Dynamic}(K,s,a,K')\), 
    \(
      \mathsf{fl}(\ell)=f_i
      \text{ implies }
      \mathsf{Fl}(\psi)\subseteq\{f_1,\ldots,f_{i-1}\}\).
\end{enumerate}
Then, in the RAT system \(\mathcal R\), there is at most one configuration
\(\langle K',s'\rangle\) satisfying
\[
  \langle K,s\rangle\trans{a}_{\mathrm{RAT}}\langle K',s'\rangle.
\]
\end{proposition}
\begin{proof}[Proof sketch]
The enabled action~$a$ fixes \(K'\), and assumption (i) fixes \(Y\).
We assign the fluents in the order of (ii), show that every successor
must have these values, and then check the static constraints.
\begin{enumerate}[leftmargin=*,label=(\roman*)]
  \item \emph{Construct a candidate state.}
    Start with the empty assignment \(s'_0=\emptyset\).
    For each \(f_i\), let \(H_i\subseteq\{f_i,\neg f_i\}\) consist of
    the heads on \(f_i\) of rules in
    \(\mathsf{Static}(K')\cup\mathsf{Dynamic}(K,s,a,K')\)
    whose bodies are true in \(s'_{i-1}\), together with any literal
    in \(Y\cap\{f_i,\neg f_i\}\). By the ordering assumption, all these bodies use
    only fluents already assigned.

    If \(H_i=\{f_i,\neg f_i\}\), the reduct for any state extending
    \(s'_{i-1}\) contains both literals and has no classical model,
    so the construction stops. If \(|H_i|=1\), its literal must
    belong to every successor extending \(s'_{i-1}\).
    If \(H_i=\emptyset\), \(f_i\) cannot be a newly introduced
    basic fluent, because \(Y\) assigns each such fluent a literal. For a defined
    fluent, neither \(f_i\) nor \(\neg f_i\) has causal support,
    so the reduct cannot determine its value uniquely and the
    construction also stops. The remaining case is a basic
    fluent present before and after the action. Only its inertia rule
    \(\ell\Leftarrow\ell\), with \(\ell\in s\cap\{f_i,\neg f_i\}\),
    can supply a literal on \(f_i\). Changing its value makes this
    rule's body false and leaves the new value without causal support.
    Thus its value must remain as in \(s\).

    After these checks, the only possible assignments are
    \[
      s'_i=
      \begin{cases}
        s'_{i-1}\cup H_i,& |H_i|=1,\\
        s'_{i-1}\cup\bigl(s\cap\{f_i,\neg f_i\}\bigr),&
          H_i=\emptyset,\ f_i\in\mathcal F^b[K]\cap\mathcal F^b[K'].
      \end{cases}
    \]
    Each \(s'_i\) assigns exactly \(f_1,\ldots,f_i\).
    If the construction reaches \(i=m\), the candidate state is
    \(s'=s'_m\).

  \item \emph{Prove uniqueness.}
    Every successor agrees with \(s'_0\). If it agrees with
    \(s'_{i-1}\), the ordering assumption gives the same set \(H_i\).
    The checks in (i) then either rule out a successor or force the
    assignment \(s'_i\). By induction, stopping excludes every
    successor, and completing the construction leaves \(s'_m\)
    as the only possible successor state.

  \item \emph{Check existence.}
    If the construction yields a complete state \(s'\), let
    \(I_{s'}=s'\cap\mathcal F[K']\) be its classical
    interpretation. Check every static constraint
    \(\bot\Leftarrow\psi\) in \(\mathsf{Static}(K')\).
    If \(s'\models\psi\) for any such rule, no transition is possible.
    Otherwise the static and dynamic rules whose bodies are true in
    \(s'\), together with the initialization facts, contribute exactly
    \(\bigcup_{i=1}^m H_i\). The inertia rules contribute exactly
    \(\mathsf{Base}_K(s)\cap s'\). Therefore
    \(
      {(T_{K,s,a,K',Y})}^{I_{s'}}
      =\bigcup_{i=1}^m H_i\ \cup\
        \bigl(\mathsf{Base}_K(s)\cap s'\bigr)
      =s'.
    \)
    This reduct has the unique model \(I_{s'}\).
    Every defined literal of \(s'\) comes from a static rule.
    Together with facts for its basic literals, these rules also give
    \({\mathsf{CT}_{K'}(\mathsf{Base}_{K'}(s'))}^{I_{s'}}=s'\),
    so \(\langle K',s'\rangle\) is causally admissible.
    Definition~\ref{def:rat-transition} now gives the transition
    from \(\langle K,s\rangle\) to \(\langle K',s'\rangle\)
    under \(a\). By the uniqueness argument in (ii), it is the only
    such transition.
    Thus a completed construction gives the unique successor
    exactly when \(s'\) satisfies all static constraints.\qedhere
\end{enumerate}
\end{proof}

The construction in the above proof either returns the unique successor configuration or
establishes that no such transition exists in \(\mathcal R\).

In the running example in Section~\ref{sec:running-example},
order the fluents of \(\mathcal F[K']\) as follows. If \(\mathit{cal}\notin K'\),
the only fluent is \(\mathit{taskDone}\). If \(\mathit{cal}\in K'\), use
\(
  \mathit{taskDone},\mathit{registered}(\mathit{cal}),\mathit{answer}(391),
  \mathit{ready}(\mathit{cal}).
\)
The combined initializer supplies one assignment, as required by (i).
Every selected dynamic rule has body \(\top\), and the two static
laws for \(\mathit{ready}(\mathit{cal})\) refer only to
\(\mathit{taskDone}\) and \(\mathit{registered}(\mathit{cal})\), which occur
earlier in this order. Thus (ii) also holds, and
Proposition~\ref{prop:unique-successor} applies to every enabled action
in the running example.

\subsection{Adding a module without changing existing behavior}
\label{sec:independent-extension}

A harness may add an optional component while an existing task continues.
It then needs to know whether loading changes the task state or prevents
an action that was previously possible. We first give conditions
for a module that shares no fluents or actions with the current description.
Section~\ref{sec:readonly-extension} then considers modules whose laws
also refer to fluents of the core or other active modules.

Let \(K\) be the active module set in the RAT system \(\mathcal R\), and let
\(M\in\Modules\setminus K\) be the module to be added. To check
\(M\)'s static laws separately, fix a complete assignment
\(Y\in\mathsf{St}(\mathcal F^b(\Sigma_M))\) and define the local
causal theory over \(\mathcal F(\Sigma_M)\) by
\[
  \mathsf{CT}^M(Y)=\{\ell\Leftarrow\top\mid\ell\in Y\}
    \cup\{h\Leftarrow\psi\mid
      (\mathsf{caused}\ h\ \mathsf{if}\ \psi)\in L_M^s\}.
\]
The states causally explained by this theory form the set
\[
  \cl^M(Y)=\{s_M\in\mathsf{St}(\mathcal F(\Sigma_M))\mid
    s_M\cap\mathsf{Lit}(\mathcal F^b(\Sigma_M))=Y,\
    I_{s_M}\in\mathsf{CE}(\mathsf{CT}^M(Y))\}.
\]
Here \(s_M\) is a complete state of \(M\), while
\(I_{s_M}=s_M\cap\mathcal F(\Sigma_M)\) is its corresponding classical
interpretation. The basic-projection equality is already enforced by
the facts for \(Y\) in \(\mathsf{CT}^M(Y)\). The causal explanation
condition is the same as in Section~\ref{sec:static-closure}.

We require \(M\) to share no symbols with \(\Dof K\), and loading
\(M\) to have no dynamic effects specified by \(\Dof K\).
We also require its initialization assignments to permit states under
\(M\)'s static laws. The following definition states these conditions.

\begin{definition}
\label{def:rat-independent-extension}
Let \(K\) be the active module set in the RAT system \(\mathcal R\).
An inactive module \(M\in\Modules\setminus K\) is an
\emph{independent extension of \(K\)} if the following conditions hold.
\begin{enumerate}[leftmargin=*,label=(\roman*)]
  \item The fluent and action vocabularies of \(M\) are disjoint
    from those of \(\Dof K\), i.e., 
    \(
      \mathcal F(\Sigma_M)\cap\mathcal F[K]=\emptyset,
      \,
      \mathcal A(\Sigma_M)\cap\mathcal A[K]=\emptyset.
    \)
  \item There is no dynamic law
    \(
      (\mathsf{caused}\ \ell\ \mathsf{if}\ \psi\
        \mathsf{after}\ \mathit{load}(M)\land\varphi)\in L^d[K].
    \)
  \item For every state \(s\) such that \(\langle K,s\rangle\) is
    causally admissible, \(\mathsf{Init}_M(s)\) is nonempty.
    Each assignment \(Y\in\mathsf{Init}_M(s)\) to \(M\)'s private
    basic fluents must permit at least one state under its static laws, i.e.,
    \(
      \cl^M(Y)\neq\emptyset,
      \text{ for every }Y\in\mathsf{Init}_M(s).
    \)
\end{enumerate}
\end{definition}

Since the core fluents belong to \(\mathcal F[K]\), condition (i)
implies \(\mathcal F^b(\Sigma_M)=\mathcal F^b_M\). Thus the initializer
assigns all basic fluents of \(M\), and condition (iii) requires that
each assignment can be extended to a state of \(M\). These are
sufficient conditions for the results below, not requirements on every
RAT module.

Because \(M\) and \(\Dof K\) share no fluents, their states can be
checked separately and then combined by set union.

\begin{lemma}
\label{lem:closure-factorization}
Let \(K\) be the active module set in the RAT system \(\mathcal R\),
and let \(M\) be an independent extension of \(K\). For complete basic
assignments \(X\) over \(\mathcal F^b[K]\) and \(Y\) over
\(\mathcal F^b(\Sigma_M)\),
\begin{equation}
  \cl_{K\cup\{M\}}(X\cup Y)
  =
  \{s\cup s_M\mid s\in\cl_K(X),\ s_M\in\cl^M(Y)\}.
  \label{eq:closure-factorization}
\end{equation}
\end{lemma}

\begin{proof}
The laws of \(\Dof K\) mention only \(\mathcal F[K]\), while the laws
of \(M\) mention only \(\mathcal F(\Sigma_M)\). The reduct of the
combined theory is therefore the union of their reducts over disjoint
atom sets. A model of this union consists of one model from each reduct.
It is unique exactly when both models are unique. This gives
Equation~\eqref{eq:closure-factorization}.
\end{proof}

When an independent module \(M\) is loaded, no dynamic rule assigns
a fluent in \(\mathcal F[K]\), and \(M\)'s laws cannot mention 
fluents in \(\mathcal F[K]\). Their values after loading are therefore determined by the
static laws of \(\Dof K\) and inertia from the state \(s\) before loading.
Inertia does not fix the basic literals of \(s\) as facts.
Static laws may still support different basic values.
To determine whether loading preserves \(s\), we therefore collect
all states explained by these static laws and inertia.
For a causally admissible state \(s\) under \(K\), define
\[
  \mathsf{Succ}^{0}_K(s)=
  \{v\in\mathsf{St}(\mathcal F[K])\mid
    I_v\in\mathsf{CE}(\mathsf{Static}(K)
      \cup\{\ell\Leftarrow\ell\mid\ell\in\mathsf{Base}_K(s)\})\}.
\]
The superscript \(0\) indicates that no dynamic rules are included.
The interpretation is \(I_v=v\cap\mathcal F[K]\).
Since \(s\) is causally admissible, evaluating these rules at \(I_s\)
gives exactly the literals of \(s\), so \(s\in\mathsf{Succ}^{0}_K(s)\).
The equality \(\mathsf{Succ}^{0}_K(s)=\{s\}\) means that static laws
and inertia permit only \(s\). By contrast,
\(\cl_K(\mathsf{Base}_K(s))\) uses facts to fix the basic values
and does not test whether they can change during loading.

For a causally admissible configuration \(\langle K,s\rangle\) and
a ground action \(a\in\mathcal A[K]\), define its possible successor
states by
\[
  \mathsf{Succ}_K(s,a)=
  \{v\mid\langle K,s\rangle\trans{a}_{\mathrm{RAT}}
         \langle\mathsf{Next}_K(a),v\rangle\}.
\]
Each state in this set assigns the fluents in
\(\mathcal F[\mathsf{Next}_K(a)]\). The set is empty if no
\(a\)-transition is possible from \(\langle K,s\rangle\).

Theorem~\ref{thm:conservative}(i) determines the possible values of
\(\mathcal F[K]\) after loading \(M\).
Part (ii) compares execution of the same action \(a\) with \(M\)
inactive and active, starting from states that agree on
\(\mathcal F[K]\). It shows that \(M\) does not change the possible
next values of those fluents.

\begin{theorem}[Preservation by an independent module]
\label{thm:conservative}
Let \(K\) be the active module set in the RAT system \(\mathcal R\),
let \(M\) be an independent extension of \(K\), and let \(s\) be
a causally admissible state under \(K\).
\begin{enumerate}[leftmargin=*,label=(\roman*)]
  \item If \(\mathit{load}(M)\in\enabled_K(s)\), then
    \(
      \{v\cap\mathsf{Lit}(\mathcal F[K])\mid
        v\in\mathsf{Succ}_K(s,\mathit{load}(M))\}
      =\mathsf{Succ}^{0}_K(s).
    \)
  \item Let \(u\) be a causally admissible state under
    \(K\cup\{M\}\). If \(u\cap\mathsf{Lit}(\mathcal F[K])=s\), then
    for every ground action
    \(a\in\mathcal A[K]\setminus\mathcal A^{\mathsf r}\),
    \(
      \{v\cap\mathsf{Lit}(\mathcal F[K])\mid
        v\in\mathsf{Succ}_{K\cup\{M\}}(u,a)\}
      =\mathsf{Succ}_K(s,a).
    \)
\end{enumerate}
\end{theorem}

\begin{proof}[Proof sketch]
We separate the rules over \(\mathcal F[K]\) from those over
\(\mathcal F(\Sigma_M)\). Since these fluent sets are disjoint,
the unique-model test can be applied to the two sets of rules separately,
as in Lemma~\ref{lem:closure-factorization}.
\begin{enumerate}[leftmargin=*,label=(\roman*)]
  \item Assume \(\mathit{load}(M)\in\enabled_K(s)\).
    During loading, the rules over \(\mathcal F[K]\) are exactly
    the static laws of \(\Dof K\) and inertia from \(s\).
    The rules over \(\mathcal F(\Sigma_M)\) form
    \(\mathsf{CT}^M(Y)\), for an assignment
    \(Y\in\mathsf{Init}_M(s)\). Hence
    \(
      \mathsf{Succ}_K(s,\mathit{load}(M))
      =\{v\cup s_M\mid v\in\mathsf{Succ}^{0}_K(s),\
          Y\in\mathsf{Init}_M(s),\ s_M\in\cl^M(Y)\}.
    \)
    Definition~\ref{def:rat-independent-extension}(iii) supplies at least
    one such \(Y\) and \(s_M\). Therefore every
    \(v\in\mathsf{Succ}^{0}_K(s)\) occurs in a loading successor,
    and every loading successor restricts to such \(v\). This proves (i).

  \item The assumption on \(u\) gives \(u=s\cup s_M\), where
    \(s_M=u\cap\mathsf{Lit}(\mathcal F(\Sigma_M))\).
    By Lemma~\ref{lem:closure-factorization}, \(s_M\) is causally
    admissible under \(M\)'s static laws. The module cannot mention
    \(a\), so it neither blocks \(a\) nor adds dynamic rules for \(a\).
    If \(a\) is blocked by \(\Dof K\), both successor sets are empty.
    Otherwise the step theory separates into
    \(T_{K,s,a,K,\emptyset}\) and \(M\)'s static laws with inertia
    from \(s_M\). The latter rules permit \(s_M\) itself, because their
    reduct at \(I_{s_M}\) is the same as for its admissibility test.
    Thus every \(v\in\mathsf{Succ}_K(s,a)\) gives an extended
    successor \(v\cup s_M\). Conversely, if \(w\) is an extended
    successor, its reduct over \(\mathcal F[K]\) must have a unique
    model. Therefore \(w\cap\mathsf{Lit}(\mathcal F[K])\) belongs to
    \(\mathsf{Succ}_K(s,a)\). This proves both inclusions in (ii).
\end{enumerate}
The causal explanations used in both parts also give causally admissible
configurations, as established after Definition~\ref{def:rat-transition}.
\end{proof}

When loading is enabled, part (i) guarantees at least one successor because
\(s\in\mathsf{Succ}^{0}_K(s)\). In that case, all loading successors retain the
values in \(s\) exactly when \(\mathsf{Succ}^{0}_K(s)=\{s\}\).
Part (ii) compares configurations that already agree on
\(\mathcal F[K]\). If \(u\cap\mathsf{Lit}(\mathcal F[K])\ne s\),
the comparison instead uses that restricted state in place of \(s\).

\paragraph{Why shared interfaces need separate conditions}
\label{rem:nonconservative-loading}
Consider a harness that loads a policy module after a task has finished.
Let \(\Dof K\) have one basic fluent \(\mathit{taskDone}\) and no
laws. Let \(P\in\Modules\setminus K\) be the policy module.
Its signature declares the shared
basic fluent \(\mathit{taskDone}\), but no private fluents.
Its initializer returns \(\{\emptyset\}\), and its only law is
\(
  \mathsf{caused}\ \bot\ \mathsf{if}\ \mathit{taskDone}.
\)
In the configuration \(\langle K,\{\mathit{taskDone}\}\rangle\),
\(\mathit{load}(P)\) is enabled because \(P\) is inactive and no
nonexecutability law blocks loading.
However, neither possible value of \(\mathit{taskDone}\) gives a successor.
If it is true after loading, \(P\)'s constraint contributes \(\bot\)
to the reduct. If it is false, no rule supports
\(\neg\mathit{taskDone}\). Thus loading has no successor.
The policy module \(P\) shares a fluent with \(\Dof K\), violating
Definition~\ref{def:rat-independent-extension}(i).
This example shows why merely refraining from assigning a shared fluent
does not guarantee preservation. A constraint that reads the fluent can
still prevent a transition.

\subsection{Adding a module that reads existing fluents}
\label{sec:readonly-extension}

A harness component may need to read the task state without changing
which actions the task can perform or their possible results. In the
running example, the calculator reads \(\mathit{taskDone}\) to determine
\(\mathit{ready}(\mathit{cal})\). Its vocabulary therefore cannot be disjoint from
the core's. We give conditions under which adding such a module preserves
the possible transitions on the fluents already in \(\Dof K\).
The module may still update its own private fluents.

Let \(K\subseteq\Modules\) be the active module set in the RAT system
\(\mathcal R\), and let \(M\in\Modules\setminus K\) be a module to be added.
Let \(F_0\) contain the fluents of \(\Dof K\), and let \(A_0\) contain
its actions other than loading and unloading. The \emph{interface} \(J\) consists of the fluents that \(M\) shares
with \(\Dof K\). The sets \(P,Q\) contain the basic and defined fluents
introduced by loading \(M\), while \(J, P, Q\) are pairwise disjoint.
Specifically, \(F_0=\mathcal F[K]\), \(A_0=\mathcal A[K]\setminus\mathcal A^{\mathsf r}\), \(J=\mathcal F(\Sigma_M)\cap F_0\), \(P=\mathcal F^b(\Sigma_M)\setminus F_0\), \(Q=\mathcal F^d(\Sigma_M)\setminus F_0\), and \(\mathcal F(\Sigma_M)=J\cup P\cup Q\).

We require the module not to cause or forbid a value in \(F_0\), block an action
in \(A_0\), or prevent a transition because its private values cannot be
assigned. The following sufficient conditions ensure this. We call
\(M\) a \emph{read-only extension over \(K\)} when they hold:
\begin{enumerate}[leftmargin=*,label=(\roman*)]
  \item \emph{Static laws determine the new defined values.}
    Every static head belongs to \(\mathsf{Lit}(Q)\), so the module
    adds no static constraint or static cause for a fluent in \(F_0\cup P\).
    There is an ordering for all fluents \(q_1,\ldots,q_m\) in \(Q\), such that each law
    with head \(q_i\) or \(\neg q_i\) has a body \(\psi\) satisfying
    \(\mathsf{Fl}(\psi)\subseteq J\cup P\cup\{q_1,\ldots,q_{i-1}\}\).
    For an assignment
    \(X\in\mathsf{St}(F_0\cup P\cup\{q_1,\ldots,q_{i-1}\})\), put
    \[
      H_i(X)=\{\ell\in\{q_i,\neg q_i\}\mid
        (\mathsf{caused}\ \ell\ \mathsf{if}\ \psi)\in L_M^s,
        \ X\models\psi\}.
    \]
    Require \(|H_i(X)|=1\) for every such \(X\) and every \(i\).
    For each fixed assignment \(X\), these laws therefore support exactly
    one of \(q_i\) and \(\neg q_i\). The supported literal may differ
    for different assignments \(X\).
  \item \emph{Dynamic laws update private basic fluents consistently.}
    Every dynamic head belongs to \(\mathsf{Lit}(P)\). For an action
    \(a\in A_0\), every dynamic law in \(L_M^d\) for \(a\) has
    the form \(a\ \mathsf{causes}\ \ell\ \mathsf{if}\ \varphi\),
    where \(\varphi\) is tested before the action. For each state
    \(u\in\mathsf{St}(F_0\cup P\cup Q)\), let
    \[
      E_a(u)=\{\ell\in\mathsf{Lit}(P)\mid
        (a\ \mathsf{causes}\ \ell\ \mathsf{if}\ \varphi)\in L_M^d,
        \ u\models\varphi\}.
    \]
    Require \(|E_a(u)\cap\{p,\neg p\}|\leq1\) for every
    \(a\in A_0\), state \(u\) above, and \(p\in P\).
    The effects may assign only some private basic fluents.
    Each unassigned fluent retains its value by inertia.

  \item \emph{Existing actions are not blocked.}
    No law \(\mathsf{nonexecutable}\ a\ \mathsf{if}\ \varphi\)
    in \(L_M^x\) has \(a\in A_0\).

  \item \emph{Loading has an initialization choice.}
    For every state \(s\in\mathsf{St}(F_0)\) for which
    \(\langle K,s\rangle\) is causally admissible,
    \(\mathsf{Init}_M(s)\ne\emptyset\).
    By Definition~\ref{def:module}, each allowed initial assignment
    \(Y\in\mathsf{Init}_M(s)\) is a complete assignment to \(P\).
    It contains exactly one of \(p\) and \(\neg p\) for each \(p\in P\),
    and no literals outside \(\mathsf{Lit}(P)\).
\end{enumerate}

The four conditions defining a read-only extension are sufficient for
preservation, not restrictions on general RAT systems. Condition (i)
permits several static laws to support the same literal, but requires
the stated dependency order and exactly one supported value for each
fixed assignment \(X\). Condition (ii) permits conditional updates to
only some fluents in \(P\), provided the effects are consistent.
The module must also permit initialization at every causally admissible
state under \(K\), add no static constraints, and introduce no blocking
laws for actions in \(A_0\).

For example, condition (i) permits the calculator's readiness to be
specified by three laws:
\[
\begin{aligned}
  &\mathsf{caused}\ \mathit{ready}(\mathit{cal})\ \mathsf{if}\
    \mathit{registered}(\mathit{cal})\land\neg\mathit{taskDone},\\
  &\mathsf{caused}\ \neg\mathit{ready}(\mathit{cal})\ \mathsf{if}\
    \neg\mathit{registered}(\mathit{cal}),\\
  &\mathsf{caused}\ \neg\mathit{ready}(\mathit{cal})\ \mathsf{if}\
    \mathit{taskDone}.
\end{aligned}
\]
The last two may both apply, but they support the same literal.
These laws determine the same readiness value as the two laws in the
running example. Condition (ii) likewise permits a single conditional
effect without requiring a rule for its negation when the condition is false.

For an independent module, Section~\ref{sec:independent-extension}
uses \(\cl^M(Y)\) to collect the local states permitted by the basic
assignment \(Y\). A read-only module also reads the interface \(J\),
whose values must be supplied by a state \(v\in\mathsf{St}(F_0)\).
For each fixed \(v\) and \(Y\in\mathsf{St}(P)\), condition (i)
determines exactly one assignment to \(Q\).
Using the stated dependency order, define
\[
\begin{gathered}
  u_0=v\cup Y,\qquad
  u_i=u_{i-1}\cup H_i(u_{i-1})\quad(1\leq i\leq m),\\
  \mathsf{ext}_M(v,Y):=u_m
    =v\cup Y\cup\bigcup_{i=1}^{m}H_i(u_{i-1}).
\end{gathered}
\]
Thus \(\mathsf{ext}_M(v,Y)\in\mathsf{St}(F_0\cup P\cup Q)\)
extends \(v\) with basic assignment \(Y\) and the defined values
supported by \(M\)'s static laws. It denotes one state, not a set of
states. If \(m=0\), the union is empty and \(\mathsf{ext}_M(v,Y)=v\cup Y\).

\begin{theorem}[Preservation by a read-only module]
\label{thm:readonly-extension}
Let \(K\) be the active module set in the RAT system \(\mathcal R\),
let \(M\) be a read-only extension over \(K\), and let \(s\)
be a causally admissible state under~\(K\).
Write \(F_0=\mathcal F[K]\),
\(A_0=\mathcal A[K]\setminus\mathcal A^{\mathsf r}\), and let
\(P=\mathcal F^b(\Sigma_M)\setminus F_0\) denote the private basic
fluents introduced by \(M\).
\begin{enumerate}[leftmargin=*,label=(\roman*)]
  \item \emph{Actions of \(\Dof K\).} For every basic assignment \(Y\in\mathsf{St}(P)\),
    \(u=\mathsf{ext}_M(s,Y)\) is causally admissible under
    \(K\cup\{M\}\), and for every ground action \(a\in A_0\),
    \[
      \{v\cap\mathsf{Lit}(F_0)\mid
        v\in\mathsf{Succ}_{K\cup\{M\}}(u,a)\}
      =\mathsf{Succ}_K(s,a).
    \]
  \item \emph{Loading.} If \(\mathit{load}(M)\in\enabled_K(s)\)
    and \(L^d[K]\) contains no law of the form
    \[
      \mathsf{caused}\ \ell\ \mathsf{if}\ \psi\
        \mathsf{after}\ \mathit{load}(M)\land\varphi,
    \]
    then
    \[
      \mathsf{Succ}_K(s,\mathit{load}(M))
      =\{\mathsf{ext}_M(v,Y)\mid
        v\in\mathsf{Succ}^{0}_K(s),\ Y\in\mathsf{Init}_M(s)\}.
    \]
\end{enumerate}
\end{theorem}
\begin{proof}[Proof sketch]
We determine the private values and compare the reducts on \(F_0\).
The laws of \(\Dof K\) mention only \(F_0\), while the module's
heads belong to \(\mathsf{Lit}(P\cup Q)\). For these literal-headed
causal theories and a complete state \(w\), its classical
interpretation \(I_w\) is a causal explanation
exactly when the reduct is the literal set \(w\).
Thus private heads cannot supply a missing literal on \(F_0\) or
remove a conflicting literal or \(\bot\).

\emph{Part (i).} Let \(Y\in\mathsf{St}(P)\) and
\(u=\mathsf{ext}_M(s,Y)\). Since \(s\) is causally admissible,
its static laws and basic facts give the reduct \(s\).
Adding \(Y\) and the heads \(H_i(u_{i-1})\) gives exactly \(u\),
so \(u\) is causally admissible. Here \(u_i\) are the assignments in
the construction of \(\mathsf{ext}_M(s,Y)\).
For \(a\in A_0\), condition (iii) leaves executability unchanged.
If \(a\) is blocked, both successor sets are empty. Otherwise,
condition (ii) gives the consistent effects \(E_a(u)\).
An effect fixes its fluent's value. If it changes that value, the inertia
rule for the previous value has a false body. Without an effect, inertia
is the only rule that can support a value of that private basic fluent,
since no static head belongs to \(\mathsf{Lit}(P)\).
The resulting complete assignment is
\(
  Y'=E_a(u)\ \cup\
    \bigcup_{\substack{p\in P,\,E_a(u)\cap\{p,\neg p\}=\emptyset}}
      \bigl(u\cap\{p,\neg p\}\bigr).
\)

For every \(s'\in\mathsf{Succ}_K(s,a)\), condition (i) determines
\(\mathsf{ext}_M(s',Y')\). Its step reduct is exactly that state,
so it is an extended successor. Conversely, every extended successor
has the private basic assignment \(Y'\). Removing its reduct heads
on \(P\cup Q\) leaves exactly the step reduct of \(\Dof K\),
evaluated at the restricted state on \(F_0\). This reduct must equal
that restricted state and contain no \(\bot\). The restricted state
therefore belongs to \(\mathsf{Succ}_K(s,a)\), proving both inclusions.

\emph{Part (ii).} The step has no dynamic rules because \(L^d[K]\)
contains no dynamic law for \(\mathit{load}(M)\), and \(M\) is not
yet active. For each
\(Y\in\mathsf{Init}_M(s)\), initializer facts fix the values in \(Y\).
On \(F_0\), the step theory contains exactly the static laws of
\(\Dof K\) and inertia used to define \(\mathsf{Succ}^{0}_K(s)\).
For each state \(v\) explained by those rules, condition (i) determines
\(\mathsf{ext}_M(v,Y)\).
Conversely, every loading successor must use such a \(v\) and \(Y\),
and its values on \(Q\) are fixed by the same construction. This proves
the equality in (ii). The set is
nonempty because \(\mathsf{Init}_M(s)\ne\emptyset\) and
\(s\in\mathsf{Succ}^{0}_K(s)\).
In both parts, these causal explanations also give causally admissible
successor configurations, as shown after Definition~\ref{def:rat-transition}.
\end{proof}

Part (i) says that the module neither removes nor adds a possible
successor on \(F_0\) for an action in \(A_0\). Starting from its extended
state \(u\), a harness can therefore perform the same action sequences over \(A_0\)
as from \(s\), with the same possible state sequences on \(F_0\).
Part (ii) describes loading itself. Under its hypotheses, loading has
at least one successor. Every such successor agrees with \(s\) on \(F_0\)
iff \(\mathsf{Succ}^{0}_K(s)=\{s\}\). Otherwise, the static laws already in
\(\Dof K\) may permit other values, as explained in
Section~\ref{sec:independent-extension}.

\subsection{Loading modules in either order}
\label{sec:loading-order}

When a harness needs two new components, it may load either one first.
We give conditions under which both orders are possible and lead to the
same set of final configurations. Disjoint module vocabularies are not
enough. An initializer may read fluents outside its module's signature,
and the first load may change the values read by the second initializer.
The conditions below ensure that each module offers the same initial
assignments in either order.

Let \(\Config=\langle K,s\rangle\) be a causally admissible configuration
of the RAT system \(\mathcal R\), and let \(\alpha\) be a finite ground
action sequence. Using the execution relation
\(\Config\trans{\alpha}_{\mathrm{RAT}}\Config'\) from
Section~\ref{sec:semantics}, define the set of configurations reachable
by this sequence as
\[
  \mathsf{Post}_{\alpha}(\Config)
    =\{\Config'\mid\Config\trans{\alpha}_{\mathrm{RAT}}\Config'\}.
\]

Two distinct inactive modules \(M,N\in\Modules\setminus K\) are
\emph{jointly independent over \(K\)} if each is an independent extension
of \(K\) and
\(
  \mathcal F(\Sigma_M)\cap\mathcal F(\Sigma_N)=\emptyset,
  \,
  \mathcal A(\Sigma_M)\cap\mathcal A(\Sigma_N)=\emptyset.
\)
Loading \(N\) adds its fluent values to the input of \(M\)'s initializer.
We require that removing those values from the input leaves the
initializer's output unchanged, and impose the same requirement on
\(N\)'s initializer when \(M\) is loaded. Formally, the initializers are
\emph{unchanged by loading the other module over \(K\)} if
\begin{equation}
    \mathsf{Init}_M(v)=
      \mathsf{Init}_M\bigl(v\cap\mathsf{Lit}(\mathcal F[K])\bigr), \qquad
    \mathsf{Init}_N(w)=
      \mathsf{Init}_N\bigl(w\cap\mathsf{Lit}(\mathcal F[K])\bigr).
  \label{eq:initializer-invariance}
\end{equation}
The first equality must hold for every state \(v\) causally admissible
under \(K\cup\{N\}\), and the second for every state \(w\) causally
admissible under \(K\cup\{M\}\).
The intersections retain exactly the literals over \(\mathcal F[K]\).
Each initializer may still offer several assignments or depend on values
of these retained fluents.

\begin{theorem}[Same final configurations in either loading order]
\label{thm:loading-order}
Let \(\Config=\langle K,s\rangle\) be a causally admissible configuration
of the RAT system \(\mathcal R\), and let \(M,N\) be jointly independent
modules over the active module set \(K\). Assume the following conditions.
\begin{enumerate}[leftmargin=*,label=(\roman*)]
  \item \emph{Unique state for the given basic assignment}, i.e.,
    \(\cl_K(\mathsf{Base}_K(s))=\{s\}\).
  \item \emph{Both loads are enabled}, i.e.,
    \(\{\mathit{load}(M),\mathit{load}(N)\}\subseteq\enabled_K(s)\).
  \item \emph{Unchanged initialization choices}, i.e.,
    both equalities in Equation~\eqref{eq:initializer-invariance} hold.
  \item \emph{No basic-fluent heads in static laws}, i.e., every static head
    in \(L^s[K]\cup L^s_M\cup L^s_N\) is a defined literal or \(\bot\).
\end{enumerate}
Then both loading sequences have an execution and the same set of final
configurations, i.e.,
\[
  \mathsf{Post}_{\langle\mathit{load}(M),\mathit{load}(N)\rangle}(\Config)
  =
  \mathsf{Post}_{\langle\mathit{load}(N),\mathit{load}(M)\rangle}(\Config)
  \ne\emptyset.
\]
\end{theorem}

\begin{proof}[Proof sketch]
We first show that either first load preserves \(s\) and permits the
second load. We then describe the final configurations explicitly.
Definition~\ref{def:rat-independent-extension}(ii) excludes dynamic laws
for \(\mathit{load}(M)\) and \(\mathit{load}(N)\) from
\(\Dof K\). Together with condition (iv), this leaves inertia as the
only support for basic fluents in \(\mathcal F[K]\), so their values
cannot change. With those values fixed, condition (i) leaves
\(s\) as the only causally admissible state on \(\mathcal F[K]\). Thus
\(\mathsf{Succ}^{0}_K(s)=\{s\}\), and either first load preserves \(s\)
on \(\mathcal F[K]\). Both loading actions belong to the core action set,
while each module's action signature is disjoint from \(\mathcal A[K]\).
Neither module can therefore add a dynamic or nonexecutability law for
either loading action. The other module remains inactive, and
the nonexecutability conditions in \(\Dof K\) retain their truth values.
Condition (ii) therefore keeps the second load enabled.

Equation~\eqref{eq:initializer-invariance} gives the second module the
same initialization choices as at \(s\), because the intermediate state
agrees with \(s\) on \(\mathcal F[K]\).
Definition~\ref{def:rat-independent-extension}(iii) ensures that each
of these choices permits a local admissible state.
The final configurations are precisely
\(\langle K\cup\{M,N\},s\cup s_M\cup s_N\rangle\), where
\(Y_M\in\mathsf{Init}_M(s)\), \(s_M\in\cl^M(Y_M)\),
\(Y_N\in\mathsf{Init}_N(s)\), and \(s_N\in\cl^N(Y_N)\).

During the second load, the same argument preserves \(s\) on
\(\mathcal F[K]\). The first module's basic assignment also persists
by (iv), while its defined values must have causal support from its
static laws.
Its resulting local state therefore belongs to \(\cl^M(Y_M)\) or
\(\cl^N(Y_N)\), respectively. Every final configuration has the form
above. Conversely, choose any configuration of that form. Load each
module with its specified assignment and local state, retaining the first
module's state during the second load. At each step, the causal reduct
separates over the disjoint fluent sets of the active descriptions and
explains their union.
Both orders therefore reach this configuration. Definition~\ref{def:rat-independent-extension}(iii)
ensures that the initializer outputs and the required sets of local
states are nonempty.
\end{proof}

Without condition (iv) of Theorem~\ref{thm:loading-order}, order can
matter even with disjoint vocabularies
and constant initializers. Let the core state be empty. Let \(M\)
introduce a basic fluent \(x\), the static law
\(\mathsf{caused}\ x\ \mathsf{if}\ x\), and the constant initializer
\(\{\{\neg x\}\}\). Let \(N\) introduce a basic fluent \(y\), no laws,
and initializer \(\{\{\neg y\}\}\). All the other hypotheses hold.
Loading \(M\) and then \(N\) can end in either
\(\{\neg x,\neg y\}\) or \(\{x,\neg y\}\), because \(M\)'s static law
can support \(x\) on the second step. If \(N\) is loaded first, loading
\(M\) last fixes \(\neg x\), so only \(\{\neg x,\neg y\}\) is possible.
Thus the two orders have different final states under standard causal
semantics.

\paragraph{Reloading need not restore the previous state}
Loading a module again uses its initializer, not its previous private
state. For example, a module \(M\) may keep a private basic fluent
\(\mathit{connected}\), initialized to false. Take an empty core state,
no static or nonexecutability laws, the action law
\(\mathit{connect}\ \mathsf{causes}\ \mathit{connected}\), and
\(\mathsf{Init}_M(s)=\{\{\neg\mathit{connected}\}\}\) for every input.
Then the following transitions exist,
\[
  \langle\{M\},\{\mathit{connected}\}\rangle
  \trans{\mathit{unload}(M)}_{\mathrm{RAT}}
  \langle\emptyset,\emptyset\rangle
  \trans{\mathit{load}(M)}_{\mathrm{RAT}}
  \langle\{M\},\{\neg\mathit{connected}\}\rangle.
\]
The displayed source can be reached by performing the action \(\mathit{connect}\) after an earlier load.
Unloading removes the value of \(\mathit{connected}\), and the next
load again initializes it to false. A harness that needs to preserve a
connection or other private state must specify how that information is
retained and restored.

\subsection{Reasoning tasks and their computational complexity}
\label{sec:rat-complexity}

A harness can use RAT to check a proposed transition, determine whether
an action can occur, or find an execution that achieves a goal.
We define these reasoning tasks and compare their computational costs.

\paragraph{Decision tasks}
We consider a finite ground RAT system \(\mathcal R\), with
explicitly listed signatures and laws and propositional law conditions.
Let \(\Config=\langle K,s\rangle\) be a causally admissible configuration
of \(\mathcal R\).
A \emph{goal condition} \(Q\) is a propositional formula whose atomic
tests ask whether a module is active, a fluent is present, or a fluent
is true. A truth test is false when the fluent is absent.
We write \(\Config\models Q\) when these tests make \(Q\) true at
\(\Config\). For example, a goal may require that the calculator module
\(\mathit{cal}\) is active and \(\mathit{registered}(\mathit{cal})\) is true.
We consider the following five tasks.
\begin{enumerate}[leftmargin=*,label=(\roman*)]
  \item \emph{Transition checking.} Given a ground action
    \(a\in\mathcal A[K]\) and a candidate configuration \(\Config'\),
    does \(\Config\trans{a}_{\mathrm{RAT}}\Config'\) hold?
  \item \emph{Successor existence.} Given a ground action
    \(a\in\mathcal A[K]\), is there a configuration \(\Config'\) such
    that \(\Config\trans{a}_{\mathrm{RAT}}\Config'\)?
  \item \emph{Possible and necessary goal satisfaction.} Given a ground
    action \(a\in\mathcal A[K]\) and a goal condition \(Q\), possible
    satisfaction asks whether some configuration \(\Config'\) satisfies
    both \(\Config\trans{a}_{\mathrm{RAT}}\Config'\) and
    \(\Config'\models Q\). Necessary satisfaction asks whether every
    configuration \(\Config'\) with
    \(\Config\trans{a}_{\mathrm{RAT}}\Config'\) satisfies \(Q\).
    Note that, if no such configuration exists, then the necessary condition is still true.
  \item \emph{Execution of a given action sequence.} Given a sequence
    \(\alpha=\langle a_1,\ldots,a_n\rangle\) of \(n\geq1\) ground
    actions from \(\mathcal A(\Sigma^*)\) and a goal condition \(Q\),
    can this sequence be executed from \(\Config\) and end at a
    configuration satisfying \(Q\)? Formally, does some \(\Config'\)
    satisfy \(\Config\trans{\alpha}_{\mathrm{RAT}}\Config'\) and
    \(\Config'\models Q\)?
    Taking \(Q=\top\) asks only whether the sequence can be executed.
  \item \emph{Reachability.} Given a goal condition \(Q\), can any finite
    action sequence, including the empty sequence, take \(\Config\) to a
    configuration satisfying \(Q\)?
    This is \emph{unbounded reachability}.
    For \emph{bounded reachability}, a nonnegative bound \(b\) is
    also supplied, and the sequence must contain at most \(b\) actions.
\end{enumerate}

\begin{theorem}[Complexity of finite-ground RAT reasoning]
\label{thm:rat-complexity}
For the finite ground RAT systems above, with initializer membership
represented by propositional formulas, the five tasks have the following
complexity.
\begin{enumerate}[leftmargin=*,label=(\roman*)]
  \item \emph{Transition checking} is in \(\mathsf P\).
  \item \emph{Successor existence} is \(\mathsf{NP}\)-complete.
  \item \emph{Possible goal satisfaction} is \(\mathsf{NP}\)-complete,
    while \emph{necessary goal satisfaction} is
    \(\mathsf{coNP}\)-complete.
  \item \emph{Execution of a given action sequence} is
    \(\mathsf{NP}\)-complete.
  \item \emph{Bounded reachability} is
    \(\mathsf{NP}\)-complete, while \emph{unbounded reachability}
    is \(\mathsf{PSPACE}\)-complete.
\end{enumerate}
\end{theorem}

\begin{proof}[Proof sketch]
The upper bounds use a polynomial test for a supplied transition.
For the lower bounds of the finite search tasks, we use one reduction
from propositional satisfiability.
Given a propositional formula \(\varphi(p_1,\ldots,p_r)\), take a core
with no fluents or laws and one module \(M\) whose only fluents are
the private basic fluents \(p_1,\ldots,p_r\).
Its initializer permits every complete assignment,
\(\mathsf{Init}_M(v)=\mathsf{St}(\{p_1,\ldots,p_r\})\). Its only law is
\(\mathsf{caused}\ \bot\ \mathsf{if}\ \neg\varphi\).
Loading from \(\langle\emptyset,\emptyset\rangle\) chooses an assignment
\(Y\). The initialization facts explain \(Y\), and the constraint permits
it exactly when \(Y\models\varphi\). Thus loading has a successor iff
\(\varphi\) is satisfiable. This construction has polynomial size.

\begin{enumerate}[leftmargin=*,label=(\roman*)]
\item \emph{Transition checking.}
For a proposed transition from \(\langle K,s\rangle\) to
\(\langle K',s'\rangle\) with action \(a\), check the state vocabularies,
source admissibility, \(a\in\enabled_K(s)\), and
\(K'=\mathsf{Next}_K(a)\).
The target state fixes the initialization assignment
\(Y=s'\cap\mathsf{Lit}(\mathcal F^b[K']\setminus\mathcal F^b[K])\).
For loading \(M\), test whether $Y\in \mathsf{Init}_M(s)$. 
Let \(I_{s'}\) be the classical interpretation represented by \(s'\).
Since every causal head is a literal or \(\bot\), the remaining test is
\({\bigl(T_{K,s,a,K',Y}\bigr)}^{I_{s'}}=s'\).
It only evaluates rule bodies and compares literal sets.
The same test on \(\mathsf{CT}_K(\mathsf{Base}_K(s))\) checks source
admissibility. Target admissibility follows as shown after
Definition~\ref{def:rat-transition}. All checks are polynomial.

\item \emph{Successor existence.}
Guess a configuration and verify the transition by (i), giving the
\(\mathsf{NP}\) upper bound. The loading construction above gives
\(\mathsf{NP}\)-hardness.

\item \emph{Possible and necessary goal satisfaction.}
For possible satisfaction, guess a configuration, check its transition
by (i), and evaluate \(Q\) there. This gives membership in
\(\mathsf{NP}\), and \(Q=\top\) in the loading construction gives
hardness. A counterexample to necessary satisfaction is a verified
successor where \(Q\) is false, giving membership in \(\mathsf{coNP}\).
With \(Q=\bot\), necessary satisfaction holds exactly when the
construction has no successor, or equivalently when \(\varphi\) is
unsatisfiable. This proves \(\mathsf{coNP}\)-hardness.

\item \emph{Execution of a given action sequence.}
Guess the \(n\) successive configurations, check each transition by (i),
and test \(Q\) at the final configuration. The explicitly listed
sequence makes this a polynomial-size certificate, giving membership
in \(\mathsf{NP}\).
The loading construction with the one-action sequence
\(\langle\mathit{load}(M)\rangle\) and \(Q=\top\) proves hardness.

\item \emph{Reachability.}
For the bound \(b\), guess at most \(b\) actions and their successive
configurations. Verify the transitions and the final goal.
This gives membership in \(\mathsf{NP}\).
For hardness, use the loading construction with \(b=1\)
and let \(Q\) require \(M\) to be active. The goal is false initially and
reachable within one step exactly when \(\varphi\) is satisfiable.

For unbounded reachability, let \(m=|\Modules|\) and
\(f=|\mathcal F(\Sigma^*)|\). There are at most \(2^m3^f\)
configurations, since each module is active or inactive and each fluent
is absent, true, or false. A reachable configuration has a path of at
most \(2^m3^f-1\) steps without repeated configurations.
A nondeterministic search stores only the current and next configurations.
Using (i) to verify each step gives membership in
\(\mathsf{NPSPACE}=\mathsf{PSPACE}\).

For hardness, encode propositional STRIPS planning, whose plan-existence
problem is \(\mathsf{PSPACE}\)-complete~\cite{bylander1994}.
Use no optional modules and make the planning atoms core basic fluents.
For each operator \(a\), with precondition \(P_a\) and consistent effect
set \(E_a\), add
\(\mathsf{nonexecutable}\ a\ \mathsf{if}\ \neg P_a\) and
\(a\ \mathsf{causes}\ \ell\) for each \(\ell\in E_a\).
With no static laws, these effects and inertia give exactly the planning
transitions. Reaching its goal is therefore the same problem in RAT.
\end{enumerate}
\end{proof}

\section{From RAT to causal theories and answer set programs}
\label{sec:compilation}

A RAT execution may change which fluents and laws are present. To
represent it by one causal theory, we keep an atom for every declared
fluent and use additional atoms to record which modules are active.
We first define this representation of a configuration. We then add
time indices and causal rules to represent a finite execution. Finally,
we apply the causal-theory translation of Ferraris et al.~\cite{ferrarisEtAl2012}
to obtain an answer set program.

Throughout this section, let
\(\mathcal R=\langle\Modules,\mathcal D_{\mathsf{core}},
{\{\mathsf{Mod}(M)\}}_{M\in\Modules}\rangle\) be a finite ground RAT system.
For an initial configuration \(\Config_0\) and an action sequence
\(\alpha\) of length \(h\), we denote the resulting causal theory by
\(\Gamma_h(\mathcal R,\Config_0,\alpha)\) and the answer set program by
\(\Pi_h(\mathcal R,\Config_0,\alpha)\).
Both represent exactly the RAT executions from \(\Config_0\) with those
action labels.

\subsection{Representing active modules in a fixed vocabulary}
\label{sec:guarded-representation}

In the running example, \(\mathit{registered}(\mathit{cal})\) records
whether the calculator has been registered for use. This fluent belongs
to the RAT state only while the calculator module is active. The
translation instead uses one fixed set of atoms for all configurations.
It keeps \(\mathit{registered}(\mathit{cal})\) in that set even before
loading and uses the atom \(\mathit{active}_{\mathit{cal}}\) to record whether
the fluent is currently part of the RAT state.

\emph{Atoms and interpretations.}
The signature \(\Sigma^*=\Sigma_{\mathsf{core}}\cup
\bigcup_{M\in\Modules}\Sigma_M\) contains all fluents and actions
declared in the RAT system \(\mathcal R\). Introduce a fresh atom
\(\mathit{active}_M\) for each module \(M\in\Modules\).
The fixed propositional vocabulary for configurations is
\[
  \sigma^*=\mathcal F(\Sigma^*)\cup
    \{\mathit{active}_M\mid M\in\Modules\}.
\]
Represent a configuration \(\langle K,s\rangle\) by the classical
interpretation
\[
  J_{K,s}=(s\cap\mathcal F[K])\cup
    \{\mathit{active}_M\mid M\in K\}\subseteq\sigma^*.
\]
As in Section~\ref{sec:background}, an interpretation is the set of
atoms assigned true. Thus \(J_{K,s}\) records the values in \(s\),
makes exactly the modules in \(K\) active, and assigns false to every
fluent outside \(\mathcal F[K]\).

\emph{Testing whether a symbol is present.}
For a fluent or action symbol \(x\) declared in \(\Sigma^*\), define
\[
  \mathsf{vis}(x)=
  \begin{cases}
    \top,&x\in\Sigma_{\mathsf{core}},\\
    \displaystyle\bigvee_{\{M\in\Modules\mid x\in\Sigma_M\}}
      \mathit{active}_M,&x\notin\Sigma_{\mathsf{core}}.
  \end{cases}
\]
Here \(x\in\Sigma_M\) means that \(M\) declares \(x\), and an empty
disjunction is \(\bot\). The formula \(\mathsf{vis}(x)\) is true in
\(J_{K,s}\) exactly when \(x\) is present in \(\Sigma[K]\).
For a state formula \(\psi\), define
\(
  \mathsf{vis}(\psi)=
    \bigwedge_{f\in\mathsf{Fl}(\psi)}\mathsf{vis}(f).
\)
An empty conjunction is \(\top\). This formula tests whether every
fluent in \(\psi\) is present. As discussed in
Section~\ref{sec:semantics}, \(\mathsf{Fl}(\psi)\) is taken before
simplifying \(\psi\).

\emph{The running example.}
The fixed vocabulary for the calculator example is
\[
  \sigma^*=\{\mathit{taskDone},\mathit{registered}(\mathit{cal}),
    \mathit{answer}(391),\mathit{ready}(\mathit{cal}),
    \mathit{active}_{\mathit{cal}}\}.
\]
For the configurations \(\Config_0=\langle K_0,s_0\rangle\) before
loading and \(\Config_1=\langle K_1,s_1\rangle\) immediately after
loading in Section~\ref{sec:running-transitions},
\(
  J_{K_0,s_0}=\emptyset,\,
  J_{K_1,s_1}=\{\mathit{active}_{\mathit{cal}}\}.
\)
All fluent atoms are false in both interpretations. Nevertheless,
\(\mathsf{vis}(\mathit{registered}(\mathit{cal}))
=\mathit{active}_{\mathit{cal}}\) is false in the first and true in the
second. It distinguishes an absent registration fluent from one that
is present and false. The same test applies to
\(\mathit{answer}(391)\) and \(\mathit{ready}(\mathit{cal})\), whereas
\(\mathsf{vis}(\mathit{taskDone})=\top\).

This gives a fixed vocabulary and an interpretation for each
configuration. It does not yet specify which transitions are possible.
The following causal rules provide that part of the translation.

\subsection{A causal theory for a finite execution}
\label{sec:rat-causal-translation}

To represent \(h\) actions, we use a copy of \(\sigma^*\) at each time
\(0,\ldots,h\). Causal rules relate these copies so that their
interpretations describe successive RAT configurations. We first fix
the action at each step. Corollary~\ref{cor:rat-planning} then allows
the actions to be chosen for planning.

Let \(h\geq0\) be an integer. Fix a causally admissible initial configuration
\(\Config_0=\langle K_0,s_0\rangle\), and
\(\alpha=\langle a_0,\ldots,a_{h-1}\rangle\) with \(a_t\in\mathcal A(\Sigma^*)\).
We use the following notation.
\begin{itemize}[leftmargin=*]
  \item For each atom \(p\in\sigma^*\), its fresh copy \(p^t\)
    records its value at time \(t\), where \(0\leq t\leq h\).
    For a formula \(F\) over \(\sigma^*\), \(\tau_t(F)\) replaces
    each atom \(p\) by \(p^t\) and preserves the Boolean connectives.
    In particular, \(\tau_t(\neg p)=\neg p^t\),
    \(\tau_t(\top)=\top\), and \(\tau_t(\bot)=\bot\).
  \item For each action \(a\in\mathcal A(\Sigma^*)\) and time
    \(0\leq t<h\), a fresh atom \(a^t\) is true exactly when
    \(a\) occurs between time \(t\) and \(t+1\).
    For fluents and actions written with arguments, we place the time
    index before the arguments, as in
    \(\mathit{registered}^t(\mathit{cal})\) and \(\mathit{load}^t(M)\).
  \item A law of module \(M\) applies only when
    \(\mathit{active}_M^t\) holds. To write the same rule for core
    laws, we write \(\mathit{active}_{\mathsf{core}}^t\) for \(\top\).
  \item For each causally admissible configuration
    \(\Config=\langle K,s\rangle\), inactive module
    \(M\in\Modules\setminus K\), assignment
    \(Y\in\mathsf{Init}_M(s)\), and time \(0\leq t<h\), introduce
    a fresh \emph{selector atom} \(q_{\Config,M,Y}^t\).
    It records the choice of \(Y\) when \(M\) is loaded from
    \(\Config\) at time \(t\).
\end{itemize}

The translation below produces a causal theory. Each part adds a specified
set of rules to it. Their bodies test conditions on a candidate
execution, and their heads must causally explain all its atom values.

\begin{definition}[Causal translation of a finite RAT execution]
\label{def:rat-causal-compilation}
Let \(\mathcal R\) be a finite ground RAT system,
\(\Config_0=\langle K_0,s_0\rangle\) a causally admissible initial
configuration, and \(\alpha=\langle a_0,\ldots,a_{h-1}\rangle\) a
sequence of \(h\geq 0\) actions in \(\mathcal A(\Sigma^*)\).
The theory \(\Gamma_h(\mathcal R,\Config_0,\alpha)\) is the union of
the seven sets of ground causal rules in (a)--(g). Its atoms are
\[
  \{p^t\mid p\in\sigma^*,\ 0\leq t\leq h\}
  \ \cup\
  \{a^t\mid a\in\mathcal A(\Sigma^*),\ 0\leq t<h\},
\]
together with the selector atoms \(q_{\Config,M,Y}^t\) for every possible causally admissible configuration
    \(\Config=\langle K,s\rangle\), inactive module
    \(M\in\Modules\setminus K\), assignment
    \(Y\in\mathsf{Init}_M(s)\), and time \(0\leq t<h\).
\begin{enumerate}[leftmargin=*,label=(\alph*)]
  \item \emph{Initial configuration.}
    Fix the initial module set by
    \[
      \mathit{active}_M^0\Leftarrow\top\quad(M\in K_0),
      \qquad
      \neg\mathit{active}_M^0\Leftarrow\top
        \quad(M\in\Modules\setminus K_0).
    \]
    Include \(\tau_0(\ell)\Leftarrow\top\) for every
    \(\ell\in\mathsf{Base}_{K_0}(s_0)\).
    For each \(\ell\in s_0\cap\mathsf{Lit}(\mathcal F^d[K_0])\), include
    \(\bot\Leftarrow\neg\tau_0(\ell)\).
    These constraints require the specified defined-fluent values without
    supplying their causal support, which must come from (c).
    The rules in (f) assign false to fluents absent at time zero.

  \item \emph{Fixed action occurrences.}
    These facts require exactly the action \(a_t\) specified by
    \(\alpha\) at each time \(0\leq t<h\).
    \[
      {a_t}^t\Leftarrow\top,\qquad
      \neg a^t\Leftarrow\top
      \quad(a\in\mathcal A(\Sigma^*)\setminus\{a_t\}).
    \]

  \item \emph{Static causal laws.}
    These rules express within-state causes from the core and the
    modules active at that time.
    For every \(M\in\{\mathsf{core}\}\cup\Modules\), law
    \(\mathsf{caused}\ h_0\ \mathsf{if}\ \psi\) in
    \(L_M^s\), and \(0\leq t\leq h\),
    include
    \[
      \tau_t(h_0)\Leftarrow\tau_t(\psi)\land\mathit{active}_M^t.
    \]

  \item \emph{Dynamic causal laws and executability.}
    These rules express effects and blocked actions using the laws
    active before the action.
    For every \(M\in\{\mathsf{core}\}\cup\Modules\), law
    \(\mathsf{caused}\ \ell\ \mathsf{if}\ \psi\
      \mathsf{after}\ a\land\varphi\) in \(L_M^d\), and \(0\leq t<h\),
    include
    \[
      \tau_{t+1}(\ell)\Leftarrow{}
        \mathit{active}_M^t\land a^t\land\tau_t(\varphi)\land\tau_{t+1}(\psi)
        \land\tau_{t+1}(\mathsf{vis}(\psi))
         \land\tau_{t+1}(\mathsf{vis}(\mathsf{fl}(\ell))).
    \]
    For the same range of \(M\), each law
    \(\mathsf{nonexecutable}\ a\ \mathsf{if}\ \varphi\)
    in \(L_M^x\) and \(0\leq t<h\), include
    \[
      \bot\Leftarrow a^t\land\tau_t(\varphi)\land\mathit{active}_M^t.
    \]
    Here \(\mathit{active}_M^t\) requires the law's module to be active
    before the action. The formula \(\varphi\) is tested at time \(t\),
    and \(\psi\) at time \(t+1\). The two \(\mathsf{vis}\) tests require
    all fluents in \(\psi\) and the fluent of \(\ell\) to be present
    after the action.

  \item \emph{Module changes and available actions.}
    These rules implement \(\mathsf{Next}_K(a)\) and require the
    action to be present before it occurs.
    For each \(M\in\Modules\) and \(0\leq t<h\), include
    \[
    \begin{aligned}
      \mathit{active}_M^{t+1}
        &\Leftarrow \mathit{load}^t(M),\\
      \neg\mathit{active}_M^{t+1}
        &\Leftarrow \mathit{unload}^t(M),\\
      \mathit{active}_M^{t+1}
        &\Leftarrow \mathit{active}_M^t\land
          \mathit{active}_M^{t+1}\land\neg\mathit{unload}^t(M),\\
      \neg\mathit{active}_M^{t+1}
        &\Leftarrow \neg\mathit{active}_M^t\land
          \neg\mathit{active}_M^{t+1}\land\neg\mathit{load}^t(M),\\
      \bot&\Leftarrow \mathit{load}^t(M)\land\mathit{active}_M^t,\\
      \bot&\Leftarrow \mathit{unload}^t(M)\land\neg\mathit{active}_M^t.
    \end{aligned}
    \]
    For every \(a\in\mathcal A(\Sigma^*)\) and \(0\leq t<h\), also include
    \(\bot\Leftarrow a^t\land\neg\tau_t(\mathsf{vis}(a))\).
    The third and fourth rules support the value of \(\mathit{active}_M\) at time
    \(t+1\) when it agrees with the value at time \(t\). Loading
    supplies support for becoming active, and unloading for becoming
    inactive. No rule supports any other change on \(\mathit{active}_M\).

  \item \emph{Inertia and values for absent fluents.}
    For each basic fluent \(f\in\mathcal F^b(\Sigma^*)\), each
    \(\ell\in\{f,\neg f\}\), and \(0\leq t<h\), include
    \[
      \tau_{t+1}(\ell)\Leftarrow
      \tau_t(\ell)\land\tau_{t+1}(\ell)
      \land\tau_t(\mathsf{vis}(f))\land\tau_{t+1}(\mathsf{vis}(f)).
    \]
    For a basic fluent present at both times, these rules support its value
    if it stays unchanged. A changed value instead needs support from a
    dynamic or static causal law.

    For each fluent \(f\in\mathcal F(\Sigma^*)\setminus
    \mathcal F(\Sigma_{\mathsf{core}})\) and \(0\leq t\leq h\), also include
    \[
      \neg f^t\Leftarrow\neg\tau_t(\mathsf{vis}(f)).
    \]
    These false values represent absence, not a value in the RAT state.
    The inertia rule requires \(f\) to be present at both times, so it
    cannot carry such a value into a newly loaded module.

  \item \emph{Initializer selection.}
    These rules choose one assignment permitted by the initializer
    whenever a module is loaded, and require its new basic fluents
    to have those values.
    Enumerate every causally admissible configuration
    \(\Config=\langle K,s\rangle\) of \(\mathcal R\),
    every inactive module \(M\in\Modules\setminus K\), and every time
    index \(t\) with \(0\leq t<h\).
    We call a pair \((\Config,M)\) a \emph{loading context}. It specifies
    the source configuration and the module being loaded.
    For each such context, use the selectors
    \(q_{\Config,M,Y}^t\), where \(Y\in\mathsf{Init}_M(s)\), and define
    \[
e_{\Config,M}^t
        = \bigwedge_{N\in K}\mathit{active}_N^t
        \land\bigwedge_{N\in\Modules\setminus K}\neg\mathit{active}_N^t
        \land\bigwedge_{\ell\in s}\tau_t(\ell) \land\mathit{load}^t(M).
        \]
    Here \(e_{\Config,M}^t\) tests that the source is exactly
    \(\Config\) and this
    step loads \(M\).
The atom \(q_{\Config,M,Y}^t\) records the choice
    of assignment \(Y\). When \(e_{\Config,M}^t\) is true, exactly
    one selector for this context must be true. When it is false, every
    selector for the context must be false.
    Include the following rules, where the assignments \(Y,Y'\) range over
    \(\mathsf{Init}_M(s)\).
    \[
    \begin{aligned}
      q_{\Config,M,Y}^t
        &\Leftarrow q_{\Config,M,Y}^t\land e_{\Config,M}^t,\\
      \neg q_{\Config,M,Y}^t
        &\Leftarrow\neg q_{\Config,M,Y}^t\land e_{\Config,M}^t,\\
      \neg q_{\Config,M,Y}^t
        &\Leftarrow\neg e_{\Config,M}^t,\\
      \bot&\Leftarrow e_{\Config,M}^t\land
        \bigwedge_{Y\in\mathsf{Init}_M(s)}\neg q_{\Config,M,Y}^t,\\
      \bot&\Leftarrow e_{\Config,M}^t\land
        q_{\Config,M,Y}^t\land q_{\Config,M,Y'}^t
        \qquad(Y\ne Y'),\\
      \tau_{t+1}(\ell)&\Leftarrow q_{\Config,M,Y}^t
        \qquad(\ell\in Y).
    \end{aligned}
    \]
    The fourth line requires at least one choice and is included once per
    context, even if \(\mathsf{Init}_M(s)=\emptyset\). In that case
    its empty conjunction is \(\top\), so it becomes
    \(\bot\Leftarrow e_{\Config,M}^t\) and forbids loading in that context.
\end{enumerate}
\end{definition}

\emph{Translation of the running example.}
Take \(\mathcal R_{\times}\) from Section~\ref{sec:running-example},
\(\Config_0=\langle\emptyset,\{\neg\mathit{taskDone}\}\rangle\), and
\(\alpha=\langle\mathit{load}(\mathit{cal})\rangle\).
The initializer supplies the single assignment
\[
  Y=\{\neg\mathit{registered}(\mathit{cal}),
       \neg\mathit{answer}(391)\}.
\]
The rules in (a), (b), and (f) fix the atom values at time zero and
the action occurrence. The rules in (e) make
\(\mathit{active}_{\mathit{cal}}^1\) true, and (g) makes
\(q_{\Config_0,\mathit{cal},Y}^0\) true and every other selector false.
After substituting these values and removing the resulting trivial
rules, the causal theory for the four fluent atoms at time one is
\[
\begin{aligned}
  \neg\mathit{taskDone}^1
    &\Leftarrow\neg\mathit{taskDone}^1,\\
  \neg\mathit{registered}^1(\mathit{cal})
    &\Leftarrow\top,\\
  \neg\mathit{answer}^1(391)
    &\Leftarrow\top,\\
  \mathit{ready}^1(\mathit{cal})
    &\Leftarrow\mathit{registered}^1(\mathit{cal})
      \land\neg\mathit{taskDone}^1,\\
  \neg\mathit{ready}^1(\mathit{cal})
    &\Leftarrow\neg\mathit{registered}^1(\mathit{cal})
      \lor\mathit{taskDone}^1.
\end{aligned}
\]
The first rule gives inertia for the core fluent.
The next two rules give the initializer values, and the last two
are the calculator's static laws. The calculator contributes no
dynamic law to this step because it is inactive before loading.
The only causally explained interpretation assigns false to all four
fluent atoms. Together with \(\mathit{active}_{\mathit{cal}}^1\),
it represents the configuration
\[
  \Config_1=\bigl\langle\{\mathit{cal}\},
    \{\neg\mathit{taskDone},\neg\mathit{registered}(\mathit{cal}),
      \neg\mathit{answer}(391),\neg\mathit{ready}(\mathit{cal})\}\bigr\rangle
\]
derived in Section~\ref{sec:running-transitions}.

The initialization rules in (g) enumerate source configurations and
assignment choices, so this finite translation may be exponentially
large. For initializers that explicitly list assignments independently
of the input state, the implementation in Section~\ref{sec:evaluation}
uses selectors indexed only by module, assignment, and time. The
Supplementary Material explains why this smaller encoding represents
the same executions for that fragment.

To state correctness, we define how to recover a RAT execution from an
interpretation of the translated theory. Activation atoms give the module
set at each time, and literals over the resulting visible fluent set give
its state. For an interpretation \(J\) of the
translated signature, define
\begin{align}
  K_t(J)&=\{M\in\Modules\mid J\models\mathit{active}_M^t\},
  \label{eq:decode-active-set}\\
  s_t(J)&=\{\ell\in\mathsf{Lit}(\mathcal F[K_t(J)])
             \mid J\models\tau_t(\ell)\}.
  \label{eq:decode-state}
\end{align}

\begin{theorem}[RAT--causal-theory correspondence]
\label{thm:rat-causal-correspondence}
For the finite RAT system \(\mathcal R\), a causally admissible initial
configuration \(\Config_0\), action sequence
\(\alpha=\langle a_0,\ldots,a_{h-1}\rangle\), and configurations
\(\Config_1,\ldots,\Config_h\),
\[
  \Config_0\trans{a_0}_{\mathrm{RAT}}\Config_1\cdots
  \trans{a_{h-1}}_{\mathrm{RAT}}\Config_h
\]
iff \(\Gamma_h(\mathcal R,\Config_0,\alpha)\) has a causally explained
interpretation \(J\) with
\[
  \Config_t=\langle K_t(J),s_t(J)\rangle
  \qquad(0\leq t\leq h).
\]
\end{theorem}

\begin{proof}[Proof sketch]
We compare the rules that explain fluent values at each time with the
single-step theory in Equation~\eqref{eq:step-causal-theory}.
A literal-headed causal reduct explains an interpretation exactly when
it contains the literal true at each atom, contains no false literal,
and does not contain \(\bot\).
For a causal explanation \(J\) of \(\Gamma_h\), abbreviate its decoded
module set and state as \(K_t=K_t(J)\) and \(s_t=s_t(J)\).

\begin{enumerate}[leftmargin=*,label=(\roman*)]
  \item \emph{Recover the initial configuration and action labels.}
    The rules in (a) fix the values of \(\Config_0\), while (c) explains its
    defined-fluent values. The rules in (f) set absent fluents to false.
    Thus decoding gives \(\Config_0\). The facts in (b) fix every action
    label. The rules in (d)--(e) exclude blocked or absent actions and
    require \(K_{t+1}=\mathsf{Next}_{K_t}(a_t)\).

  \item \emph{Recover each transition.}
    Suppose the decoded \(\Config_t=\langle K_t,s_t\rangle\) is causally
    admissible. If \(a_t\) loads \(M\), the rules in (g) choose one
    \(Y\in\mathsf{Init}_M(s_t)\). Otherwise put \(Y=\emptyset\).
    On the time-\((t+1)\) copies of fluents in \(\mathcal F[K_{t+1}]\), the heads of
    \(\Gamma_h^J\) are exactly those of
    \(
      {\bigl(T_{K_t,s_t,a_t,K_{t+1},Y}\bigr)}^{I_{s_{t+1}}},
    \)
    with time indices added. Here
    \(I_{s_{t+1}}=s_{t+1}\cap\mathcal F[K_{t+1}]\) is the classical
    interpretation of the decoded state. The rules in (c), (d), (f), and
    (g) give, respectively, static rules, dynamic rules, inertia, and
    initializer facts. Hence this single-step theory explains
    \(s_{t+1}\). Only static rules can explain its defined-fluent
    values, so \(s_{t+1}\) is also causally admissible under \(K_{t+1}\).
    Together with (i), these are the transition requirements of
    Definition~\ref{def:rat-transition}. Induction gives the execution.

  \item \emph{Represent a given execution.}
    Conversely, assign the activation and present-fluent values from
    each \(\Config_t\), set absent fluents to false, and fix the action
    labels to \(\alpha\). At every loading step, choose an initializer
    assignment used by that transition and make its selector true.
    Set every other selector false, including all selectors at steps
    without loading. The same comparison with the single-step reduct shows that
    (a)--(g) supply exactly the assigned literal for every atom and no
    \(\bot\). The resulting interpretation is therefore the unique
    model of its reduct, and it decodes to the given execution.
\end{enumerate}
\end{proof}

If no action loads or unloads a module, the construction represents
executions of the one ordinary action description \(\Dof K\).

\begin{corollary}[Executions with a fixed module set]
\label{cor:fixed-causal-reduction}
Let \(\Config_0=\langle K,s_0\rangle\) be causally admissible in the RAT system
\(\mathcal R\), and let
\(\alpha=\langle a_0,\ldots,a_{h-1}\rangle\) contain only actions in
\(\mathcal A[K]\setminus\mathcal A^{\mathsf r}\).
Every causal explanation \(J\) of
\(\Gamma_h(\mathcal R,\Config_0,\alpha)\) has \(K_t(J)=K\) for all
\(0\leq t\leq h\). As \(J\) ranges over these explanations, the decoded
state sequences are exactly the sequences, i.e., 
\[
  s_0(J)\trans{a_0}_{\Dof K}s_1(J)\cdots
     \trans{a_{h-1}}_{\Dof K}s_h(J)
\]
under the ordinary causal transition relation of
Section~\ref{sec:action-theory-foundation}.
\end{corollary}

\begin{proof}[Proof sketch]
There is no loading or unloading, so the rules in (e) keep \(K\) fixed
and all initializer selectors are false. Theorem~\ref{thm:rat-causal-correspondence}
then gives exactly the RAT executions at this fixed \(K\).
Proposition~\ref{prop:fixed-theory} identifies each of their transitions
with the ordinary transition under \(\Dof K\).
\end{proof}

So far, the input specifies every action. Planning instead leaves the
actions to be chosen. We replace only the occurrence facts, keeping all
rules that determine which transitions are permitted.

\begin{corollary}[Bounded planning encoding]
\label{cor:rat-planning}
Let \(\mathcal R\) be a finite ground RAT system with a causally
admissible initial configuration \(\Config_0\), and let \(h\geq0\)
be an integer specifying the number of action steps.
Replace (b) of Definition~\ref{def:rat-causal-compilation} by, for every
\(a\in\mathcal A(\Sigma^*)\) and \(0 \leq t<h\),
\[
  a^t\Leftarrow a^t,\qquad
  \neg a^t\Leftarrow\neg a^t,
\]
and the exactly-one constraints, where \(a,b\) range over
\(\mathcal A(\Sigma^*)\),
\[
  \bot\Leftarrow\bigwedge_{a\in\mathcal A(\Sigma^*)}\neg a^t,
  \qquad
  \bot\Leftarrow a^t\land b^t\quad(a\ne b).
\]
Call the resulting theory
\(\Gamma_h^{\mathsf{plan}}(\mathcal R,\Config_0)\).
Its decoded causal explanations are exactly all length-\(h\) RAT
executions from \(\Config_0\), including the action at each step.
\end{corollary}

\begin{proof}
For each action atom, the two occurrence rules support the truth value
assigned by the candidate interpretation. The constraints require exactly
one action atom to be true at each step. Each resulting sequence \(\alpha\)
is covered by Theorem~\ref{thm:rat-causal-correspondence}. Action visibility,
nonexecutability tests, and all successor conditions remain unchanged.
\end{proof}

\subsection{Translation to answer set programs}
\label{sec:rat-asp-translation}

The theory \(\Gamma_h\) is now an ordinary nonmonotonic causal theory.
Every head is a literal or \(\bot\), so we can apply the corresponding
translation of Ferraris et al.~\cite[Sections~2.4 and~5.2]{ferrarisEtAl2012}.
We write the translation as a program with nested bodies, using
the semantics in Section~\ref{sec:asp-foundation}. The Supplementary
Material gives the conversion to normal
rules using auxiliary atoms and expands the proofs.

For a finite literal-headed causal theory \(T\) over atoms \(\sigma\),
introduce two ASP atoms \(p^+\) and \(p^-\) for each \(p\in\sigma\).
They represent its true and false values. In particular, \(p^-\) is an
ordinary ASP atom, not a negated expression. Write each Boolean body \(G\)
with negations only on atoms, and let \([G]\) replace \(p\) by \(p^+\)
and \(\neg p\) by \(p^-\), preserving \(\top,\bot,\land,\lor\).
The same notation applies to a literal head.
The nested program \(\mathsf{Tr}[T]\) contains
\[
\begin{aligned}
  [\ell]&\leftarrow\mathsf{not}\,\mathsf{not}\,[G]
      &&\text{for each }\ell\Leftarrow G\text{ in }T,\\
  &\leftarrow[G]
      &&\text{for each }\bot\Leftarrow G\text{ in }T.
\end{aligned}
\]
It also contains, for every \(p\in\sigma\),
\begin{equation}
  \leftarrow p^+,p^-,
  \qquad
  \leftarrow\mathsf{not}\ p^+,\mathsf{not}\ p^-.
  \label{eq:asp-two-valuedness}
\end{equation}
The body \([G]\) may contain conjunctions and disjunctions, and
\(\mathsf{not}\,\mathsf{not}\,[G]\) applies default negation twice
to that entire expression. These bodies use the nested syntax, not the
normal-rule syntax.
Equation~\eqref{eq:asp-two-valuedness} requires exactly one truth value
but does not provide support for either one.
Double default negation tests the body in the candidate answer set.
The rule becomes a fact in the reduct when that test is true, just as
a true causal body contributes its head to the causal reduct.
Replacing \(\mathsf{not}\,\mathsf{not}\,[G]\) by \([G]\) would be
incorrect. For example, \(p\Leftarrow p\) explains \(p\), whereas
\(p\leftarrow p\) alone does not derive \(p\).

For a classical interpretation \(J\subseteq\sigma\), the corresponding
set of ASP atoms is
\[
  J^{\pm}=
    \{p^+\mid p\in J\}\cup\{p^-\mid p\in\sigma\setminus J\}.
\]
The following is the Boolean literal-headed case of the cited result.

\begin{theorem}[Causal-theory--ASP correspondence~{\cite{ferrarisEtAl2012}}]
\label{thm:causal-asp-correspondence}
For every finite Boolean literal-headed causal theory \(T\) and every
classical interpretation \(J\) over its signature,
\[
  J\in\mathsf{CE}(T)
  \quad\text{iff}\quad
  J^{\pm}\in\mathsf{AS}(\mathsf{Tr}[T]).
\]
\end{theorem}

\begin{proof}[Proof sketch]
At \(J^{\pm}\), the translated nonconstraint rules give precisely
the signed heads of \(T^J\) as facts in the ASP reduct.
The constraints reject conflicting or incomplete truth values and
violated causal constraints. Minimality requires each chosen value to
be supported by a fact. Thus the answer-set test holds exactly when
\(T^J\) uniquely determines \(J\).
\end{proof}

Apply this translation to \(\Gamma_h\), giving
\[
  \Pi_h(\mathcal R,\Config_0,\alpha)
    =\mathsf{Tr}[\Gamma_h(\mathcal R,\Config_0,\alpha)].
\]
For an answer set \(A\), let \(J_A=\{p\mid p^+\in A\}\) be its classical
interpretation. Equation~\eqref{eq:decode-active-set}
and Equation~\eqref{eq:decode-state} then recover the modules and state at each time.

\begin{corollary}[RAT executions represented by answer sets]
\label{cor:asp-path}
Let \(\mathcal R\) be a finite ground RAT system,
\(\Config_0\) a causally admissible initial configuration, and
\(h\in\{0,1,\ldots\}\) the number of steps. Let
\(\alpha=\langle a_0,\ldots,a_{h-1}\rangle\) be a sequence of \(h\)
ground actions in \(\mathcal A(\Sigma^*)\). The sequences
\[
  \bigl\langle
    \langle K_0(J_A),s_0(J_A)\rangle,\ldots,
    \langle K_h(J_A),s_h(J_A)\rangle
  \bigr\rangle,
  \qquad A\in\mathsf{AS}(\Pi_h(\mathcal R,\Config_0,\alpha)),
\]
are exactly the RAT executions from \(\Config_0\) with action sequence
\(\alpha\), where \(J_A=\{p\mid p^+\in A\}\) is the classical
interpretation recovered from the answer set \(A\).
Translating \(\Gamma_h^{\mathsf{plan}}\) instead gives all
length-\(h\) executions from \(\Config_0\), with their selected actions.
\end{corollary}

\begin{proof}
Combine Theorem~\ref{thm:causal-asp-correspondence} with
Theorem~\ref{thm:rat-causal-correspondence} and
Corollary~\ref{cor:rat-planning}.
\end{proof}

Auxiliary-atom translations can eliminate nested expressions while
preserving answer sets on the original atoms~\cite{pearceEtAl2002}.
For the programs used here, auxiliary atoms allow us to express every
translated rule as normal rules and constraints, without disjunctive heads.
The resulting programs can be evaluated by ASP solvers such as
clasp, with grounding supplied by gringo~\cite{gebserEtAl2019}.
The Supplementary Material gives the construction and
proves that removing the auxiliary atoms recovers exactly the answer
sets of \(\mathsf{Tr}[T]\).

\section{Experiments with harness applications}
\label{sec:evaluation}

We first compare RAT descriptions with recorded executions of DeepSeek
Harness, Pi, and OpenHands. In these local tests, the finite RAT models
predict the observed outcomes of tool calls and component changes in the
three official implementations.
We also examine which
properties from Section~\ref{sec:rat-properties} apply. We then use
generated RAT models to test configuration computation and planning
with ASP, measuring how grounding and search costs grow and when the
implementation reaches its time or memory limit.
The Supplementary Material lists the source repositories, tested commits,
package versions, and test inputs.
OpenAI Codex assisted with preparing and revising the instance-specific
ASP encodings, test scenarios, and queries. Section~S8 of the Supplementary
Material describes the scope of this assistance and the verification
procedures.

\subsection{Three harness applications}
\label{sec:harness-applications}
\label{sec:deepseek-case}
\label{sec:harness-comparison}

A \emph{workflow} here is a sequence of related tool calls and component
changes. We give a complete RAT description of a DeepSeek file-editing
workflow. Three Pi workflows and one OpenHands workflow illustrate other
tool changes. For all examples, the Supplementary Material provides the
complete RAT descriptions, ASP code, recorded test sequences, and
detailed comparisons.

In the following tests, each counted \emph{operation} is one attempt to call a tool, change a
component, or write a file externally. A successful operation is carried
out by the tested implementation. A rejected attempt is refused and
leaves the represented configuration unchanged. From the configuration
before each attempt, we compute the RAT successors by direct causal
evaluation and by ASP. Each method must give exactly the next
configuration obtained from the recorded observations for a successful
operation, and no successor for a rejected attempt.
Agreement on a rejection therefore counts as a correct prediction.

\subsubsection{DeepSeek Harness: loading and unloading the observation-policy plugin}

\emph{The workflow.}
An agent reads and changes an existing file through DeepSeek's
\emph{file-tool plugin}. It provides three operations.
A \emph{read} returns the file contents without changing them.
A \emph{write} overwrites the entire contents with the supplied text.
An \emph{edit} changes specified matching text within the file and
leaves the rest unchanged.

DeepSeek also provides an optional \emph{observation-policy plugin}.
This is a separate plugin that checks whether the agent has observed
the file's current version before allowing a write or edit.\footnote{See
\url{https://github.com/deepseek-ai/deepseek-harness}, under
\texttt{packages/fs}: \texttt{tool-fs} is the file-tool plugin and
\texttt{fs-observation-policy} is the observation-policy plugin. The Supplementary
Material records the tested commit.}
Reading the file establishes that observation. A successful write or edit
updates the observation to the version it produces. A change made
outside these tools makes the previous observation outdated.

Unloading the observation-policy plugin permits writes without this
check. Reloading the observation-policy plugin starts a new observation
record, so the agent must observe the file again. We use RAT to describe
how loading and unloading the observation-policy plugin changes which
writes are permitted.

\emph{The RAT representation.}
We represent the observation-policy plugin by the single module
\(\mathit{policy}\). The core records the file contents and describes
the file operations, including those in the file-tool plugin.

Table~\ref{tab:deepseek-sequence} gives the recorded test sequence.
Before Stage~1, the observation-policy plugin has not been loaded.
The file \texttt{document.txt} contains one line, initially
\texttt{\textquotedbl{}alpha=1\textquotedbl{}}. To test a change made outside the agent's tool
calls, the test program performs an \emph{external write}. This operation
overwrites the entire file directly through the filesystem, without
calling the file-tool plugin. 

\begin{table}[!htb]
\centering
\caption{Recorded DeepSeek sequence. The filesystem service and file-tool
plugin remain available throughout. Initially, the observation-policy
plugin is absent and the file contains \texttt{\textquotedbl{}alpha=1\textquotedbl{}}.}
\label{tab:deepseek-sequence}
\begin{tabularx}{\linewidth}{@{}r>{\raggedright\arraybackslash}X>{\raggedright\arraybackslash}X@{}}
\toprule
Stage & Operations & Observed result \\
\midrule
1 & Load the observation-policy plugin, then write \texttt{\textquotedbl{}alpha=2\textquotedbl{}}. &
  Loading succeeds. Write rejected because the file has not been observed.
  File remains \texttt{\textquotedbl{}alpha=1\textquotedbl{}}. \\
2 & Read, then edit \texttt{\textquotedbl{}alpha=1\textquotedbl{}} to \texttt{\textquotedbl{}alpha=2\textquotedbl{}}. &
  Both calls succeed. File becomes \texttt{\textquotedbl{}alpha=2\textquotedbl{}}. \\
3 & External write of \texttt{\textquotedbl{}alpha=external\textquotedbl{}},
  then write \texttt{\textquotedbl{}alpha=3\textquotedbl{}}. &
  External write succeeds, making the file
  \texttt{\textquotedbl{}alpha=external\textquotedbl{}}. The subsequent write is rejected as
  the observation is outdated. \\
4 & Read again, then write \texttt{\textquotedbl{}alpha=3\textquotedbl{}}. &
  Both calls succeed. File becomes \texttt{\textquotedbl{}alpha=3\textquotedbl{}}. \\
5 & Unload the observation-policy plugin, perform an external write of
  \texttt{\textquotedbl{}alpha=external-again\textquotedbl{}}, then write \texttt{\textquotedbl{}alpha=4\textquotedbl{}}. &
  Unloading succeeds. External write makes the file
  \texttt{\textquotedbl{}alpha=external-again\textquotedbl{}}. The subsequent write succeeds without
  another read, making the file \texttt{\textquotedbl{}alpha=4\textquotedbl{}}. \\
6 & Reload the observation-policy plugin, then write \texttt{\textquotedbl{}alpha=5\textquotedbl{}}. &
  Reloading clears the observation record. Write rejected.
  File remains \texttt{\textquotedbl{}alpha=4\textquotedbl{}}. \\
7 & Read, then write \texttt{\textquotedbl{}alpha=5\textquotedbl{}}. &
  Both calls succeed. File becomes \texttt{\textquotedbl{}alpha=5\textquotedbl{}}. \\
\bottomrule
\end{tabularx}
\end{table}

The following description defines fluents, actions, laws, and
initial values used for this case.

\begin{samepage}
\emph{(1) Representing the file contents.}
We use seven basic fluents \(c_0,\ldots,c_6\) for the file contents
used in the test. Fluent \(c_i\) is true when the file contains the line
\texttt{\textquotedbl{}alpha=value\textquotedbl{}}, where \texttt{value} is entry \(i\) of the list
\(
  (\texttt{1},\ \texttt{2},\ \texttt{3},\ \texttt{4},\ \texttt{5},\
    \texttt{external},\ \texttt{external-again}),
\)
numbered from zero. For example, \(c_0\) means that the file contains
\texttt{\textquotedbl{}alpha=1\textquotedbl{}}. Static constraints below require exactly one
of these fluents to be true. The complete assignment for contents
\(c_i\) is
\[
  Y_i=\{c_i\}\cup\{\neg c_j\mid 0\leq j\leq6,\ j\ne i\}
  \qquad(0\leq i\leq6).
\]
The executable model uses these same seven indicators.
\end{samepage}

\emph{(2) Actions, fluents, and modules.}
The action \(\mathit{read}\) reads the file without changing its contents.
The action \(\mathit{write}_i\) uses the file-tool plugin to overwrite
the entire contents with the text represented by \(c_i\).
The action \(\mathit{external}_i\) is an external write of that same
text directly to disk, without calling the file-tool plugin. The action
\(\mathit{edit}_0\) changes the unique occurrence of
\texttt{\textquotedbl{}alpha=1\textquotedbl{}} to \texttt{\textquotedbl{}alpha=2\textquotedbl{}} and leaves the rest unchanged.
The ordinary
action set is
\[
  A_0=\{\mathit{read},\mathit{edit}_0\}
       \cup\{\mathit{write}_i,\mathit{external}_i\mid i=0,\ldots,6\}.
\]
We use \(A_{\mathsf{file}}=\{\mathit{read},\mathit{edit}_0\}
\cup\{\mathit{write}_i\mid i=0,\ldots,6\}\) for the file-tool calls and
\(A_{\mathsf{mut}}=A_{\mathsf{file}}\setminus\{\mathit{read}\}\)
for those that change the file.

The private basic fluent \(\mathit{seen}\) of module \(\mathit{policy}\)
records whether the agent has observed the file since the
observation-policy plugin was loaded.
The private basic fluent \(\mathit{fresh}\) records whether that
observation is still current. 
The RAT system and its signatures are
\[
\begin{aligned}
  \mathcal R_{\mathsf{fs}}
    &=\left\langle\{\mathit{policy}\},\mathcal D_{\mathsf{core}},
       \{\langle\mathcal D_{\mathit{policy}},
                  \mathsf{Init}_{\mathit{policy}}\rangle\}\right\rangle,\\
  \Sigma_{\mathsf{core}}
    &=\left\langle \{c_0,\ldots,c_6\},\emptyset,
       A_0\cup\{\mathit{load}(\mathit{policy}),
                 \mathit{unload}(\mathit{policy})\}\right\rangle,\\
  \Sigma_{\mathit{policy}}
    &=\left\langle\{\mathit{seen},\mathit{fresh}\},
       \emptyset,A_0\right\rangle.
\end{aligned}
\]
For \(X\in\{\mathsf{core},\mathit{policy}\}\), let
\(\mathcal D_X=\langle\Sigma_X,L_X^s,L_X^d,L_X^x\rangle\).
We give all the laws below.

\emph{(3) Laws for file operations.}
The static laws in \(L_{\mathsf{core}}^s\) require exactly one
file content.
\[
\begin{aligned}
  &\mathsf{caused}\ \bot\ \mathsf{if}\
     \neg c_0\land\cdots\land\neg c_6,\\
  &\mathsf{caused}\ \bot\ \mathsf{if}\ c_i\land c_j
     \qquad(0\leq i<j\leq6).
\end{aligned}
\]
The dynamic laws in \(L_{\mathsf{core}}^d\) assign every literal
in \(Y_i\) when a write sets the file to \(c_i\). The edit action
changes \(c_0\) to \(c_1\).
\[
\begin{aligned}
  &a\ \mathsf{causes}\ \ell
    &&\bigl(i=0,\ldots,6,\ a\in\{\mathit{write}_i,\mathit{external}_i\},
                      \ \ell\in Y_i\bigr),\\
  &\mathit{edit}_0\ \mathsf{causes}\ \ell\ \mathsf{if}\ c_0
    &&(\ell\in Y_1).
\end{aligned}
\]
The core has one nonexecutability law. The fixed edit arguments
require content \(c_0\).
\[
  \mathsf{nonexecutable}\ \mathit{edit}_0\ \mathsf{if}\ \neg c_0.
\]

\emph{(4) Laws and initial values of the observation-policy plugin.}
The module \(\mathit{policy}\) has no static laws, so \(L_{\mathit{policy}}^s=\emptyset\).
Its dynamic laws in \(L_{\mathit{policy}}^d\) are
\[
\begin{aligned}
  &a\ \mathsf{causes}\ \ell
    &&(a\in A_{\mathsf{file}},\
        \ell\in\{\mathit{seen},\mathit{fresh}\}),\\
  &\mathit{external}_i\ \mathsf{causes}\ \neg\mathit{fresh}
    &&(i=0,\ldots,6).
\end{aligned}
\]
A successful file-tool call records the current version, while an
external write makes that observation outdated. The laws in \(L_{\mathit{policy}}^x\)
are
\[
  \mathsf{nonexecutable}\ a\ \mathsf{if}\
    \neg\mathit{seen}\lor\neg\mathit{fresh}
    \qquad(a\in A_{\mathsf{mut}}).
\]
For every input partial valuation \(v\), the initializer is
\[
  \mathsf{Init}_{\mathit{policy}}(v)
    =\bigl\{\{\neg\mathit{seen},\neg\mathit{fresh}\}\bigr\}.
\]

\emph{(5) Initial configuration.}
The initial configuration is
\[
  \Config^{\mathsf{fs}}_0=\langle\emptyset,Y_0\rangle.
\]
The file contains \texttt{\textquotedbl{}alpha=1\textquotedbl{}},
and the observation-policy plugin is absent. This is the configuration
at the start of Table~\ref{tab:deepseek-sequence}. Loading or unloading
\(\mathit{policy}\) changes the restrictions on the core file actions,
not the availability of the file-tool plugin itself.

\emph{Why the model gives the observed results.}
Loading or reloading the observation-policy plugin sets
\(\mathit{seen}\) and \(\mathit{fresh}\) false, blocking the writes
in Stages~1 and~6. Reading sets both true, permitting the edit in
Stage~2 and the write in Stage~7. The external write in Stage~3 makes
\(\mathit{fresh}\) false, blocking the subsequent write. In Stage~5,
unloading the observation-policy plugin removes this blocking law,
so the write succeeds even after an external write.
Neither loading nor unloading the observation-policy plugin changes
the file contents.

We compare four recorded DeepSeek sequences with the corresponding RAT
descriptions in the Supplementary Material. One extends
Table~\ref{tab:deepseek-sequence} with setup, file-tool removal, and a
rejected read; another tests filesystem loss and restoration. The remaining
two load the observation-policy and file-tool plugins in opposite orders. DeepSeek Harness carries out 31 of the 39 attempted
operations and refuses eight calls. Direct causal evaluation and ASP
each correctly predict every execution or rejection and the resulting
configurations, including the file contents.

\subsubsection{Pi: disabling tools, planning mode, and internal calls}
We use three Pi workflows to distinguish disabling a tool, restricting
file changes, and calling a tool that is not listed for direct use by
the language model. Each workflow has its own RAT model.
\begin{enumerate}[leftmargin=*]
  \item \emph{Disabling a registered tool.}
    Pi's official dynamic-tool example registers an echo tool that
    returns a message containing the supplied text. The test calls the
    tool, disables it, and enables it again. Direct calls are rejected while
    the tool is disabled, but its registered definition
    remains.\footnote{See \url{https://github.com/earendil-works/pi},
    \path{packages/coding-agent/examples/extensions/dynamic-tools.ts}.
    The Supplementary Material records the tested revision.}

    In RAT, registration loads module \(\mathit{echo}\), whose private
    basic fluent \(\mathit{echoEnabled}\) is initialized true.
    Disabling sets this fluent false, and enabling sets it true.
    The action \(\mathit{echo}\) represents a call to the tool.
    The module contains the nonexecutability law
    \[
      \mathsf{nonexecutable}\ \mathit{echo}\ \mathsf{if}\
        \neg\mathit{echoEnabled}.
    \]
    The module stays loaded. Disabling the tool changes a fluent,
    rather than removing the action from the active description.

  \item \emph{Restricting file changes in planning mode.}
    Pi's planning-mode extension permits reading and searching a task
    file but rejects a write and a shell command that would create
    another file. After leaving that mode, a write updates the task
    file.\footnote{See \url{https://github.com/earendil-works/pi},
    \path{packages/coding-agent/examples/extensions/plan-mode/index.ts}.}

    In RAT, module \(\mathit{planPolicy}\) remains loaded.
    Its basic fluent \(\mathit{planMode}\) is true in planning mode.
    Its nonexecutability laws then block the file-changing actions.
    For example, the action \(\mathit{writeDone}\) writes the updated
    task file and is blocked by
    \[
      \mathsf{nonexecutable}\ \mathit{writeDone}\ \mathsf{if}\
        \mathit{planMode}.
    \]
    Leaving planning mode sets \(\mathit{planMode}\) false, permitting
    the write without unloading \(\mathit{planPolicy}\).

  \item \emph{Calling a tool through another tool.}
    We provide a local MCP search server and a query tool for this test.
    The query tool calls the search tool through Pi's MCP client.
    Pi does not list the search tool among the tools the language model
    can call directly. Instead, its \texttt{codemode} setting permits
    the internal call. The internal call succeeds, while a direct search
    call is rejected. If the server withdraws the search tool, the
    internal call also fails. Restoring the tool permits the internal
    call again.\footnote{The test follows Pi's MCP integration tests at
    \url{https://github.com/earendil-works/pi},
    \path{packages/coding-agent/test/suite/agent-session-mcp.test.ts}.}

    In RAT, module \(\mathit{search}\) provides three ground search
    actions for two fixed query strings. We write
    \(\mathit{directSearch}\) for the direct query using the first string,
    and \(\mathit{internalSearch}_1\) and \(\mathit{internalSearch}_2\) for the
    internal queries using the first and second strings, respectively.
    Its private basic fluent \(\mathit{declared}\) records whether the
    search tool is listed for direct calls.
    For every input partial valuation \(v\), initialization and the
    direct-call restriction are given by
    \[
      \begin{gathered}
        \mathsf{Init}_{\mathit{search}}(v)
          =\{\{\neg\mathit{declared}\}\},\\
        \mathsf{nonexecutable}\ \mathit{directSearch}\
          \mathsf{if}\ \neg\mathit{declared}.
      \end{gathered}
    \]
    Neither internal action has this restriction. Thus loading
    \(\mathit{search}\) permits internal queries but not direct ones.
    Server withdrawal is represented by
    \(\mathit{unload}(\mathit{search})\), which removes all three
    search actions. Reloading restores the internal calls.
\end{enumerate}
We test the echo-tool, planning-mode, and direct and internal search
workflows in Pi and in the corresponding RAT descriptions. The three
tests contain six, eight, and ten attempted operations, respectively.
Pi carries out 18 operations and refuses six calls. Direct causal
evaluation and ASP each correctly predict every execution or rejection
and the resulting tool status and file contents.

\subsubsection{OpenHands: removing a tool while retaining a report}
\emph{The workflow.}
A task may use a temporary tool and save its result in a file. Removing
the tool should prevent further calls, but need not delete the saved
result. We test these two effects with an OpenHands MCP server that can
add and remove an arithmetic tool during a session. The tool is named
\texttt{extra} in the upstream test and returns twice its integer
input.\footnote{See \url{https://github.com/OpenHands/software-agent-sdk},
\path{tests/sdk/mcp/test_mcp_tool_list_changed.py}.
The Supplementary Material records the tested revision.}

Our test creates \texttt{REPORT.txt}, registers this tool, and calls it
with input \(21\), obtaining \(42\). A separate call to OpenHands'
file editor writes that result into the report.\footnote{The workload
combines MCP tools with native file tools, following the pattern in
\url{https://github.com/OpenHands/software-agent-sdk},
\path{examples/01_standalone_sdk/07_mcp_integration.py}.}
The server then removes the arithmetic tool. Calling its previously
saved SDK definition reaches the server and fails, while reading the
report still returns the saved result. Registering the tool again
allows another successful call.

\emph{The RAT representation.}
The core's basic fluents \(\mathit{reportExists}\) and
\(\mathit{reportHasResult}\) record whether the report exists and
contains the result. Registration loads module \(\mathit{extra}\).
The ground action \(\mathit{extra21}\) represents the arithmetic call
with the fixed input \(21\). The module's private basic fluent
\(\mathit{resultReturned}\) records whether this call has returned
\(42\) since the module was last loaded. It is initially false and
becomes true when \(\mathit{extra21}\) occurs. For every input partial
valuation \(v\), the signature, initializer, and effect law are
\[
  \begin{gathered}
    \Sigma_{\mathit{extra}}
      =\langle\{\mathit{resultReturned}\},\emptyset,
        \{\mathit{extra21}\}\rangle,\\
    \mathsf{Init}_{\mathit{extra}}(v)
      =\{\{\neg\mathit{resultReturned}\}\},\\
    \mathit{extra21}\ \mathsf{causes}\ \mathit{resultReturned}.
  \end{gathered}
\]
This law records the returned result for the fixed input, rather than
implementing integer multiplication. A separate file-editor action
makes \(\mathit{reportHasResult}\) true. The file editor accepts supplied
text, so the model does not require the arithmetic call before writing.

\emph{Why the report remains available.}
Unloading \(\mathit{extra}\) removes its action and the private fluent
\(\mathit{resultReturned}\). Inertia preserves both core report fluents,
so the saved report can still be read. Reloading restores the action and
initializes \(\mathit{resultReturned}\) to false. Unlike the DeepSeek observation-policy plugin example,
this example removes an action rather than changing when it is permitted.
We compare the registered tools after each change notification has been
processed. The model does not describe notification delivery.

We test the same sequence in OpenHands and in the corresponding RAT
description. It contains 13 attempts at tool calls and registration
changes, including writing and reading the report. OpenHands carries
out 12 and refuses the call to the removed tool. Direct causal evaluation
and ASP each correctly predict every execution or rejection and the
resulting configurations, including the record of the returned result
and the saved report.

\subsubsection{Using the RAT properties in these examples}
The properties in Section~\ref{sec:rat-properties} apply to the RAT
descriptions above as follows. The Supplementary Material gives the
checks of their conditions.
\begin{itemize}[leftmargin=*]
  \item \emph{Predicting one next configuration.}
    The DeepSeek file-editing model, the three Pi models, and the
    OpenHands report model satisfy Proposition~\ref{prop:unique-successor}.
    Each initializer permits one assignment, and the laws satisfy the
    required dependency order. Thus each enabled action has at most
    one successor from a given configuration. Exhaustive evaluation
    of these finite models also confirms that every enabled action
    has a successor at every causally admissible configuration. A controller
    can therefore compute the next configuration predicted by the model
    before executing an enabled action.
  \item \emph{Adding a tool without changing existing behavior.}
    The OpenHands module \(\mathit{extra}\) adds the action
    \(\mathit{extra21}\) and the private fluent \(\mathit{resultReturned}\).
    Its effect law changes only this private fluent. It satisfies the
    independent-extension conditions of Theorem~\ref{thm:conservative}.
    In this model, loading \(\mathit{extra}\) preserves the report state
    and the possible effects of the existing file-editor actions on the
    two report fluents.
  \item \emph{Restricting existing actions.}
    The DeepSeek observation-policy plugin blocks writes without a
    current observation. Pi's planning-mode extension blocks file changes
    while planning mode is active. Both add nonexecutability laws for
    actions already provided by the core, so the extension theorems do
    not guarantee that those actions remain executable. A designer must
    instead check calls against the added laws. With the observation-policy
    plugin loaded, the DeepSeek model requires another read before writing
    after an external file change. The Pi model requires leaving planning
    mode before a write.
\end{itemize}
Each application model has one optional module, so these tests do not
illustrate the two-module loading-order result.

\subsection{Configuration computation and planning on generated models}
\label{sec:controlled-study}
\label{sec:reasoning-performance}

We implement the causal rules of Section~\ref{sec:compilation} directly
in ASP for explicitly listed, state-independent initializer assignments.
We test two reasoning tasks on generated RAT systems. For a RAT system
\(\mathcal R\), an initial configuration \(\Config_0\), and an
integer \(h\geq0\), the tasks are as follows.
\begin{enumerate}[leftmargin=*]
  \item \emph{Configuration computation.} Given an action sequence
    \(\alpha=\langle a_0,\ldots,a_{h-1}\rangle\), find an execution
    \(\Config_0\trans{a_0}\cdots\trans{a_{h-1}}\Config_h\), or establish
    that none exists.
  \item \emph{Planning.} Given a goal and a set of allowed actions,
    find an execution from \(\Config_0\) that reaches the goal in
    at most \(h\) actions. The goal specifies fluent values and may also
    require modules to be active or inactive in the final configuration.
\end{enumerate}
In these models, the initial configuration and an executable action
sequence uniquely determine the configuration after each action.
For these models and the bounds below, one query with exactly \(h\)
actions decides the bounded planning task. The Supplementary Material
gives the argument. Timings measure that query, not incremental search
over lengths.

\paragraph{Three model families}
We increase \(n\), the number of tools, providers, or jobs in the
following models, so we can measure how computation time and
ground-program size grow with the problem size.
Complete laws and generation procedures are in the Supplementary Material.
\begin{itemize}[leftmargin=*]
  \item \emph{Tool sequences.} A harness may collect data, analyze it,
    and write a report using different tools. We model \(n\) tools,
    each an initially unloaded module. Each tool must be loaded before
    use, and the tools must be used once each in a fixed order.
    Configuration computation follows a supplied sequence that loads,
    uses, and unloads each tool in that order. Planning must use all
    tools and leave them unloaded. Each tool requires a load, a use,
    and an unload, so we use \(h=3n\). Loading and unloading need
    not follow the supplied sequence.
  \item \emph{Provider choices.} A harness may select optional services,
    such as search, translation, or storage, subject to compatibility
    requirements. We model \(n\) candidate providers as initially
    unloaded modules. For each provider in index order, an action
    either loads its module or records that it is not selected.
    Planning chooses all \(n\) decisions to satisfy a generated
    Boolean formula describing the allowed combinations, with \(h=n\).
    Configuration computation instead takes the choices as given and
    computes which modules are active after each action. 
  \item \emph{Resource allocation.} A harness may reserve a separate
    workspace for each job in a batch. Let \(n\) be the number of
    jobs and \(r\) the number of workspaces. Each workspace
    is represented by an initially unloaded resource module. Jobs
    are assigned in order to distinct loaded resources, with no
    unloading or reuse. With \(r=n\), a plan needs \(n\) loads and
    \(n\) assignments, so \(h=2n\). Configuration computation follows
    a supplied load-and-assign sequence for this case. We also test
    planning with \(r=n-1\) and \(h=2n-1\). There is then one fewer
    workspace than jobs, so no plan exists at any length.
\end{itemize}

\paragraph{Test cases and repetitions}
A test case consists of a model and one of the two reasoning queries
above. We use the following sizes and repetition counts.
\begin{itemize}[leftmargin=*]
  \item \emph{Tools.} We test \(n\in\{4,16,64,128,256\}\), with
    one case per size for each reasoning task. Each case is run three
    times, giving 5 cases and 15 runs per task.
  \item \emph{Providers.} We test \(n\in\{80,120,160,240,320\}\).
    At each size, 10 seeds generate 10 different instances. Each
    instance is tested once for configuration computation and once
    for planning, giving 50 cases and 50 runs per task.
  \item \emph{Resources.} Configuration computation uses
    \(n\in\{4,6,8,10,12,14,16,18,20,24,28,32\}\) and \(r=n\), giving 12 cases.
    Feasible planning uses \(n\in\{4,6,8,10,12,14,16,20\}\), giving
    8 cases with \(r=n\). Infeasible planning uses
    \(n\in\{4,6,8,10,12,14,16\}\), giving 7 cases with \(r=n-1\).
    Each case is run three times.
\end{itemize}

\paragraph{Measurement protocol}
We used a Mac mini with an Apple M4 processor (10 CPU cores), 24 GiB
RAM, macOS 26.2, Python 3.14.3, and clingo 5.8.2. Queries ran
sequentially, each in a fresh process with one solver thread. Each
query had a 1,200-second wall-clock budget and an 8 GiB worker
resident-memory stopping threshold. The supplement gives monitoring
details and observed memory overruns. Reported times cover model
construction, ASP generation, parsing and grounding, search, and
result decoding. We report a median query time only when every run
in its group finishes. Program size is
the number of ground rules, including choice rules and constraints,
not file size in bytes.
The project webpage\footnote{\artifacturl} provides the code,
complete query inputs, per-run records, and reproduction instructions.
The feasible planning case at \(n=20\) was added in a separate batch
using the same machine, frozen code, and measurement settings.

\begin{figure}[!htbp]
\centering
\includegraphics[width=\linewidth]{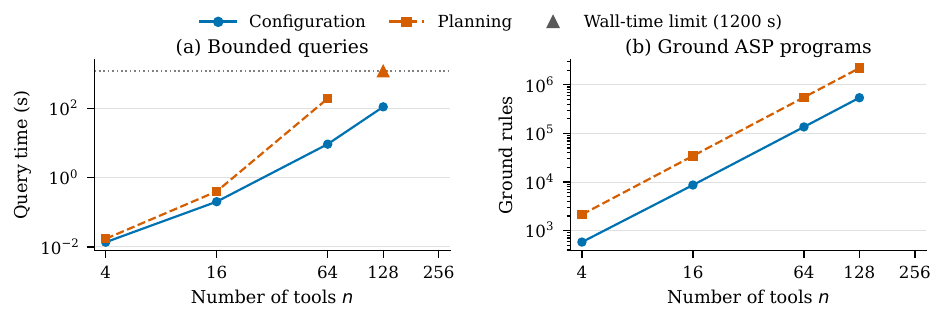}
\caption{Tool-sequence results, with
horizon \(3n\). Each point summarizes three runs of one case.
Bars show interquartile ranges. Triangles mark timeouts.}
\label{fig:bounded-tools}
\medskip
\includegraphics[width=\linewidth]{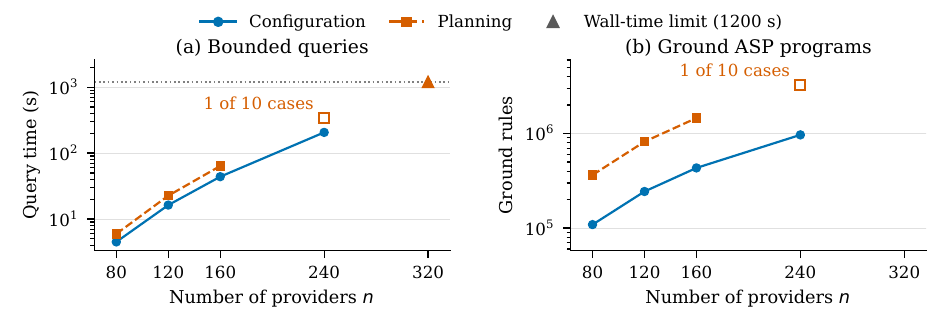}
\caption{Provider results, with
horizon \(n\). Each group has ten distinct cases, run once each.
The hollow points at \(n=240\) show the time and ground-rule count
of the only completed planning query, not ten-case medians.}
\label{fig:bounded-providers}
\medskip
\includegraphics[width=\linewidth]{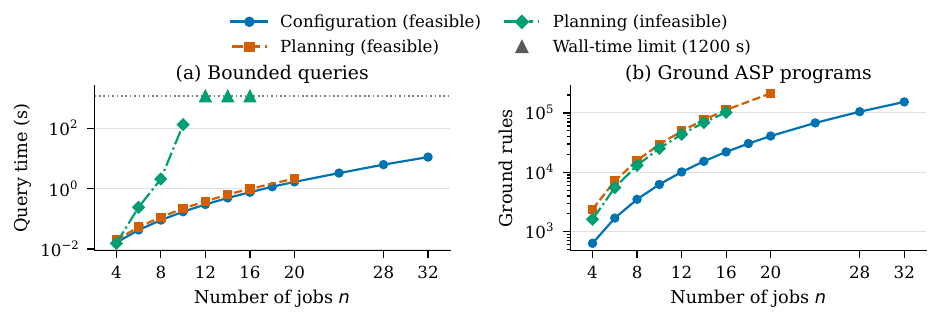}
\caption{Resource-allocation results for \(n\) jobs.
Each point summarizes three runs of one case. Feasible instances use
\(h=2n\), and infeasible instances use \(h=2n-1\). Triangles mark timeouts.}
\label{fig:bounded-capacity}
\end{figure}

\paragraph{Results}
Table~\ref{tab:bounded-results} summarizes the 137 test cases and
211 runs for the three model families. The \emph{Cases} column counts distinct queries, and \emph{Repeats}
gives the number of runs per case. Thus \emph{Runs} is their product.
The four outcome columns count runs as follows. For \emph{SAT}, the solver finds an execution satisfying the
    query. For \emph{UNSAT}, it reports that no execution satisfies
    the query.
\emph{Timeout} and \emph{Memory} count runs stopped at the
    time or memory limit without either answer.
Each run has exactly one of these four outcomes, so
\(
  \text{Runs}=\text{SAT}+\text{UNSAT}+\text{Timeout}+\text{Memory}.
\)
Figures~\ref{fig:bounded-tools}--\ref{fig:bounded-capacity} show how time
and ground-program size change with problem size. Full numerical
tables are in the Supplementary Material. 

\begin{table}[htbp]
\centering
\caption{Query and run counts for the three generated model families. C denotes configuration computation and P planning.}
\label{tab:bounded-results}
\small
\setlength{\tabcolsep}{4pt}
\begin{tabular}{@{}llrrrrrrr@{}}
\toprule
Family & Task & Cases & Repeats & Runs & SAT & UNSAT & Timeout & Memory \\
\midrule
Tools & C & 5 & 3 & 15 & 12 & 0 & 0 & 3 \\
Tools & P & 5 & 3 & 15 & 9 & 0 & 3 & 3 \\
Providers & C & 50 & 1 & 50 & 40 & 0 & 0 & 10 \\
Providers & P & 50 & 1 & 50 & 19 & 12 & 1 & 18 \\
Resources, feasible & C & 12 & 3 & 36 & 36 & 0 & 0 & 0 \\
Resources, feasible & P & 8 & 3 & 24 & 24 & 0 & 0 & 0 \\
Resources, infeasible & P & 7 & 3 & 21 & 0 & 12 & 9 & 0 \\
\midrule
Total & & 137 & -- & 211 & 140 & 24 & 13 & 34 \\
\bottomrule
\end{tabular}
\end{table}

Figures~\ref{fig:bounded-tools}--\ref{fig:bounded-capacity} compare
the two reasoning tasks as \(n\) increases.
\begin{enumerate}[leftmargin=*]
  \item \emph{Tool sequences require substantially more time for planning.}
    Both tasks finish all three runs for \(n=4,16,64\). For
    \(n=64\), median times are 9.37 seconds for configuration
    computation and 194.84 seconds for planning. For \(n=128\),
    configuration computation takes a median 112.18 seconds, while
    all three planning runs time out during search. For \(n=256\),
    all runs stop during grounding at the memory threshold.
  \item \emph{Provider choices are limited mainly by grounding.}
    Both tasks finish all ten cases at each of \(n=80,120,160\).
    For \(n=240\), configuration computation has median query and
    grounding times of 206.12 and 202.44 seconds. Only one planning
    case finishes (UNSAT), while nine stop during grounding at the
    memory threshold. The hollow points ``1 of 10 cases'' report the
    completed case's 342.46 seconds and 3,282,605 ground rules,
    not group averages.
    No case finishes for \(n=320\). Configuration-query SAT results
    concern execution of the given choices, not their compatibility.
  \item \emph{Infeasible resource queries can require substantial search.}
    All feasible cases finish all three runs. Median times are
    2.19 seconds for planning at \(n=20\) and 11.31 seconds for
    configuration computation at \(n=32\). The three planning runs
    at \(n=20\) each produce 214,161 ground rules. With one fewer resource,
    all runs return UNSAT for \(n=4,6,8,10\), taking a median
    134.95 seconds at \(n=10\). For \(n=12,14,16\), all nine runs
    time out during search despite having at most 102,341 ground
    rules. Resource counting proves infeasibility, but these solver
    results remain timeouts.
    Even without the allocation goal, at most \(2(n-1)\) actions can
    execute, fewer than the tested \(h=2n-1\). These timings therefore
    do not isolate the cost of checking the allocation goal.
\end{enumerate}

\paragraph{Checking the reported answers}
A separate checker validated all 140 returned SAT executions in the
reported study, comprising 12,756 transitions, directly against the RAT
laws. The Supplementary Material describes these checks, the
resource-allocation infeasibility argument, and the absence of independent
certificates for the 12 provider UNSAT results. Runs stopped by time or
memory limits remain without a solver answer.

\section{Related work}
\label{sec:related}

RAT studies actions that change the modules forming an action description.
The closest work concerns modular action languages, theory change, and
runtime reconfiguration.

\paragraph{Action languages and modular descriptions}
Action languages and situation calculus provide accounts of action
effects, inertia, and executability
~\cite{mccainTurner1997,giunchigliaEtAl2004,gelfondLifschitz1998,giunchigliaLifschitz1998,baral2003,reiter2001}.
RAT retains the ordinary causal semantics of a fixed action description.
Its difference from modular action languages concerns
when modules are composed. MAD imports modules, with renaming and
redefinition, to construct an action description~\cite{lifschitzRen2006}.
ALM organizes reusable knowledge into module hierarchies and separates
a theory from the structure specifying its objects
~\cite{inclezanGelfond2015,inclezanGelfond2016}.
RAT instead represents the currently active modules in each configuration.
Loading and unloading are actions that change which vocabulary and laws
apply during an execution. A module also supplies initial assignments
for its private basic fluents when it is loaded.
Thus MAD and ALM provide modular descriptions of a domain, whereas RAT
directly describes changes to the active modules within that domain.
\paragraph{Theory change and software reconfiguration}
Action-domain update and action-theory revision or contraction change
descriptions according to constraints, preferences, or minimal-change
criteria~\cite{eiterErdemFinkSenko2010,varzinczak2010}.
Evolp Action Programs allow rules to be asserted during execution,
including rules governing further change~\cite{alferesBantiBrogi2005}.
Runtime architectures, self-adaptation, and dynamic software updating
address component replacement and state migration
~\cite{kramerMagee1990,oreizyEtAl1998,garlanEtAl2004,hicksMooreNettles2001}.
RAT has a narrower purpose. It describes execution with a fixed catalogue
of modules whose membership in the active theory can change. It does not
revise laws to accommodate new information, generate arbitrary rules,
or give later rules priority over earlier ones. Nor does it implement
software replacement or state migration. Instead, its causal laws specify
the permitted state changes when declared modules are loaded or unloaded.
RAT inherits the fixed-description causal semantics. Its additional
results concern private state introduced or removed by module changes
and conditions for preserving behavior or changing loading order.
These results make no claim of greater expressive power than evolving
programs.

\paragraph{Cordis and the DeepSeek Harness theory}
The context paradigm underlying Cordis and DeepSeek Harness
~\cite{shiZhangCui2026} represents component effects by context
transformations with associated inverse operations. It also records
dependencies so that component lifecycles can respond to changes in
their providers. Its central questions concern undoing effects and
restoring dependent components under stated recovery assumptions.
RAT modules instead contain vocabulary, causal laws, and initializers.
They specify which actions and states are possible before and after a
module change, without prescribing a recovery procedure. In particular,
unloading a RAT module need not reverse its previous effects, and RAT
does not automatically restart work that depended on it. Cordis therefore
provides an account of dependency management and recovery, while RAT
provides an action-theory account of the changing laws and their permitted
executions. Relating the two requires a model of the relevant Cordis
operations, not merely identifying each plugin with a RAT module.

\section{Discussion and conclusion}
\label{sec:conclusion}

\subsection{Why use RAT if causal theories suffice?}
\label{sec:rat-value}

For a finite module catalogue and execution bound,
Section~\ref{sec:compilation} translates RAT into an ordinary causal
theory. RAT therefore does not add expressive power in this setting.
Its purpose is to let a designer describe component changes in terms of
modules, rather than encode all their consequences directly as causal rules.

In a direct encoding, the designer must represent module activation,
guard the corresponding laws, and account for initialization and the
removal of private fluents. In RAT, the designer declares each module's
vocabulary, laws, and initializer. The semantics and translation handle
these consequences of loading and unloading uniformly. For example,
removing the observation-policy plugin in
Section~\ref{sec:harness-applications} removes its observation record
and the associated blocking laws, but leaves the file and file actions
in the core. The module declaration states this separation explicitly.

RAT also states conditions for preserving behavior and changing loading
order in terms of these declarations, as shown in
Section~\ref{sec:rat-properties}. A designer can apply a result after checking its hypotheses for the RAT
system at hand, without reproving it for the causal encoding. The translation then makes the same model available to existing
reasoners.

\subsection{Relation to harness architectures}
\label{sec:harness-value}

In a RAT configuration \(\langle K,s\rangle\), \(K\) selects
the active vocabulary and laws, while \(s\) records the current fluent
values. The applications in Section~\ref{sec:harness-applications} use
this distinction to separate removing a tool or its blocking laws from
changing its visibility or permission under fixed laws. Modeling another harness requires deciding which operations change \(K\),
possibly together with \(s\), and which change only \(s\). The designer
then specifies the relevant vocabulary, laws, and initializers and checks
the predicted transitions against the implementation.

\subsection{LLM-assisted modeling and reasoning}
\label{sec:llm-assistance}

We outline a possible workflow in which an LLM drafts RAT descriptions
and proposes executions for formal checking. We have not evaluated its
effect on modeling effort or planning time. Following
LLM-assisted modeling and externally checked planning
~\cite{guanEtAl2023,kambhampatiEtAl2024,linEtAl2025clmasp}, we suggest three steps.

\begin{enumerate}[leftmargin=*]
  \item \emph{Describe the components.}
    An LLM would draft fluents, actions, laws, and initializers from interfaces,
    documentation, and recorded executions. A developer would review these
    choices and test expected and forbidden transitions. Formal checks
    can detect inconsistent descriptions.
  \item \emph{Propose and check an execution.}
    For an approved RAT model and goal, the LLM would supply actions and
    complete configurations. A checker would verify every transition and
    the final goal. By Theorem~\ref{thm:rat-complexity}, this takes
    polynomial time in the model, goal, and fully specified execution
    under the stated input representation. Supplying only the actions
    leaves an \(\mathsf{NP}\)-complete problem of finding suitable
    configurations. A failed check can guide a revised proposal.
  \item \emph{Compare with execution.}
    In this proposed workflow, a runtime checker would compare each
    observed transition with the RAT model.
    If the successor is permitted but differs from the proposed configuration,
    it would check the remaining execution from the observed successor before
    deciding whether to replan. An observation outside the model would require
    reviewing the model.
\end{enumerate}

\subsection{Limitations}

\begin{itemize}[leftmargin=*]
  \item \emph{Language and execution.}
    RAT does not directly describe concurrent or overlapping tool calls.
    Each step contains one action, so representing a call that remains
    in progress requires explicit start and finish actions and additional
    state. RAT also cannot load a module outside the declared catalogue.
    It has no built-in account of unfinished requests, resource commitments,
    or recovery when a component disappears. Such behavior needs additional
    laws or an extension of the semantics.
  \item \emph{Modeling and reasoning cost.}
    Writing a faithful RAT model requires identifying the relevant state,
    effects, blocking conditions, and initialization behavior. Tool
    interfaces alone do not provide this information, and formal checks
    cannot establish that the chosen laws match the software.
    Reasoning can also be expensive. Finding an execution remains hard
    even when checking a fully specified execution is efficient
    (Section~\ref{sec:rat-complexity}). The experiments in
    Section~\ref{sec:reasoning-performance} show that both grounding size
    and search time can limit the ASP implementation. 
    These measurements evaluate this encoding and solver, without
    comparison against alternative encodings or solvers.
\end{itemize}

\subsection{Future work}
\label{sec:future-work}

RAT describes how actions change an action theory. A proposed extension,
\emph{Harness Action Theory} (HAT), asks what happens to unfinished work
when that theory changes. Replacing a provider, for example, does not
complete a request already sent to it or release resources held by that
request. HAT would record provider choices, performed actions, and
recovery information, so that removal can account for work that must
finish or be recovered first.

\subsection{Conclusion}
RAT is an action theory in which actions can change the action theory
itself by loading and unloading declared modules. It retains ordinary
causal semantics when the active modules remain fixed. Its results give
sufficient conditions for at most one next configuration, preservation
of existing behavior, and the same possible final configurations under
either loading order.
For finite RAT systems, the complexity results and the causal and ASP
translations provide a basis for execution checking and bounded plan search.
RAT thus allows a designer to specify component changes and analyze their
effects on a harness's available actions, possible executions, and reachable
goals.

\section*{Data and code availability}
The project webpage, \artifacturl, provides the code, Supplementary
Material, and experimental records. The downloads include RAT models,
causal and ASP evaluators, harness observations, query inputs, returned
executions, and checking and plotting scripts. Source hashes and pinned
dependencies identify the tested versions. Instructions distinguish
checking saved records from running the harnesses or measurements again.
No production user data or authenticated service data were collected.

\section*{Declaration of generative AI and AI-assisted technologies in the manuscript preparation process}

During the preparation of this work, the author used OpenAI Codex to assist
with manuscript organization, drafting and revision of text, improvement of
language and readability, \LaTeX{} editing, and preparation and revision of
the ASP encoding, scenarios, queries, and accompanying documentation.
After using this tool, the author critically
reviewed, verified, and edited all AI-assisted material as needed and takes
full responsibility for the content of the publication.

\bibliographystyle{elsarticle-num}
\bibliography{references}

@article{gelfondLifschitz1998,
  author  = {Gelfond, Michael and Lifschitz, Vladimir},
  title   = {Action Languages},
  journal = {Electron. Trans. Artif. Intell.},
  volume  = {3},
  pages   = {195--210},
  year    = {1998},
}

@inproceedings{gelfondLifschitz1988,
  author    = {Gelfond, Michael and Lifschitz, Vladimir},
  title     = {The Stable Model Semantics for Logic Programming},
  booktitle = {Proceedings of the Joint International Conference and Symposium on Logic Programming (ICLP/SLP 1988)},
  pages     = {1070--1080},
  year      = {1988}
}

@article{gelfondLifschitz1991,
  author  = {Gelfond, Michael and Lifschitz, Vladimir},
  title   = {Classical Negation in Logic Programs and Disjunctive Databases},
  journal = {New Gener. Comput.},
  volume  = {9},
  number  = {3--4},
  pages   = {365--385},
  year    = {1991}
}

@article{giunchigliaEtAl2004,
  author  = {Giunchiglia, Enrico and Lee, Joohyung and Lifschitz, Vladimir and McCain, Norman and Turner, Hudson},
  title   = {Nonmonotonic Causal Theories},
  journal = {Artif. Intell.},
  volume  = {153},
  number  = {1--2},
  pages   = {49--104},
  year    = {2004}
}

@article{bylander1994,
  author  = {Bylander, Tom},
  title   = {The Computational Complexity of Propositional {STRIPS} Planning},
  journal = {Artif. Intell.},
  volume  = {69},
  number  = {1--2},
  pages   = {165--204},
  year    = {1994}
}

@article{lifschitzTangTurner1999,
  author  = {Lifschitz, Vladimir and Tang, Lappoon R. and Turner, Hudson},
  title   = {Nested Expressions in Logic Programs},
  journal = {Ann. Math. Artif. Intell.},
  volume  = {25},
  number  = {3--4},
  pages   = {369--389},
  year    = {1999}
}

@article{ferrarisEtAl2012,
  author  = {Ferraris, Paolo and Lee, Joohyung and Lierler, Yuliya and Lifschitz, Vladimir and Yang, Fangkai},
  title   = {Representing First-Order Causal Theories by Logic Programs},
  journal = {Theory Pract. Log. Program.},
  volume  = {12},
  number  = {3},
  pages   = {383--412},
  year    = {2012}
}

@inproceedings{pearceEtAl2002,
  author    = {Pearce, David and Sarsakov, Vladimir and Schaub, Torsten and Tompits, Hans and Woltran, Stefan},
  title     = {A Polynomial Translation of Logic Programs with Nested Expressions into Disjunctive Logic Programs: Preliminary Report},
  booktitle = {Proceedings of the 18th International Conference on Logic Programming (ICLP 2002)},
  pages     = {405--420},
  year      = {2002}
}

@inproceedings{jiLin2013,
  author    = {Ji, Jianmin and Lin, Fangzhen},
  title     = {{Turner}'s Logic of Universal Causation, Propositional Logic, and Logic Programming},
  booktitle = {Proceedings of the 12th International Conference on Logic Programming and Nonmonotonic Reasoning (LPNMR 2013)},
  pages     = {401--413},
  year      = {2013}
}

@inproceedings{giunchigliaLifschitz1998,
  author    = {Giunchiglia, Enrico and Lifschitz, Vladimir},
  title     = {An Action Language Based on Causal Explanation: Preliminary Report},
  booktitle = {Proceedings of the 15th National Conference on Artificial Intelligence (AAAI 1998)},
  pages     = {623--630},
  year      = {1998},
}

@inproceedings{lifschitzTurner1999,
  author    = {Lifschitz, Vladimir and Turner, Hudson},
  title     = {Representing Transition Systems by Logic Programs},
  booktitle = {Proceedings of the 5th International Conference on Logic Programming and Nonmonotonic Reasoning (LPNMR 1999)},
  pages     = {92--106},
  year      = {1999},
}

@book{reiter2001,
  author    = {Reiter, Raymond},
  title     = {Knowledge in Action: Logical Foundations for Specifying and Implementing Dynamical Systems},
  publisher = {MIT Press},
  address   = {Cambridge, MA},
  year      = {2001}
}

@inproceedings{lifschitzRen2006,
  author    = {Lifschitz, Vladimir and Ren, Wanwan},
  title     = {A Modular Action Description Language},
  booktitle = {Proceedings of the 21st National Conference on Artificial Intelligence (AAAI 2006)},
  pages     = {853--859},
  year      = {2006}
}

@inproceedings{inclezanGelfond2015,
  author    = {Inclezan, Daniela},
  title     = {A {CLib}-Inspired Library of Commonsense Knowledge in Modular Action Language {ALM}},
  booktitle = {Proceedings of the 2015 AAAI Spring Symposium on Logical Formalizations of Commonsense Reasoning (Commonsense 2015)},
  pages     = {82--88},
  year      = {2015},
}

@article{inclezanGelfond2016,
  author  = {Inclezan, Daniela and Gelfond, Michael},
  title   = {Modular Action Language {ALM}},
  journal = {Theory Pract. Log. Program.},
  volume  = {16},
  number  = {2},
  pages   = {189--235},
  year    = {2016}
}

@inproceedings{alferesBantiBrogi2005,
  author    = {Alferes, Jos{\'e} J{\'u}lio and Banti, Federico and Brogi, Antonio},
  title     = {From Logic Programs Updates to Action Description Updates},
  booktitle = {Proceedings of the 5th International Workshop on Computational Logic in Multi-Agent Systems (CLIMA V)},
  pages     = {52--77},
  year      = {2005}
}

@article{eiterErdemFinkSenko2010,
  author  = {Eiter, Thomas and Erdem, Esra and Fink, Michael and Senko, J{\'a}n},
  title   = {Updating Action Domain Descriptions},
  journal = {Artif. Intell.},
  volume  = {174},
  number  = {15},
  pages   = {1172--1221},
  year    = {2010}
}

@article{varzinczak2010,
  author  = {Varzinczak, Ivan Jos{\'e}},
  title   = {On Action Theory Change},
  journal = {J. Artif. Intell. Res.},
  volume  = {37},
  pages   = {189--246},
  year    = {2010}
}

@article{gebserEtAl2019,
  author  = {Gebser, Martin and Kaminski, Roland and Kaufmann, Benjamin and Schaub, Torsten},
  title   = {Multi-shot {ASP} Solving with clingo},
  journal = {Theory Pract. Log. Program.},
  volume  = {19},
  number  = {1},
  pages   = {27--82},
  year    = {2019}
}

@inproceedings{gebserSchaubThiele2007gringo,
  author    = {Gebser, Martin and Schaub, Torsten and Thiele, Sven},
  title     = {{GrinGo}: A New Grounder for Answer Set Programming},
  booktitle = {Proceedings of the 9th International Conference on Logic Programming and Nonmonotonic Reasoning (LPNMR 2007)},
  pages     = {266--271},
  year      = {2007}
}

@inproceedings{gebserEtAl2007clasp,
  author    = {Gebser, Martin and Kaufmann, Benjamin and Neumann, Andr{\'e} and Schaub, Torsten},
  title     = {{clasp}: A Conflict-Driven Answer Set Solver},
  booktitle = {Proceedings of the 9th International Conference on Logic Programming and Nonmonotonic Reasoning (LPNMR 2007)},
  pages     = {260--265},
  year      = {2007}
}

@inproceedings{yao2023react,
  author    = {Yao, Shunyu and Zhao, Jeffrey and Yu, Dian and Du, Nan and Shafran, Izhak and Narasimhan, Karthik and Cao, Yuan},
  title     = {{ReAct}: Synergizing Reasoning and Acting in Language Models},
  booktitle = {Proceedings of the 11th International Conference on Learning Representations (ICLR 2023)},
  year      = {2023}
}

@article{wangEtAl2024,
  author  = {Wang, Lei and Ma, Chen and Feng, Xueyang and Zhang, Zeyu and Yang, Hao and Zhang, Jingsen and others},
  title   = {A Survey on Large Language Model Based Autonomous Agents},
  journal = {Front. Comput. Sci.},
  volume  = {18},
  number  = {6},
  pages   = {186345},
  year    = {2024}
}

@article{qinEtAl2025,
  author  = {Qin, Yujia and Hu, Shengding and Lin, Yankai and Chen, Weize and Ding, Ning and Cui, Ganqu and others},
  title   = {Tool Learning with Foundation Models},
  journal = {ACM Comput. Surv.},
  volume  = {57},
  number  = {4},
  pages   = {101:1--101:40},
  year    = {2025}
}

@misc{liEtAl2026Harness,
  author       = {Li, Junjie and Xiao, Xi and Zhang, Yunbei and Liu, Chen and Zhao, Lin and Liao, Xiaoying and others},
  title        = {Agent Harness Engineering: A Survey},
  howpublished = {Online preprint},
  url          = {https://picrew.github.io/LLM-Harness/},
  month        = may,
  year         = {2026}
}

@inproceedings{wangEtAl2025OpenHands,
  author    = {Wang, Xingyao and Li, Boxuan and Song, Yufan and Xu, Frank F. and Tang, Xiangru and Zhuge, Mingchen and others},
  title     = {{OpenHands}: An Open Platform for {AI} Software Developers as Generalist Agents},
  booktitle = {Proceedings of the 13th International Conference on Learning Representations (ICLR 2025)},
  year      = {2025}
}

@article{kramerMagee1990,
  author    = {Kramer, Jeff and Magee, Jeff},
  title     = {The Evolving Philosophers Problem: Dynamic Change Management},
  journal   = {IEEE Trans. Softw. Eng.},
  volume    = {16},
  number    = {11},
  pages     = {1293--1306},
  year      = {1990}
}

@inproceedings{hicksMooreNettles2001,
  author    = {Hicks, Michael and Moore, Jonathan T. and Nettles, Scott},
  title     = {Dynamic Software Updating},
  booktitle = {Proceedings of the ACM SIGPLAN Conference on Programming Language Design and Implementation (PLDI 2001)},
  pages     = {13--23},
  year      = {2001}
}

@misc{shiZhangCui2026,
  author  = {Shi, Yifan and Zhang, Wei and Cui, Tianyi},
  title   = {A Programming Paradigm for Spatiotemporal Composability},
  howpublished = {arXiv preprint arXiv:2608.25512},
  url     = {https://arxiv.org/abs/2608.25512},
  year    = {2026}
}

@misc{deepseekArchitecture,
  author       = {{DeepSeek AI}},
  title        = {{DeepSeek} Harness Architecture},
  howpublished = {Online documentation},
  url          = {https://github.com/deepseek-ai/deepseek-harness/blob/master/docs/architecture.md},
  year         = {2026},
  note         = {{Accessed} 9 October 2026}
}

@misc{piSdk,
  author       = {{Earendil Inc.}},
  title        = {{Pi} Coding Agent {SDK} Documentation},
  howpublished = {Online documentation},
  url          = {https://pi.dev/docs/latest/sdk},
  year         = {2026},
  note         = {{Accessed} 9 October 2026}
}

@misc{openclawArchitecture,
  author       = {{OpenClaw Foundation}},
  title        = {Gateway Architecture},
  howpublished = {Online documentation},
  url          = {https://docs.openclaw.ai/concepts/architecture},
  year         = {2026},
  note         = {{Accessed} 9 October 2026}
}

@misc{codexAppServer,
  author       = {{OpenAI}},
  title        = {{Codex} App Server},
  howpublished = {Online documentation},
  url          = {https://learn.chatgpt.com/docs/app-server},
  year         = {2026},
  note         = {{Accessed} 9 October 2026}
}

@inproceedings{mccainTurner1997,
  author    = {McCain, Norman and Turner, Hudson},
  title     = {Causal Theories of Action and Change},
  booktitle = {Proceedings of the 14th National Conference on Artificial Intelligence (AAAI 1997)},
  pages     = {460--465},
  year      = {1997}
}

@book{baral2003,
  author    = {Baral, Chitta},
  title     = {Knowledge Representation, Reasoning and Declarative Problem Solving},
  publisher = {Cambridge University Press},
  address   = {Cambridge},
  year      = {2003}
}

@article{brewkaEiterTruszczynski2011,
  author  = {Brewka, Gerhard and Eiter, Thomas and Truszczy{\'n}ski, Miros{\l}aw},
  title   = {Answer Set Programming at a Glance},
  journal = {Commun. ACM},
  volume  = {54},
  number  = {12},
  pages   = {92--103},
  year    = {2011}
}

@inproceedings{oreizyEtAl1998,
  author    = {Oreizy, Peyman and Medvidovic, Nenad and Taylor, Richard N.},
  title     = {Architecture-Based Runtime Software Evolution},
  booktitle = {Proceedings of the 20th International Conference on Software Engineering (ICSE 1998)},
  pages     = {177--186},
  year      = {1998}
}

@article{garlanEtAl2004,
  author  = {Garlan, David and Cheng, Shang-Wen and Huang, An-Cheng and Schmerl, Bradley and Steenkiste, Peter},
  title   = {Rainbow: Architecture-Based Self-Adaptation with Reusable Infrastructure},
  journal = {Computer},
  volume  = {37},
  number  = {10},
  pages   = {46--54},
  year    = {2004}
}

@inproceedings{guanEtAl2023,
  author    = {Guan, Lin and Valmeekam, Karthik and Sreedharan, Sarath and Kambhampati, Subbarao},
  title     = {Leveraging Pre-trained Large Language Models to Construct and Utilize World Models for Model-based Task Planning},
  booktitle = {Advances in Neural Information Processing Systems (NeurIPS 2023)},
  pages     = {79081--79094},
  year      = {2023}
}

@inproceedings{kambhampatiEtAl2024,
  author    = {Kambhampati, Subbarao and Valmeekam, Karthik and Guan, Lin and Verma, Mudit and Stechly, Kaya and Bhambri, Siddhant and Saldyt, Lucas Paul and B Murthy, Anil},
  title     = {Position: {LLM}s Can't Plan, But Can Help Planning in {LLM}-Modulo Frameworks},
  booktitle = {Proceedings of the 41st International Conference on Machine Learning (ICML 2024)},
  pages     = {22895--22907},
  year      = {2024}
}

@inproceedings{linEtAl2025clmasp,
  author    = {Lin, Xinrui and Wu, Yangfan and Yang, Huanyu and Huang, Yuting and Zhang, Yu and Ji, Jianmin and Zhang, Yanyong},
  title     = {Improving Efficiency of Answer Set Planning with Rough Solutions from Large Language Models for Robotic Task Planning},
  booktitle = {Proceedings of the Thirty-Fourth International Joint Conference on Artificial Intelligence ({IJCAI-25})},
  pages     = {4570--4578},
  year      = {2025}
}
\end{document}